\documentclass[10pt,twocolumn,letterpaper]{article}
\usepackage{iccv}

\definecolor{iccvblue}{rgb}{0.21,0.49,0.74}
\usepackage[pagebackref,breaklinks,colorlinks,allcolors=iccvblue]{hyperref}
\usepackage{xcolor}  
\usepackage{pifont}
\usepackage{tabularx}
\usepackage{graphicx}
\usepackage{subcaption}
\usepackage{algorithm}
\usepackage{algpseudocode}
\usepackage{adjustbox}
\usepackage{array}
\usepackage{multirow}
\usepackage{makecell}
\usepackage{xcolor,colortbl}
\usepackage{dsfont}
\usepackage{pifont}
\usepackage{color} 
\definecolor{LightGray}{RGB}{220, 220, 240}
\definecolor{linkcolor}{RGB}{200, 50, 120}  

\newcommand{\datasetlabel}[1]{\rotatebox{90}{$\;\;\;\;\;\;\;\;\;\;\;\;\;\;$#1}}

\def\paperID{13670} 
\def\confName{ICCV}
\def\confYear{2025}

\title{Sparsity Outperforms Low-Rank Projections in Few-Shot Adaptation}

\author{
Nairouz Mrabah \quad
Nicolas Richet \quad
Ismail Ben Ayed \quad
Eric Granger \\
LIVIA, ILLS, Department of Systems Engineering, ÉTS Montreal, Québec, Canada \\
{\tt\small mrabah.nairouz@livia.etsmtl.ca, nicolas.richet.1@ens.etsmtl.ca,}\\ 
{\tt\small ismail.benayed@etsmtl.ca, eric.granger@etsmtl.ca}
}

\begin{document}
\maketitle

\begin{abstract}
Adapting vision-language models (VLMs) to new domains with few labeled samples remains a significant challenge due to severe overfitting and computational constraints. 
State-of-the-art solutions, such as low-rank reparameterization, mitigate these issues but often struggle with generalization and require extensive hyperparameter tuning.  
%
In this paper, a novel Sparse Optimization (SO) framework is proposed. Unlike low-rank approaches that typically constrain updates to a fixed subspace, our SO method leverages high sparsity to dynamically adjust very few parameters. We introduce two key paradigms. First, we advocate for \textit{local sparsity and global density}, which updates a minimal subset of parameters per iteration while maintaining overall model expressiveness. As a second paradigm, we advocate for \textit{local randomness and global importance}, which sparsifies the gradient using random selection while pruning the first moment based on importance. This combination significantly mitigates overfitting and ensures stable adaptation in low-data regimes. 
%
Extensive experiments on 11 diverse datasets show that SO achieves state-of-the-art few-shot adaptation performance while reducing memory overhead. 
Code is available at: \textcolor{linkcolor}{\href{https://github.com/nairouz/SO}{https://github.com/nairouz/SO}}.
\end{abstract}

\section{Introduction}
\label{sec:intro}

\begin{figure*}
\centering
\renewcommand{\arraystretch}{1.3} 
\begin{tabular}{c@{\hskip 10pt} c c c c} 
    & Pets & Flowers & Caltech & UCF \\
    \datasetlabel{LoRA} & 
    \begin{subfigure}[b]{0.22\textwidth}
        \centering
        \includegraphics[width=\linewidth]{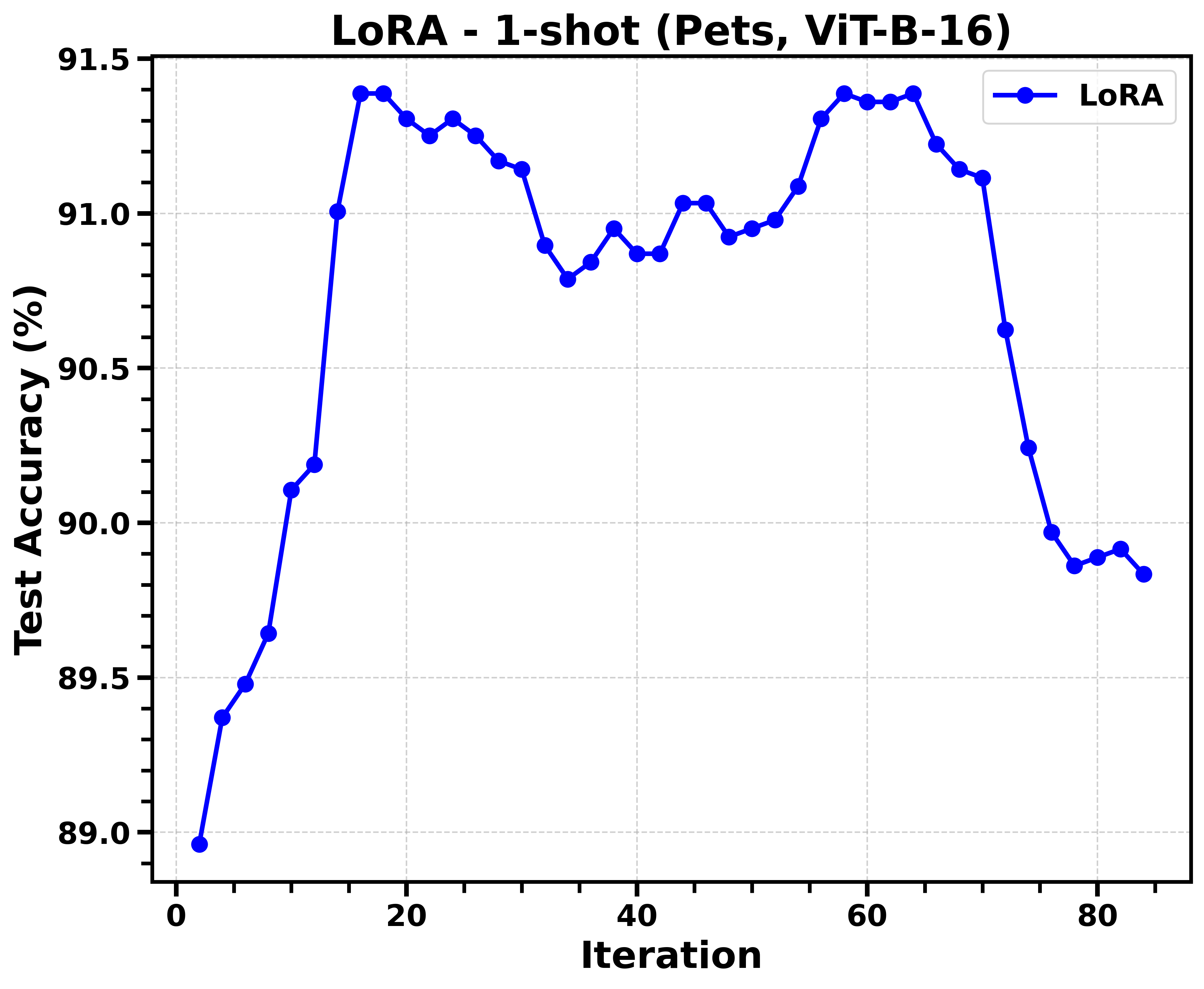}
    \end{subfigure} &
    \begin{subfigure}[b]{0.22\textwidth}
        \centering
        \includegraphics[width=\linewidth]{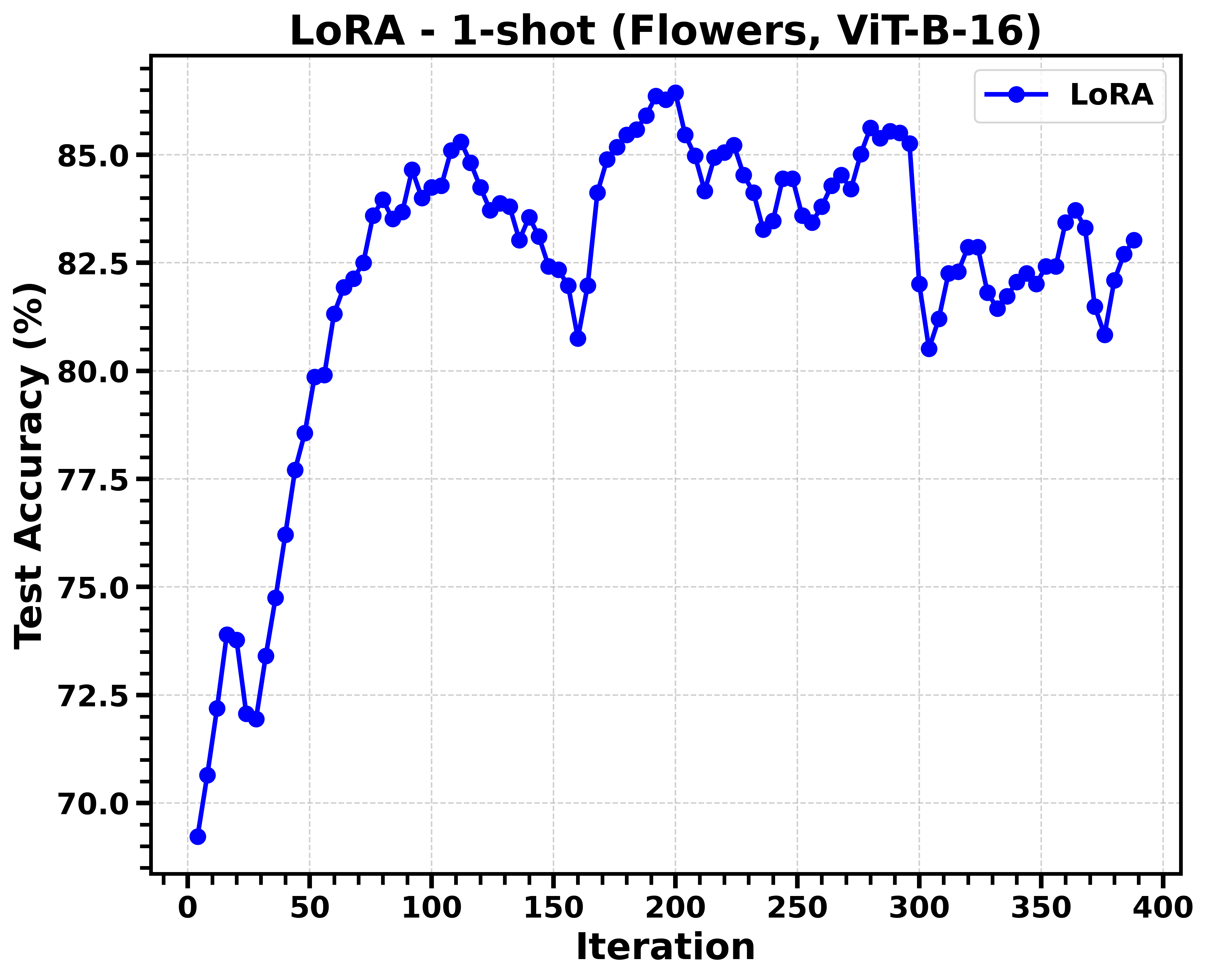}
    \end{subfigure} &
    \begin{subfigure}[b]{0.22\textwidth}
        \centering
        \includegraphics[width=\linewidth]{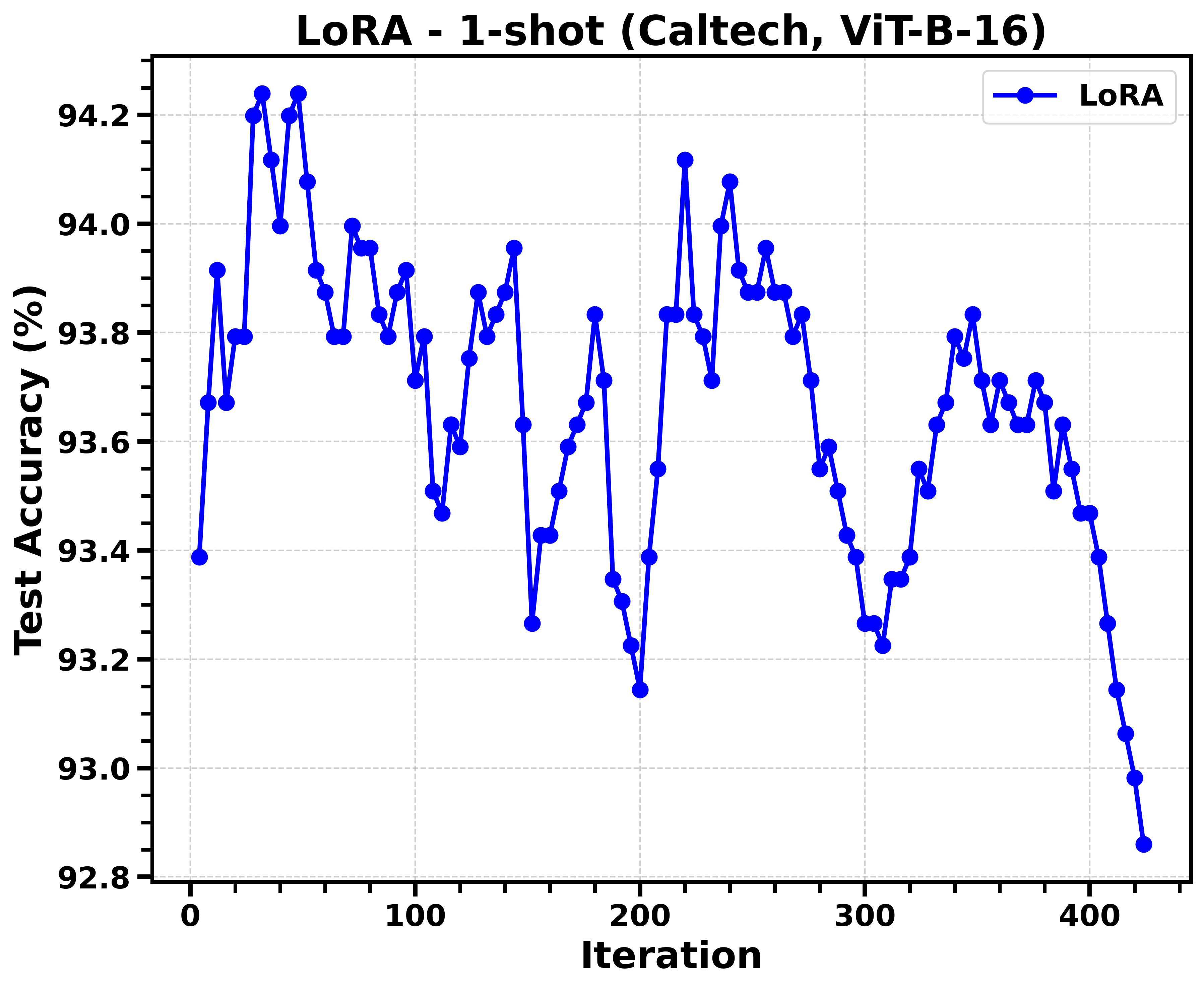}
    \end{subfigure} &
    \begin{subfigure}[b]{0.22\textwidth}
        \centering
        \includegraphics[width=\linewidth]{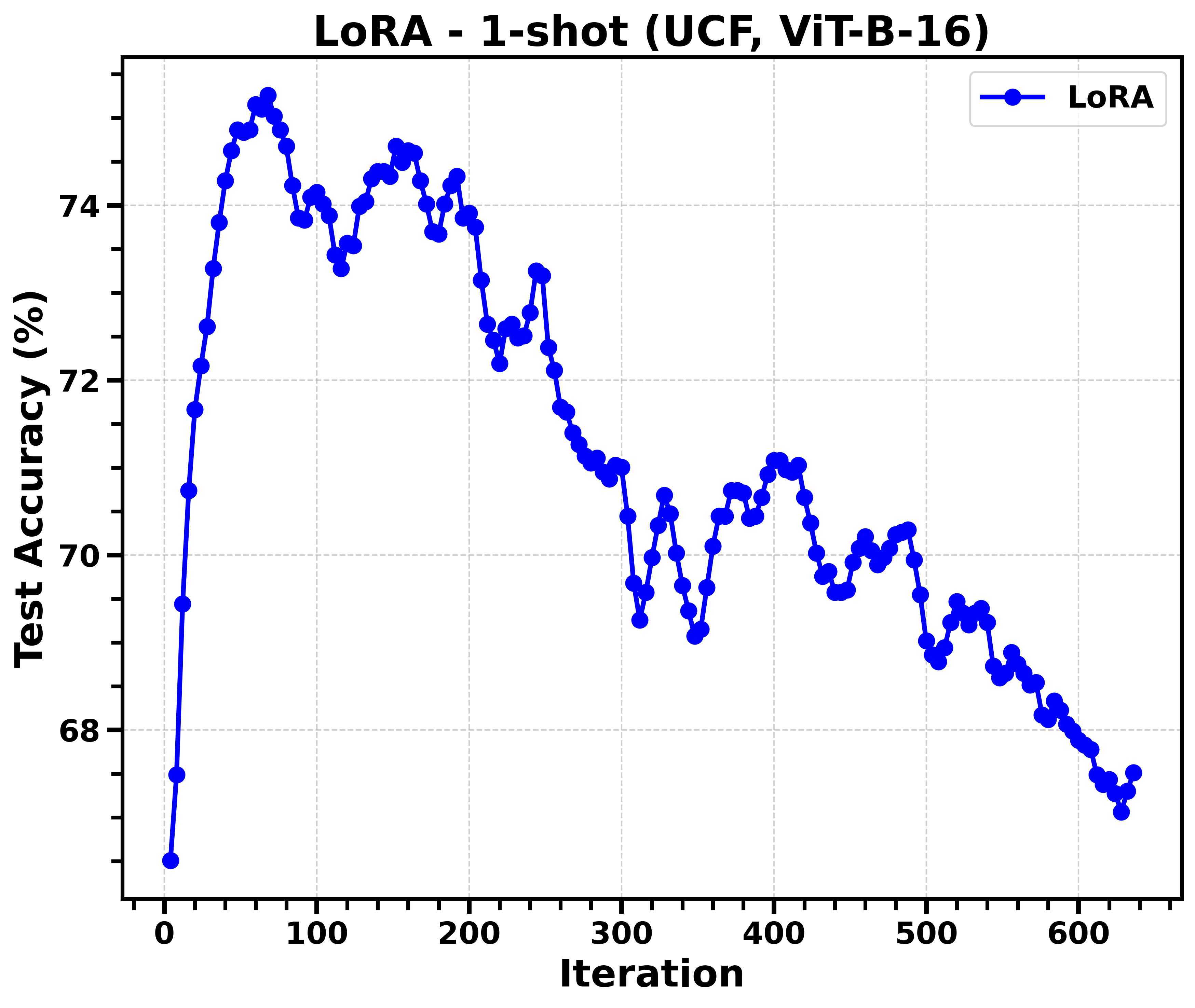}
    \end{subfigure} \\
    \datasetlabel{SO} & 
    \begin{subfigure}[b]{0.22\textwidth}
        \centering
        \includegraphics[width=\linewidth]{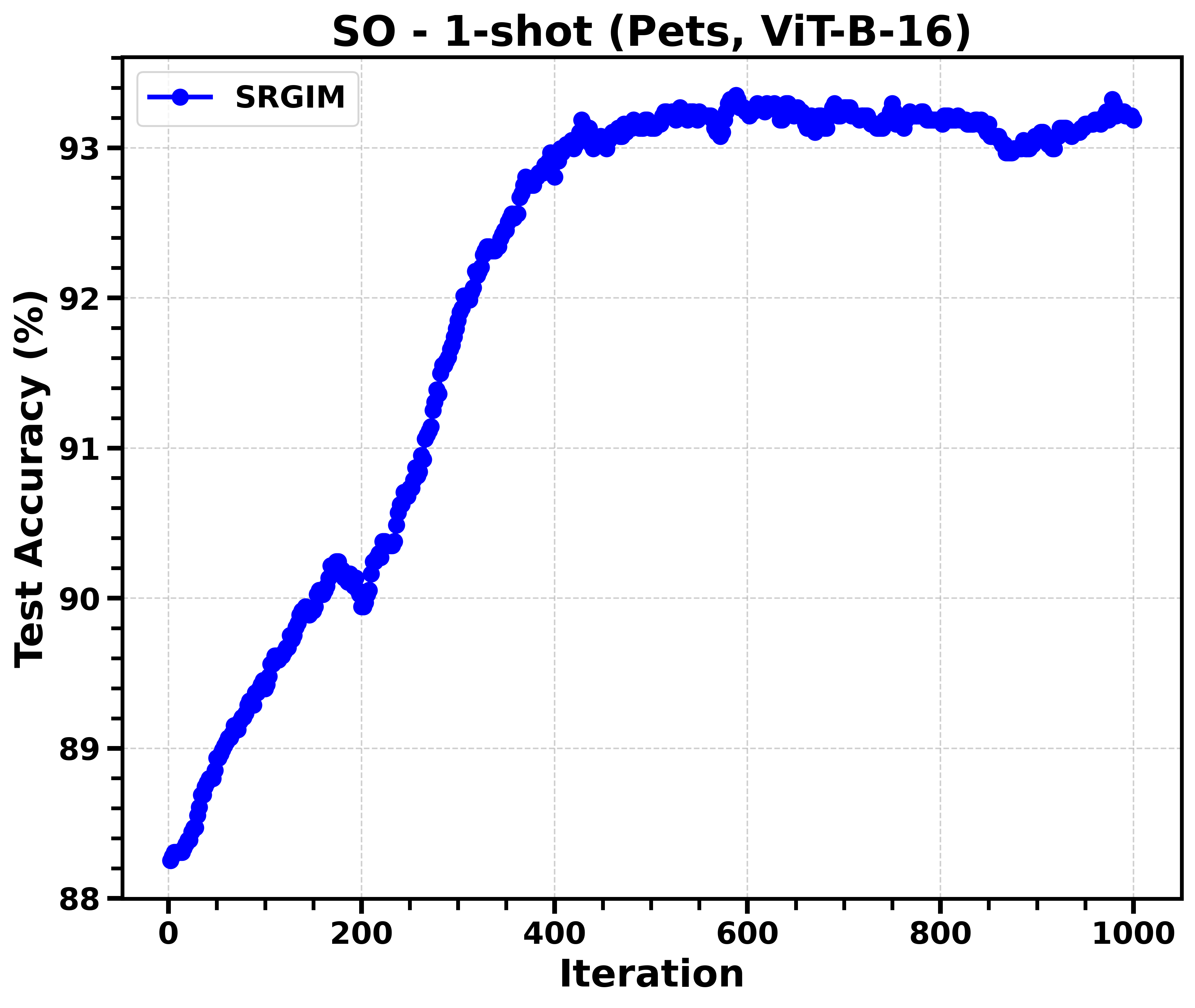}
    \end{subfigure} &
    \begin{subfigure}[b]{0.22\textwidth}
        \centering
        \includegraphics[width=\linewidth]{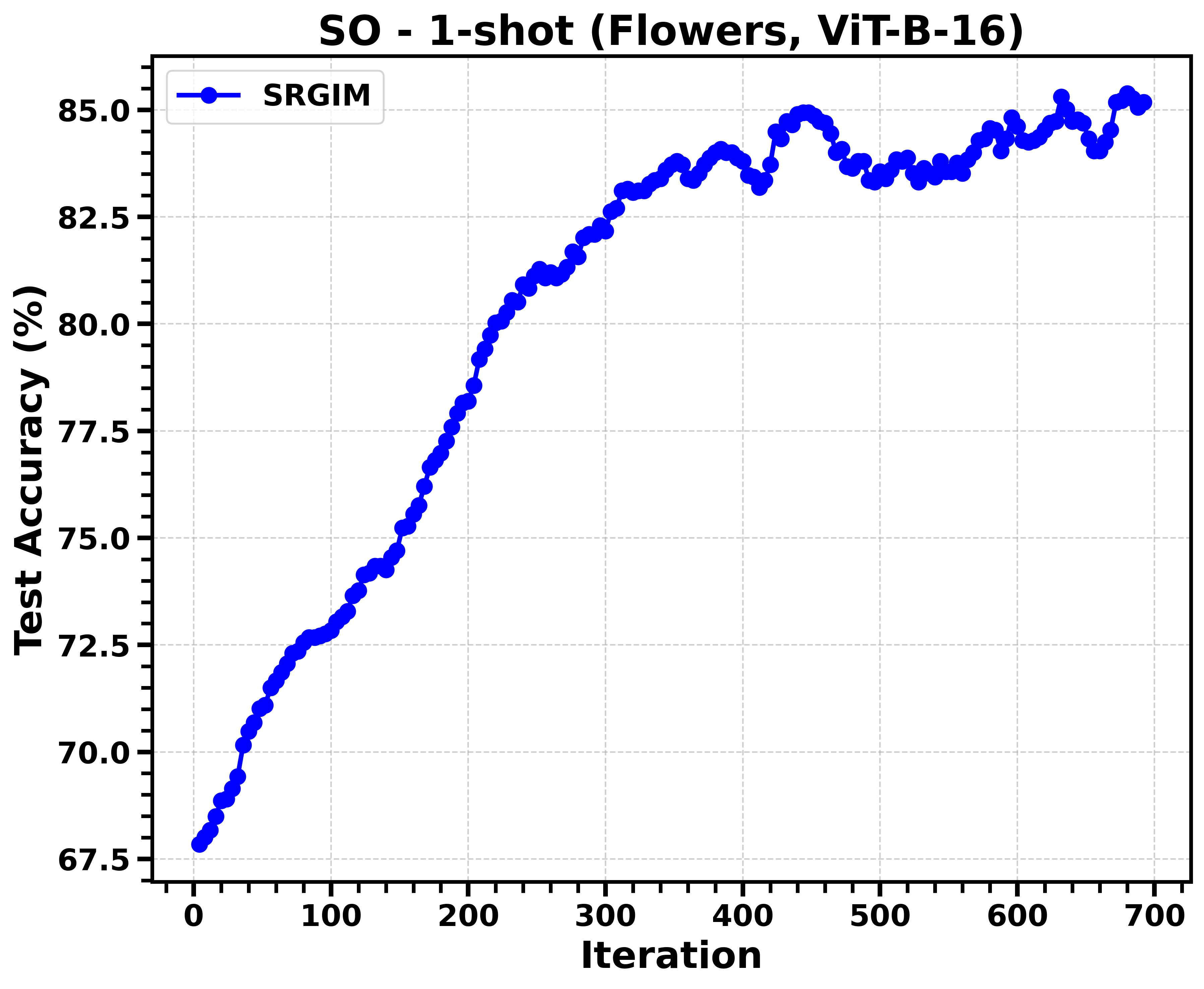}
    \end{subfigure} &
    \begin{subfigure}[b]{0.22\textwidth}
        \centering
        \includegraphics[width=\linewidth]{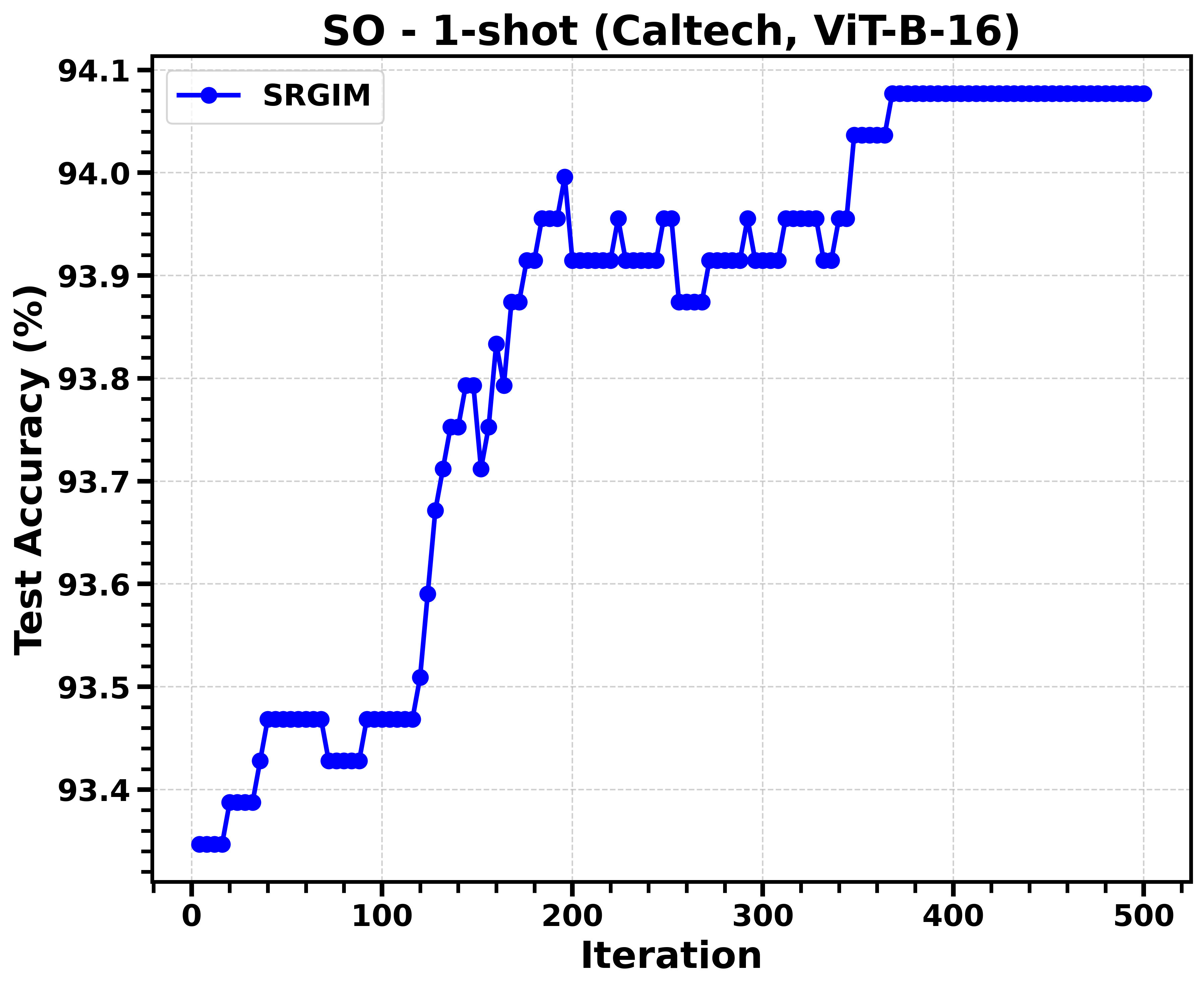}
    \end{subfigure} &
    \begin{subfigure}[b]{0.22\textwidth}
        \centering
        \includegraphics[width=\linewidth]{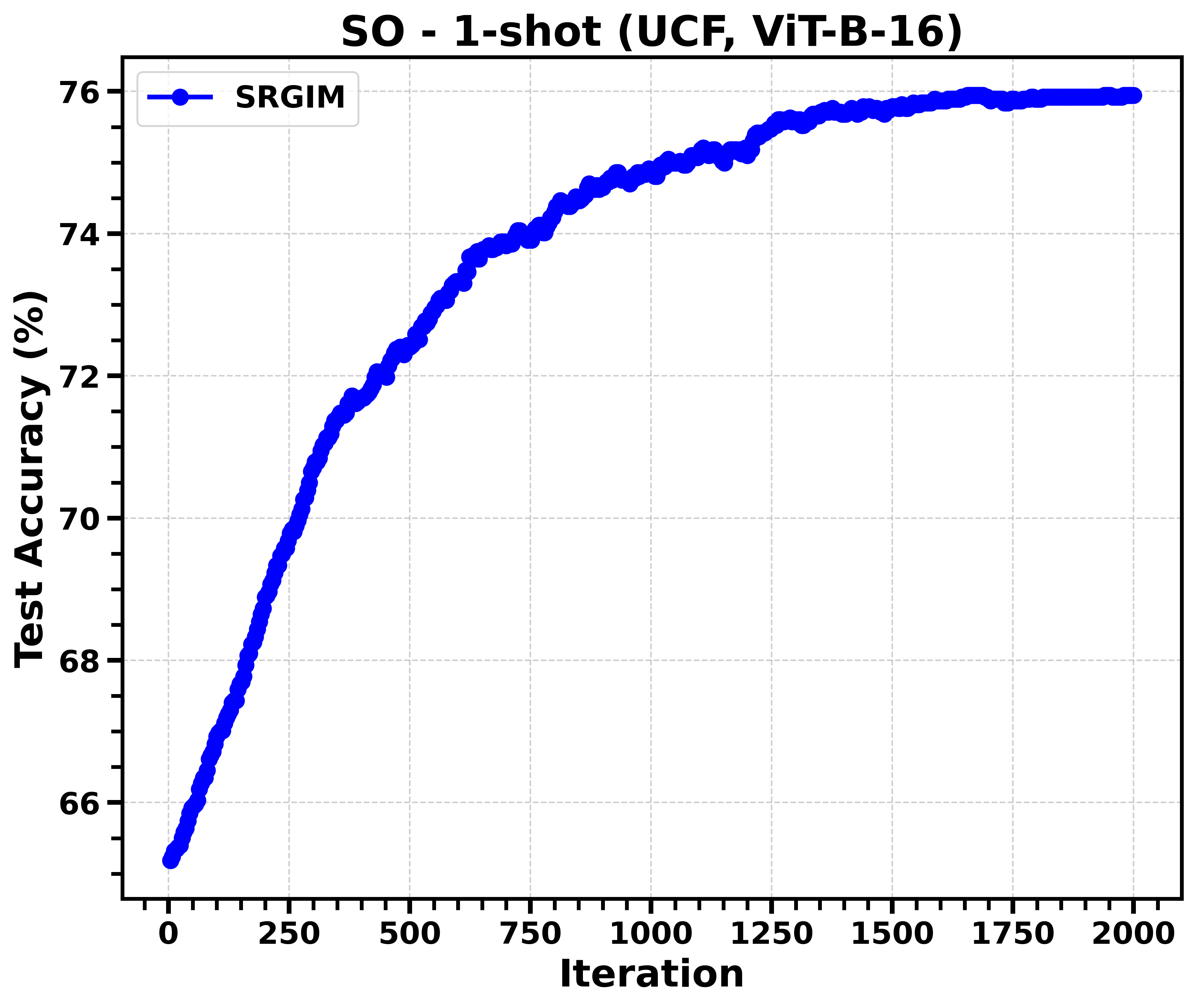}
    \end{subfigure} \\
\end{tabular}
\caption{Test accuracy over training iterations for LoRA (with rank fixed at 3) and SO (with a sparsity ratio equal to 99.95\%) in a 1-shot setting on four datasets (Pets, Flowers, Caltech, and UCF101) using a pretrained CLIP (ViT-B/16) backbone. Training stops when the loss falls below 0.01 or after 2000 iterations.}
\label{fig:matrix_of_figures_a}
\end{figure*}

\begin{figure*}
\centering
\renewcommand{\arraystretch}{1.3} 
\begin{tabular}{c@{\hskip 10pt} c c c c} 
    & Rank 2 & Rank 3 & Rank 4 & Rank 5 \\
    \datasetlabel{DTD} & 
    \begin{subfigure}[b]{0.22\textwidth}
        \centering
        \includegraphics[width=\linewidth]{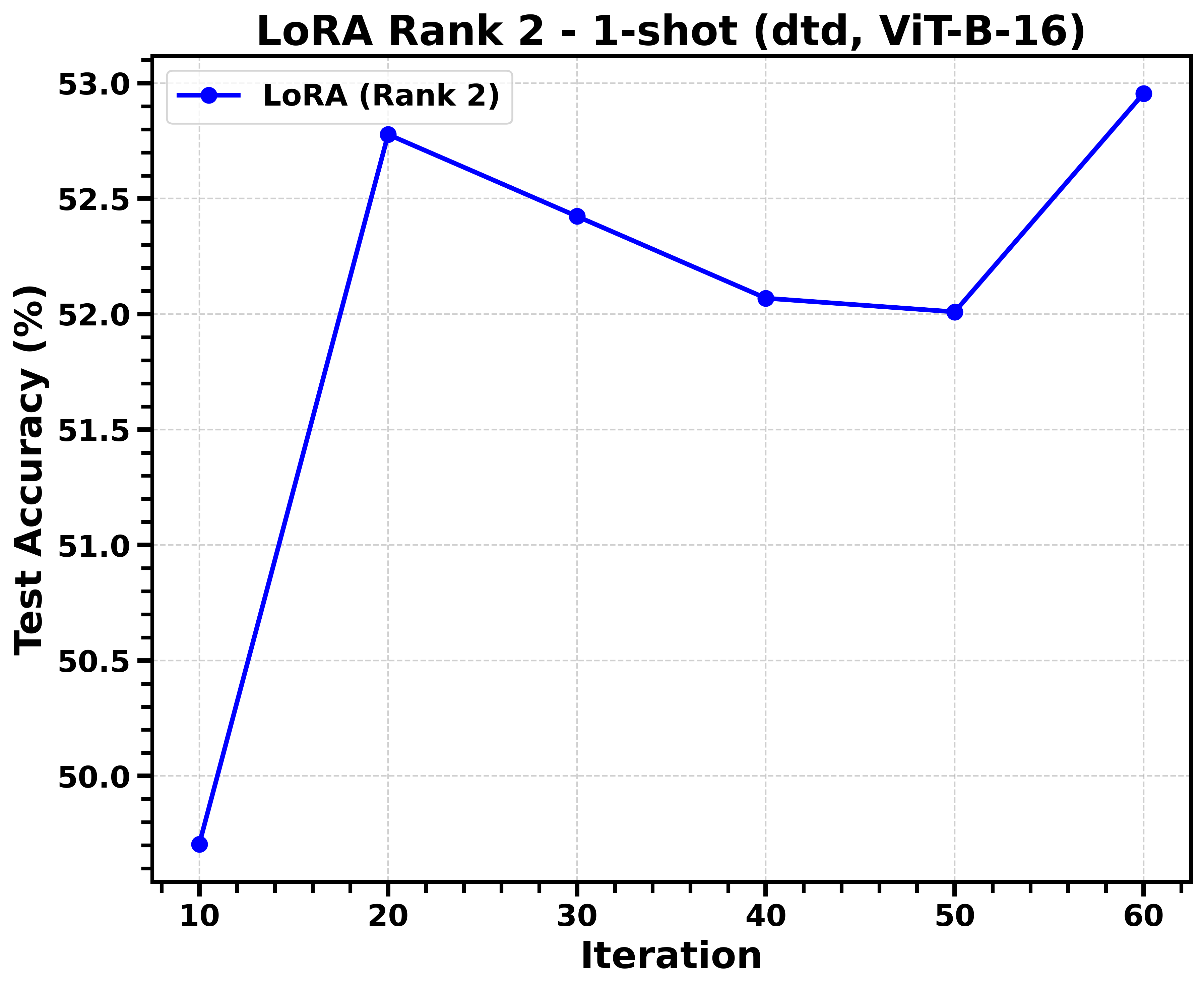}
    \end{subfigure} &
    \begin{subfigure}[b]{0.22\textwidth}
        \centering
        \includegraphics[width=\linewidth]{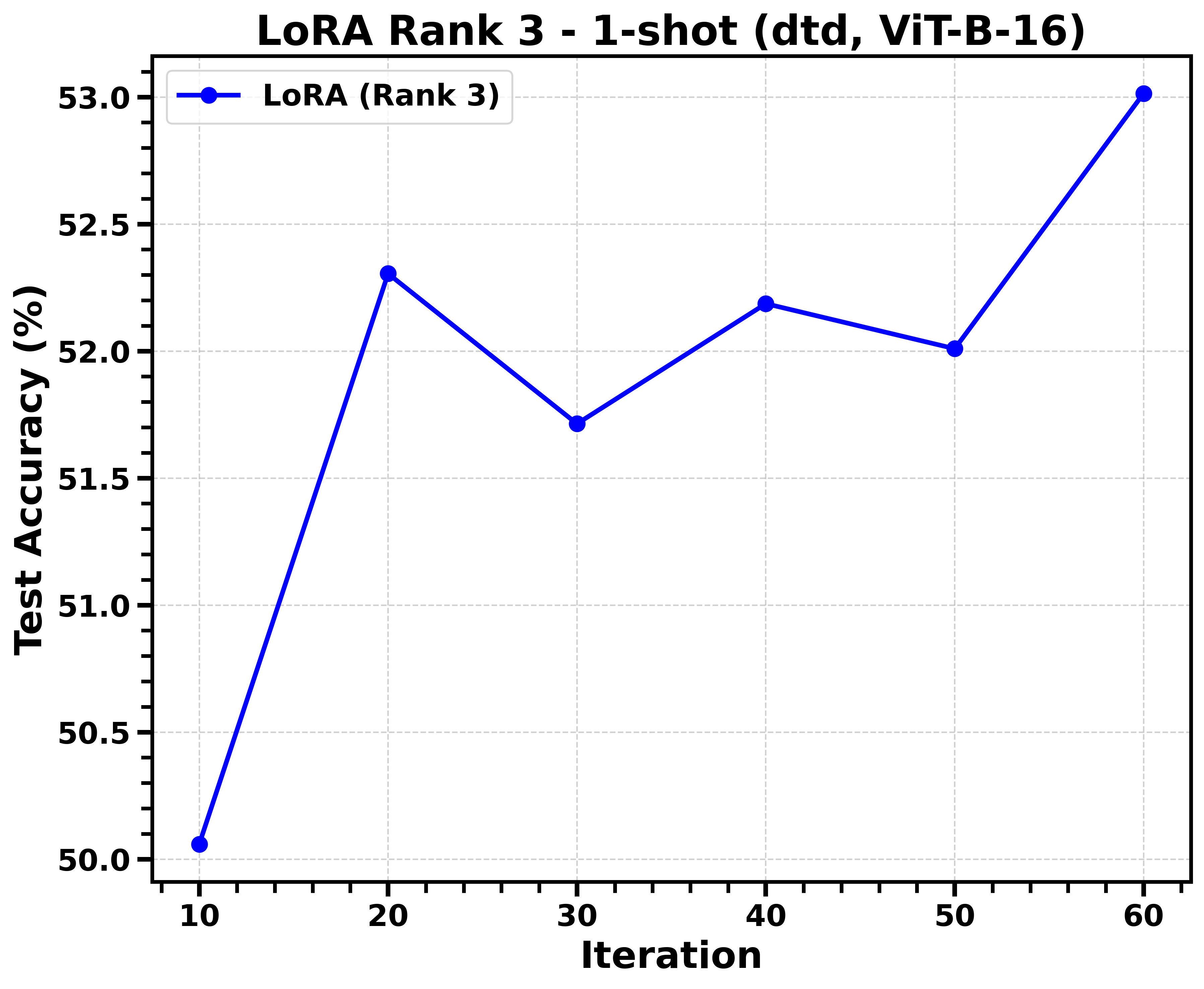}
    \end{subfigure} &
    \begin{subfigure}[b]{0.22\textwidth}
        \centering
        \includegraphics[width=\linewidth]{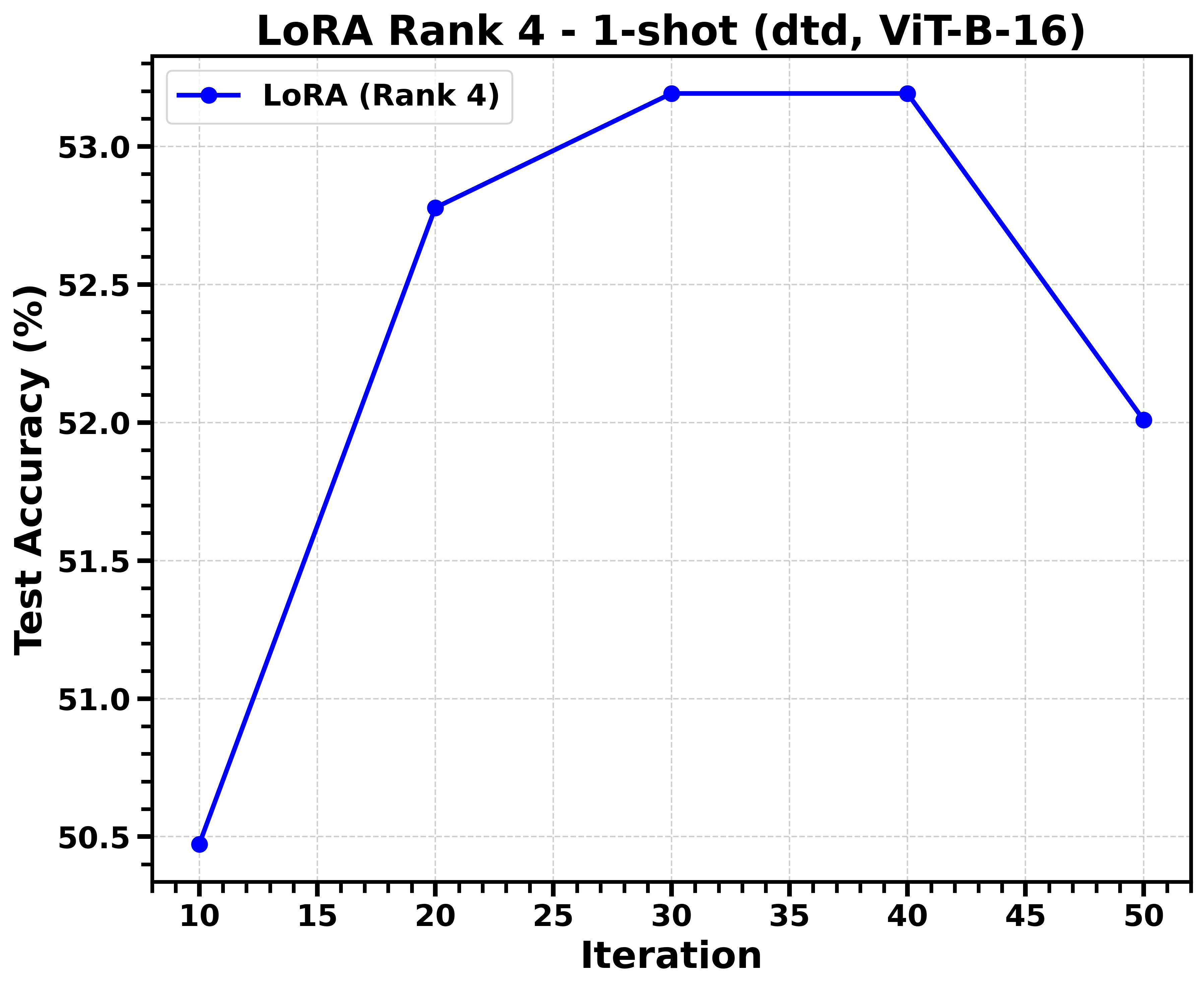}
    \end{subfigure} &
    \begin{subfigure}[b]{0.22\textwidth}
        \centering
        \includegraphics[width=\linewidth]{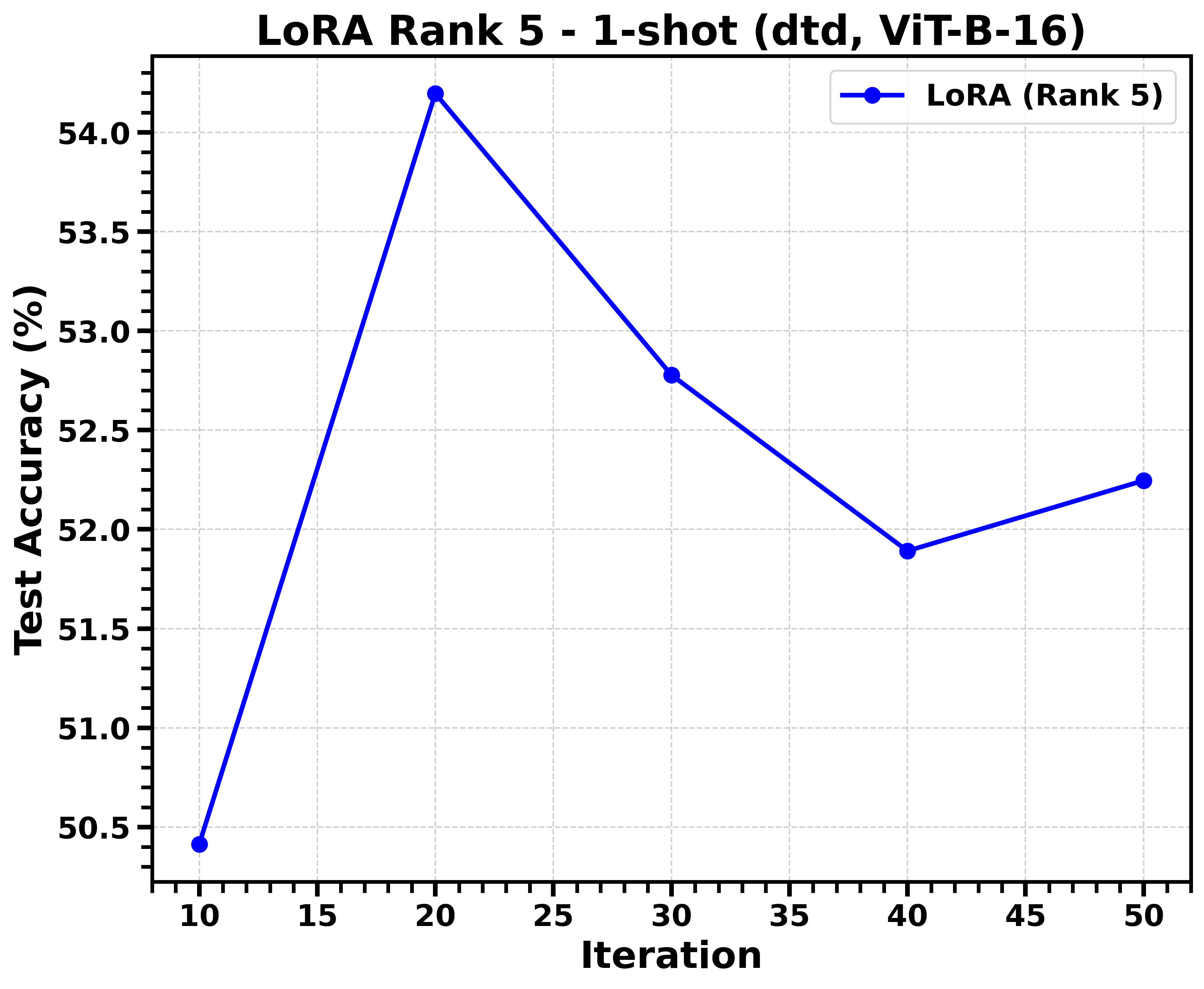}
    \end{subfigure} \\
    \datasetlabel{Pets} & 
    \begin{subfigure}[b]{0.22\textwidth}
        \centering
        \includegraphics[width=\linewidth]{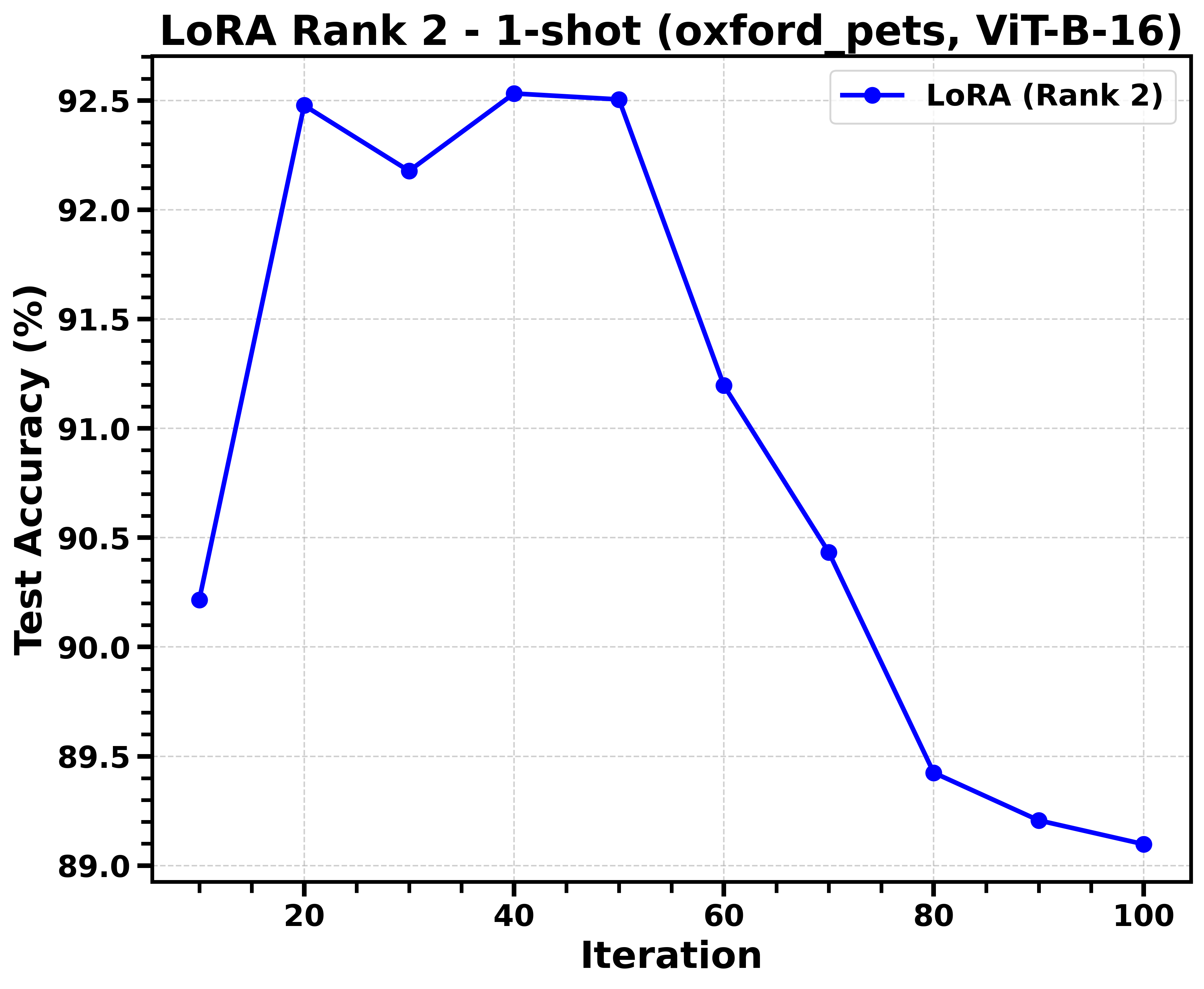}
    \end{subfigure} &
    \begin{subfigure}[b]{0.22\textwidth}
        \centering
        \includegraphics[width=\linewidth]{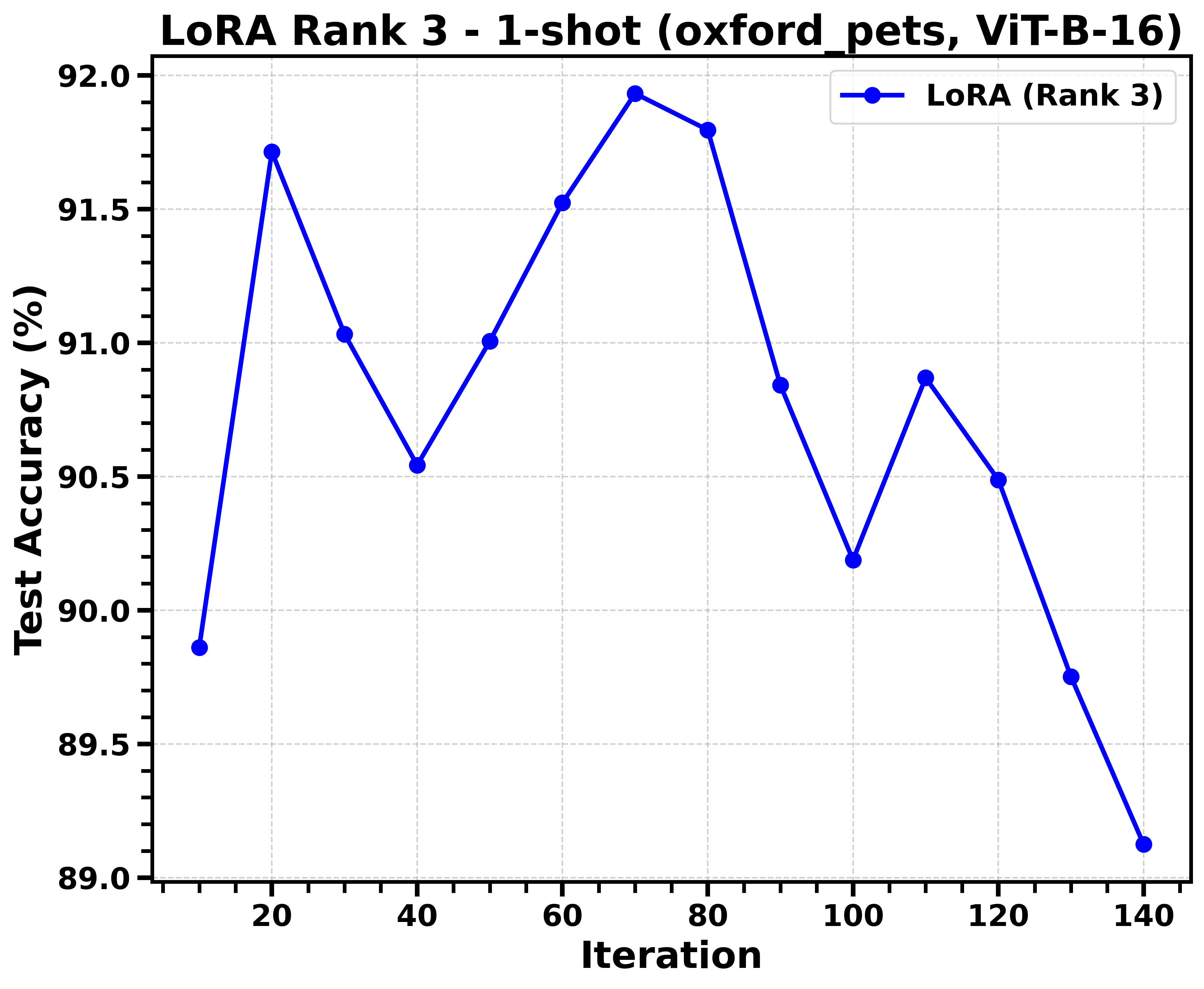}
    \end{subfigure} &
    \begin{subfigure}[b]{0.22\textwidth}
        \centering
        \includegraphics[width=\linewidth]{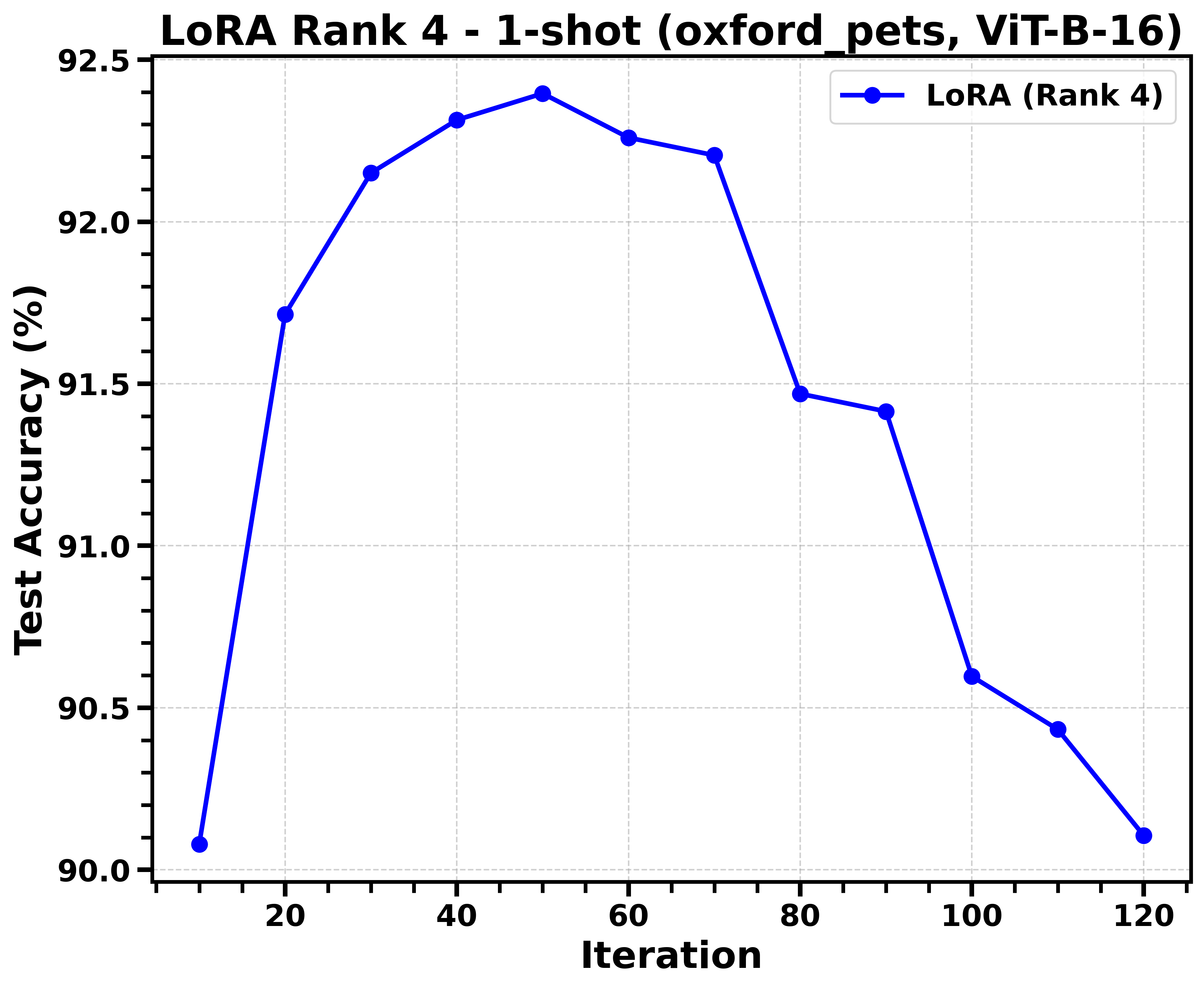}
    \end{subfigure} &
    \begin{subfigure}[b]{0.22\textwidth}
        \centering
        \includegraphics[width=\linewidth]{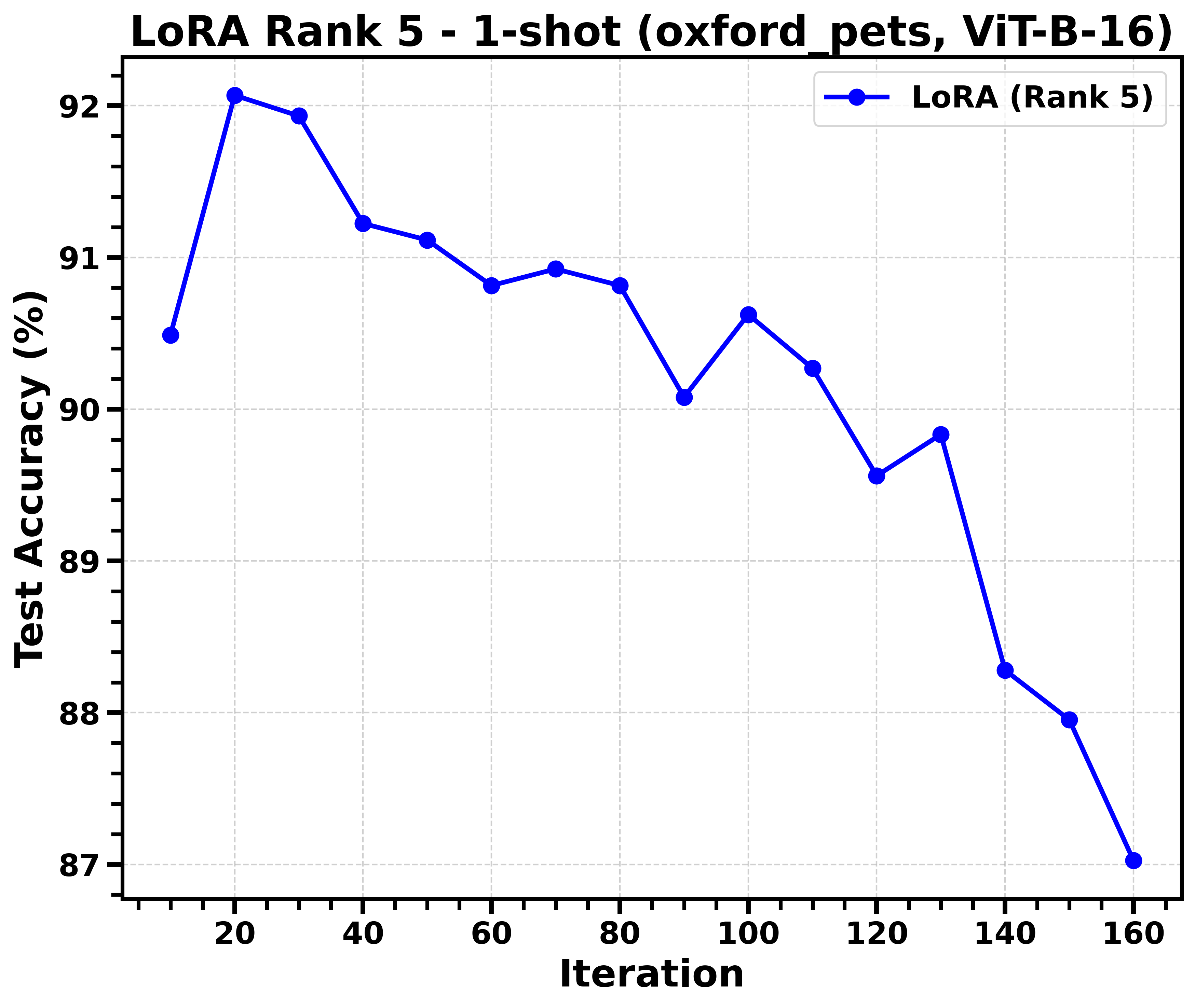}
    \end{subfigure} \\
    \datasetlabel{UCF101} & 
    \begin{subfigure}[b]{0.22\textwidth}
        \centering
        \includegraphics[width=\linewidth]{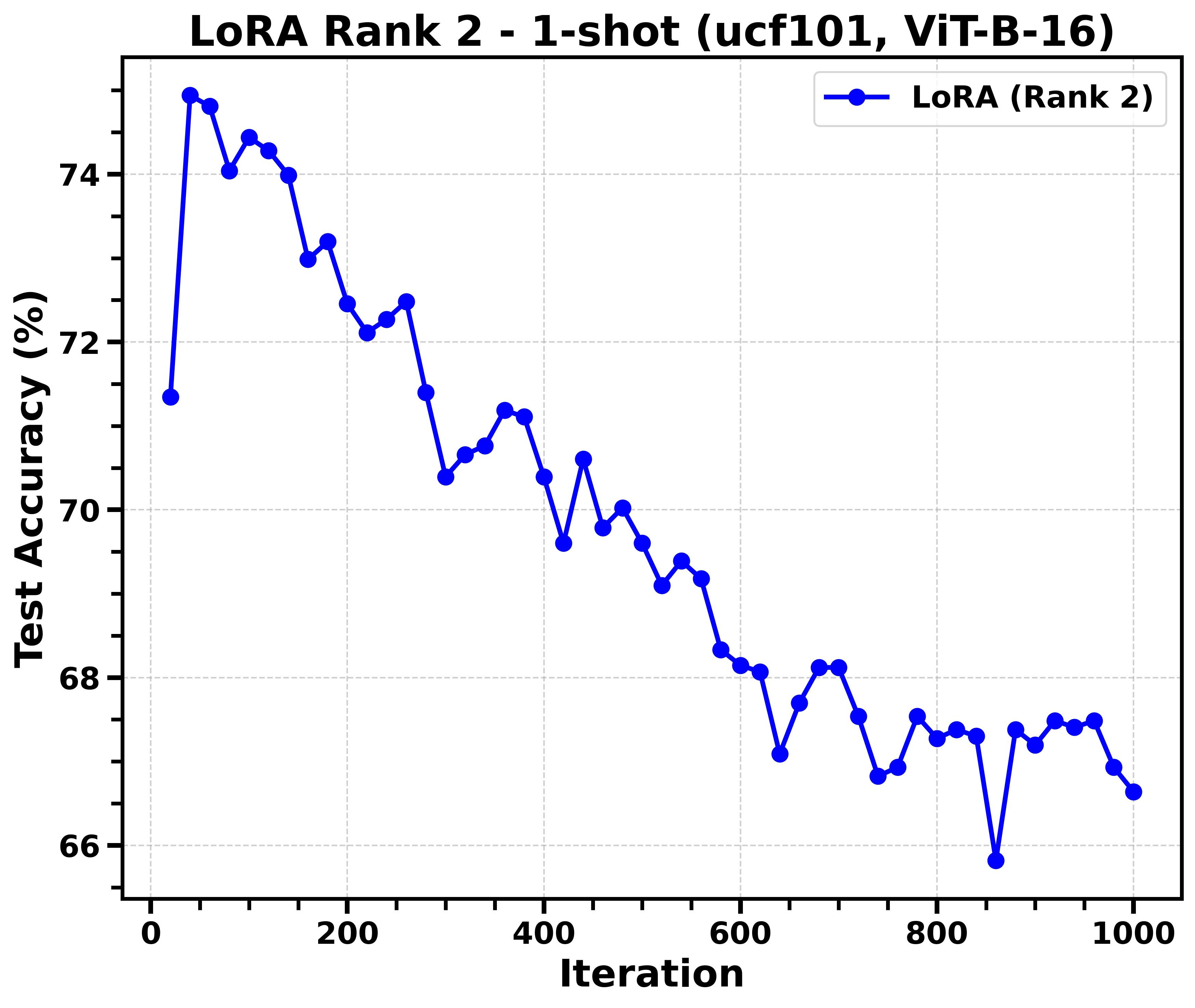}
    \end{subfigure} &
    \begin{subfigure}[b]{0.22\textwidth}
        \centering
        \includegraphics[width=\linewidth]{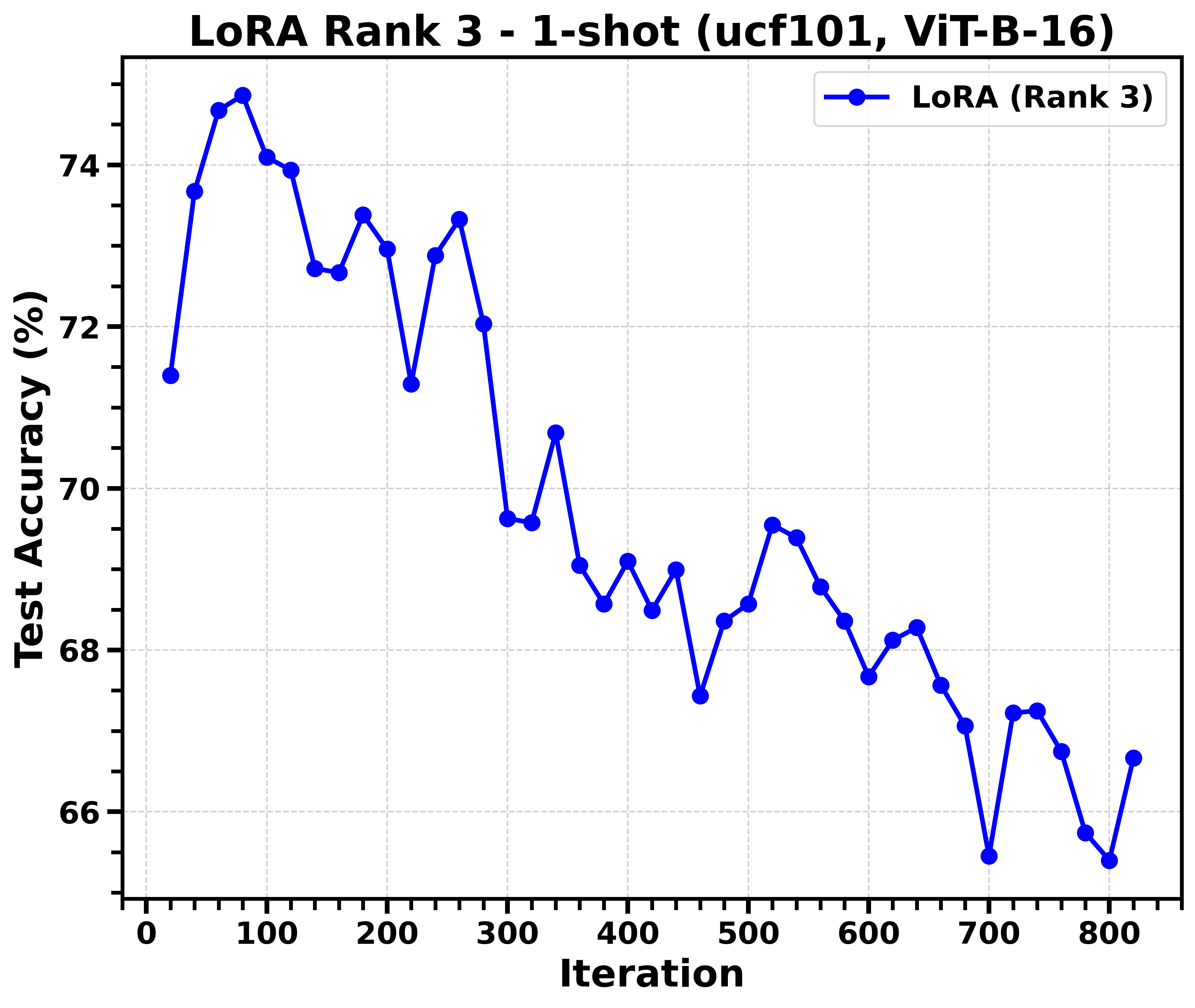}
    \end{subfigure} &
    \begin{subfigure}[b]{0.22\textwidth}
        \centering
        \includegraphics[width=\linewidth]{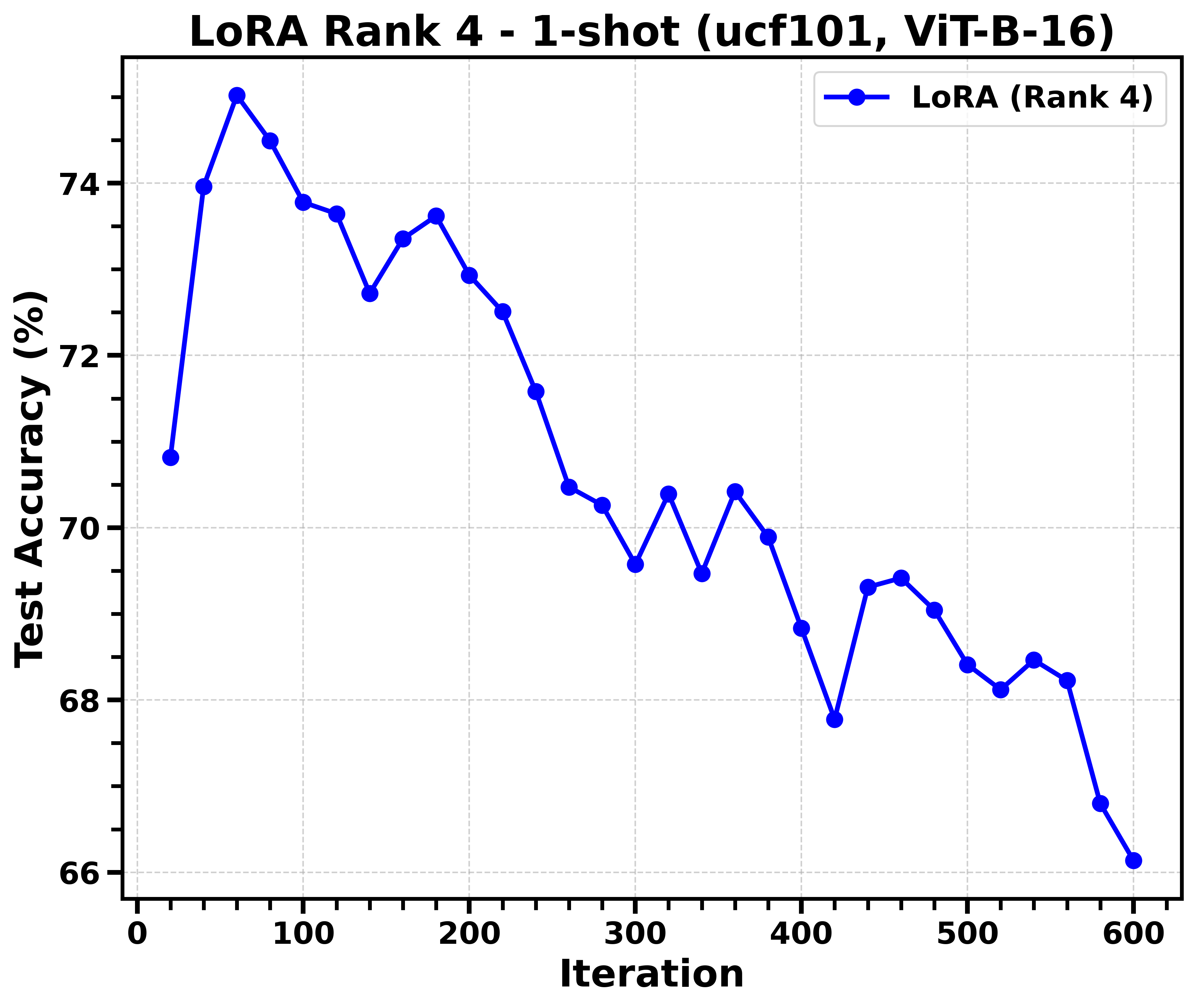}
    \end{subfigure} &
    \begin{subfigure}[b]{0.22\textwidth}
        \centering
        \includegraphics[width=\linewidth]{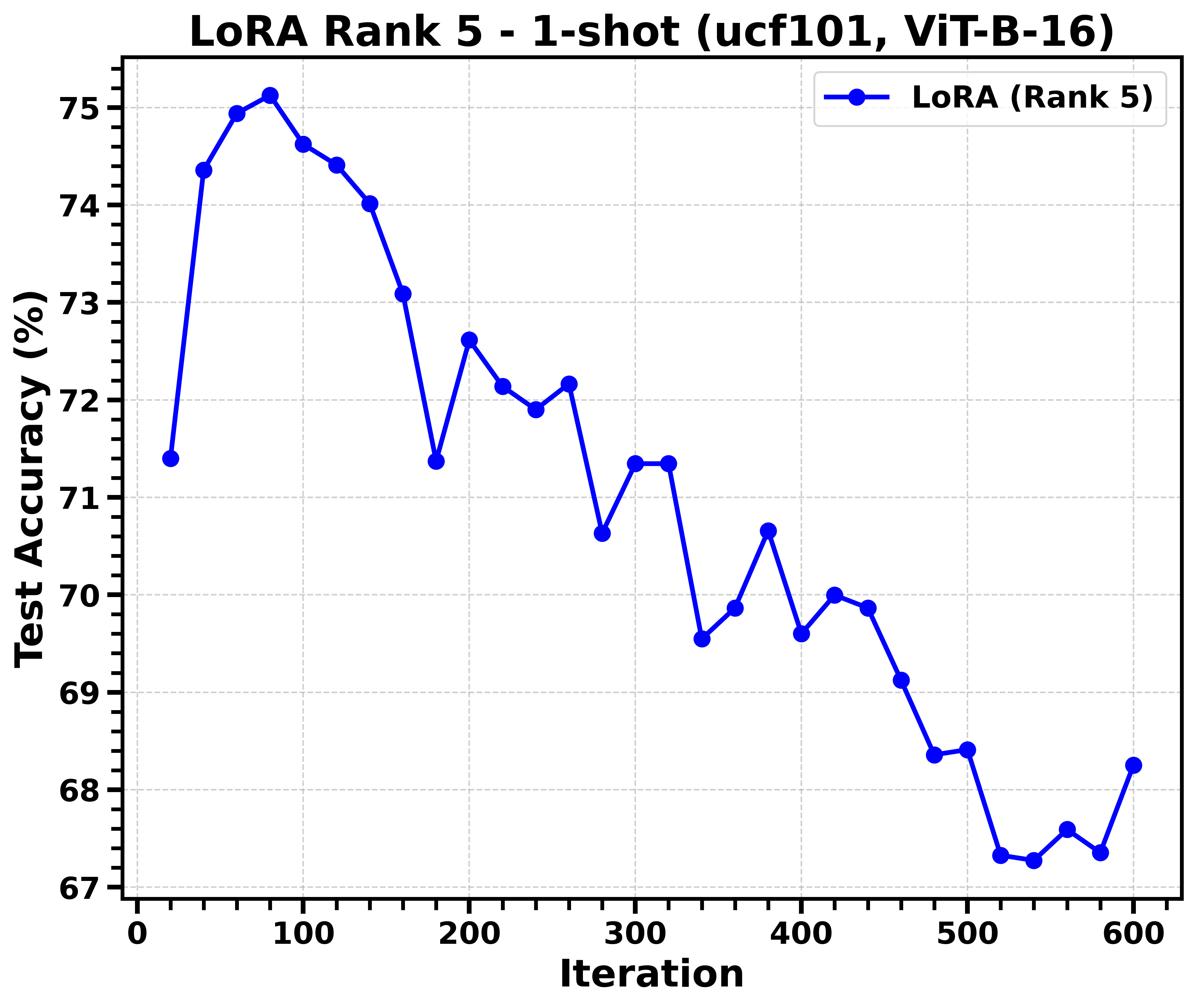}
    \end{subfigure} \\
\end{tabular}
\caption{LoRA performance in a 1-shot setting on three datasets (DTD, Oxford Pets, and UCF101), using a pretrained CLIP (ViT-B/16) backbone. We vary the LoRA rank from 2 to 5 and train for at most 2000 iterations or until the loss drops below 0.01. Each subfigure shows test accuracy over training iterations for a specific rank.}
\label{fig:matrix_of_figures_b}
\end{figure*}

The performance of deep learning (DL) models is strongly influenced by their size, with larger architectures yielding superior results. For instance, the ViT-H/14 backbone outperforms ViT-B/32 by a significant margin across multiple benchmarks \cite{cherti2023reproducible}. Despite recent architectural advancements, the best models remain highly data-dependent and require large-scale labeled datasets to achieve reliable generalization. While techniques such as data augmentation and regularization help mitigate overfitting in low-data scenarios, they are insufficient to fully resolve it \cite{balestriero2022effects}. In contrast, the human brain has more than $80$ billion neurons. Each neuron forms, on average, \textit{only} about $7,000$ synaptic connections, which results in an estimated total of $100$ to $500$ trillion synapses. Despite its tremendous capacity, the brain exhibits a remarkable ability to grasp new concepts from minimal exposure. This one-shot or few-shot learning capability might be partly attributed to the effective use of \textit{sparsity} \cite{jaaskelainen2022sparse}. By selectively activating a fraction of connections at any given time, the brain optimizes both energy consumption and computational efficiency.
 
Among the prominent DL models, vision-language models (VLMs) learn visual concepts from natural language descriptions. In particular, the contrastive language-image pretraining (CLIP) \cite{radford2021learning} model is pretrained on large-scale datasets containing hundreds of millions of image-text pairs to align visual and textual embeddings within a shared semantic space. After training, VLMs show strong zero-shot capabilities across diverse visual recognition tasks \cite{radford2021learning}. Moreover, fine-tuning VLMs yields state-of-the-art performance in various domain adaptation scenarios \cite{luo2023segclip, wu2023cora}. Despite their zero-shot capabilities, adapting VLMs to new tasks with small labeled datasets remains challenging due to pronounced overfitting and catastrophic forgetting problems. Furthermore, training such models requires vast resources, including significant GPU memory for storing millions of parameters, gradients, activations, and optimizer states. These challenges underscore the need for cost-effective optimization strategies that address overfitting.

To address the limitations of full VLM fine-tuning in few-shot learning, state-of-the-art methods leverage parameter-efficient training (PET) \cite{hu2022lora, houlsby2019parameter, kopiczko2024vera}. This strategy consists of optimizing a small number of parameters compared to the full set of weights. Additive PET methods freeze the initial weights and attach trainable and unmergeable parameters to the model. Within this category, we find the adapter \cite{gao2024clip} and the prompt tuning variants \cite{zhou2022learning}. The prevailing methods in few-shot VLMs have mainly focused on this category. Adjustable weights can be incorporated at various levels, either by embedding them directly into the input space as visual \cite{jia2022visual} or textual prompts \cite{zhou2022conditional}, or by integrating adapters throughout the network architecture \cite{zhu2023not}. 

To mitigate overfitting in few-shot adaptation and preserve the prior knowledge of VLMs, adapter-based methods typically employ three techniques. First, these small trainable modules are only integrated into the last layer of VLMs. Second, they use regularization techniques to penalize the divergence between the initial zero-shot class prototypes and the learnable class prototypes. Finally, they initialize the learnable class prototypes into a reliable region using the zero-shot prototypes. As a result, these methods have shown better results than the prompt tuning approaches \cite{yu2023task}, which often introduce computational overhead \cite{zanella2024low}. Despite these efforts, adapter-based methods are still prone to considerable overfitting \cite{silva2024closer}. Their success depends heavily on carefully tuning hyperparameters whose optimal values vary significantly across different tasks \cite{silva2024closer}. This sensitivity leads to performance degradation that undermines the generalizability of these approaches. It is worth noting that tuning these hyperparameters requires extensive grid searching with a validation set, which is not available in real-world few-shot scenarios.

Recently, low-rank reparametrization \cite{hu2022lora, liu2024dora, kopiczko2023vera} has gained popularity for domain adaptation. These approaches reparameterize the initial weights with supplementary trainable and mergeable low-rank matrices. Low-rank adaptation (LoRA) \cite{hu2022lora} was the first to apply this strategy. It reformulates weight matrices $\mathbf{W}$ as $\mathbf{W} + \mathbf{A} \cdot \mathbf{B}$, where $\mathbf{A}$ and $\mathbf{B}$ are low-rank matrices. As a key advantage, this technique helps reduce overfitting and catastrophic forgetting because of two properties. First, LoRA enables fine-grained parallel adaptation by learning small perturbations to the frozen weights \cite{he2021towards}. Second, the effective initialization of \(\mathbf{A}\) and \(\mathbf{B}\) helps in preserving prior knowledge \cite{meng2024pissa, yang2024corda}.

Building on these advantages, recent studies \cite{zanella2024low, lin2024nora} show that low-rank reparameterization methods achieve state-of-the-art performance in few-shot adaptation of VLMs. However, these methods also have notable limitations. Compelling evidence suggests that restricting weight updates to low-rank subspaces reduces model expressiveness \cite{zhao2024galore, lialin2023relora, sun2023comparative}. In this work, we argue that the low-rank projection technique exhibits severe overfitting in few-shot adaptation. Our systematic study of LoRA confirms this issue (see Fig. \ref{fig:matrix_of_figures_a}). Test-set accuracy initially improves but then declines sharply. At convergence, when the loss approaches zero, performance deteriorates significantly compared to its peak on test-set data. Moreover, we can see that the optimal number of training iterations varies widely across datasets. However, tuning hyperparameters, including the number of iterations for early stopping, on a validation set contradicts the principles of few-shot learning.

The challenges of low-rank reparameterization methods extend beyond fixing the number of training iterations. Their performance is also highly sensitive to another hyperparameter: the rank of the low-rank projection. This sensitivity is shown in Fig. \ref{fig:matrix_of_figures_b}, which illustrates LoRA’s performance for different ranks on three datasets (DTD, Oxford Pets, UCF101) in a 1-shot setting. Each dataset reveals a distinct best rank, and higher ranks sometimes lead to an early spike in accuracy but then degrade as training continues. Further experiments examining these behaviors are presented in Appendix~F of the supplementary material.

Since rank is a discrete value that cannot be lower than 1, the low-rank methods lack the flexibility to adjust very few parameters. While setting the rank to a very low value (e.g., 1 or 2) can help mitigate overfitting, this choice might reduce the model's learning capacity. This limitation arises because the trainable parameters in low-rank methods remain fixed throughout training, potentially leading to underfitting the VLM capacity. Indeed, rank selection involves a trade-off between underfitting and overfitting (see Fig. \ref{fig:matrix_of_figures_b}). For instance, in the 1-shot DTD setting, LoRA at rank 5 briefly exceeds rank 2 but then drops below it as training continues. Overall, the test accuracy of LoRA exhibits oscillations and varies unpredictably w.r.t. rank and the number of training iterations. This instability makes it difficult to reliably tune these interdependent hyperparameters without access to a validation set.

In this paper, we deviate from the current long-standing trend of low-rank adaptation and introduce two new paradigms that make sparsity work in few-shot settings. Our approach is a simple yet effective optimization strategy, referred to as sparse optimization (SO), and aligns with these paradigms. As a first paradigm, we advocate for \textit{local sparsity and global density}. Unlike the conventional low-rank approaches that update a fixed set of parameters, SO dynamically updates a minimal subset of original connections at each step. Specifically, high sparsity is enforced in both the gradient and moment updates while allowing the sparsity support to evolve dynamically throughout training. This paradigm ensures that, at any given iteration, only a minimal subset of parameters is updated to reduce the local learning capacity and prevent rapid overfitting. By dynamically changing the sparsity support, the model retains its overall expressiveness and ensures that different parameters contribute to learning over time.

As a second paradigm, we advocate for \textit{local randomness and global importance}. Similar to Adam \cite{kingma2014adam}, we leverage the gradient and the first and second moments. However, we sparsify these components in a way that addresses overfitting and ensures efficiency in terms of memory consumption. In particular, we sparsify the gradient randomly and the first moment by importance. The gradient captures local and iteration-specific information. The first moment aggregates the gradients over the whole path and, thus, reflects long-term parameter importance. Random pruning of the gradient prevents the model from relying too much on short-term and local high-magnitude updates. The importance-based selection of the first moment ensures that connections with long-term significance are updated. By emphasizing local randomness and global importance, our strategy avoids severe overfitting.
 
Concretely instantiating this paradigm, we fix a density rate \(\kappa\) and retain \(M\) elements for both the gradient and moment updates at each iteration. First, we randomly select \(M\) gradient values from the full set of parameter updates. Next, we compute the temporary first and second moments, which can contain up to \(2M\) values in the worst-case scenario where no overlap exists between the selected gradients and the previous \(M\) first moment values. We rank the \(2M\) first moment values by importance using their magnitude. Only the top \(M\) values are retained for the first moment. We then use the same indices to select the corresponding values for the second moment. This ensures alignment between the sparsified first and second moments. The selection strategy preserves the most relevant long-term updates while introducing randomness in the local updates.

\textbf{Our contributions.} 
\noindent \textbf{(1)} A novel framework is proposed that sparsifies both gradients and moments to improve generalization and mitigate overfitting in few-shot learning. Our approach differs from conventional low-rank reparameterization by focusing on sparsity.  
\noindent \textbf{(2)} We introduce the local sparsity and global density paradigm, where a dynamically evolving sparsity support updates only a minimal subset of model parameters at each iteration. This strategy preserves the overall model expressiveness while reducing its risk of overfitting. 
\noindent \textbf{(3)} We propose the local randomness and global importance paradigm, which selectively sparsifies the gradient using random selection while pruning the first moment based on importance. This selection strategy preserves the most relevant long-term updates while introducing local randomness to mitigate overfitting. 
\noindent \textbf{(4)} Our comprehensive empirical study on 11 diverse datasets compares our method to state-of-the-art low-rank approaches. Results indicate that SO can significantly improve few-shot generalization while achieving memory efficiency.
\section{Related Work}
\label{sec:related_work}

\noindent \textbf{(a) Low-Rank Methods.} 
LoRA \cite{hu2022lora} approximates weight updates by decomposing them into two low-rank matrices. As a result, it reduces the number of trainable parameters while preserving the model's frozen pretrained backbone. DoRA \cite{liu2024dora} introduces a weight decomposition approach to improve LoRA's expressiveness by decoupling weight magnitude and direction and training them separately. VeRA \cite{kopiczko2024vera} reduces the set of trainable parameters further than LoRA. In particular, VeRA freezes the pair of low-rank matrices to randomly initialized ones and only trains two small scaling vectors. PiSSA \cite{meng2024pissa} initializes the two low-rank matrices using the most informative singular vectors and their corresponding singular values from the original weight matrix while freezing the less significant components in a residual matrix. ReLoRA \cite{lialin2023relora} achieves high-rank weight updates by merging trained low-rank updates into the main model parameters at regular intervals and re-initializing the auxiliary low-rank matrices. Unlike these LoRA-based reparametrization methods, which constrain weight updates to low-rank subspaces, GaLore \cite{zhao2024galore} performs full-parameter training. Specifically, GaLore projects gradients into a low-rank subspace to reduce memory consumption while allowing full updates to the model weights.

Despite their efficiency, these methods still face major limitations. Most low-rank techniques impose a fixed-rank constraint on the updates, which restricts adaptation flexibility and may lead to underfitting when the rank is too low. Even with minimal rank, these methods struggle with generalization in few-shot adaptation due to severe overfitting. Furthermore, these approaches require careful hyperparameter tuning, particularly in selecting the optimal rank and training iterations, which significantly impact performance. Our work challenges this strategy by showing that appropriately designed sparsity-based adaptation provides superior generalization while reducing memory consumption.


\noindent \textbf{(b) Sparsity-Based Methods.} These methods update only a subset of model parameters. The selection can be structured or unstructured. Structured selection methods \cite{ma2023llmpruner, lagunas2021block, xia2022structured, chen2023lorashear, dery2024everybody} target entire modules (e.g., attention heads, layers), offering coarse-grained sparsity but often requiring architecture-specific designs that may degrade performance \cite{Cheng2024survey}. Unstructured selection methods \cite{zhang2024gradient, sun2024a, han2024sltrain, ben2022bitfit, sung2021training, xu2021raise} prune individual weights for greater flexibility. However, most unstructured methods rely on static sparsity patterns and perform importance-based gradient pruning. Moreover, none of these methods incorporate sparse moments.

In contrast, SO updates a minimal subset of parameters at each step. Furthermore, our approach uses random gradient selection to avoid local high-magnitude biases and importance-based moment pruning to retain long-term relevance. This combination enables stable adaptation in few-shot scenarios without sacrificing model expressiveness.
\section{Proposed Sparse Optimization Method}
\label{sec:proposed_method}

A comprehensive overview of few-shot adaptation in VLMs is provided in Appendix~A. Briefly, we leverage a pretrained CLIP \cite{radford2021learning} model that aligns images and texts in a shared embedding space via contrastive learning. After pretraining, CLIP can perform zero-shot classification by comparing image embeddings to class-specific text prototypes. In the few-shot scenario, we only have a small support set of labeled examples (e.g., \(K\) images per class). We adapt CLIP to this support set by minimizing the cross-entropy loss with respect to the class prototypes or learnable parameters, and we optimize the loss function using our proposed approach.

Motivated by the challenges inherent in few-shot adaptation, we introduce SO, a new optimizer designed to enhance memory efficiency while mitigating overfitting. Unlike standard optimization approaches that update all parameters at each iteration, SO dynamically updates a subset of parameters. This sparsification significantly reduces memory consumption, which makes SO suitable for training large-scale deep learning models like VLMs.

Our approach builds on the Adam optimizer and behaves identically to it when the sparsity ratio is set to zero. However, SO deviates from Adam by following two paradigms. The first paradigm is \textit{local sparsity and global density}, which ensures that a minimal subset of parameters is dynamically updated while maintaining overall model expressiveness. The second paradigm is \textit{local randomness and global importance}, which applies random pruning to the gradient updates while retaining the most significant moment estimates. These principles prevent severe overfitting and make SO suitable for scenarios like few-shot learning.

\noindent \textbf{Notation.} Let \( \Theta \in \mathbb{R}^{d} \) represent the model parameters, where \( d \) is the total number of parameters. At each iteration \( t \), the model updates some parameters to minimize a given loss function \( \mathcal{L}(\Theta_t) \). The gradient of the loss concerning all parameters is denoted as \( g_t = \nabla_{\Theta} \mathcal{L}(\Theta_t) \). We use the first and second moments, denoted as \( \mu_t \) and \( \nu_t \), respectively, to capture long-term and global parameter significance throughout the optimization process. A small constant \( \epsilon \) is introduced for numerical stability. The learning rate, denoted as \( \eta \), scales the parameter updates. We introduce a density ratio \( \kappa \in [0,1] \), which determines the fraction of gradient entries retained at each step. Let \( \mathcal{I}(\cdot) \) and \( \mathcal{V}(\cdot) \) be the functions that return the indices and values of a sparse input vector, respectively. Instead of updating all parameters, SO retains only a small subset of the gradient and generates sparse first and second moments. 

We enforce sparsity through two operations. The $\text{Top-}M$ operation selects $M$ elements with the highest magnitude. In contrast, $\text{Random-}M$ retains $M$ elements uniformly at random, preventing bias toward high-magnitude values.

\noindent \textbf{Dynamic and High-Sparsity Support.} Conventional optimization methods update the full set of parameters throughout training, which accelerates overfitting. In contrast, our method dynamically updates a minimal subset of parameters. 
The high sparsity and dynamic selection provide two key advantages. First, updating a minimal subset of parameters reduces the local learning capacity and thus mitigates overfitting and catastrophic forgetting. Second, continuously changing the sparsity support ensures that different parameters receive updates throughout training. This avoids excessive reliance on a fixed set and makes the learning process benefit from the model’s overall expressiveness. 

\noindent \textbf{Randomness in Gradient Pruning.} We sparsify the gradient vector to mitigate overfitting. The proposed optimizer applies a stochastic sampling mechanism that uniformly picks \( M \) elements from the gradient \( g_t \). Given the density ratio \( \kappa \), we define the number of retained gradient entries as $M = \lfloor \kappa d \rfloor$. 
Formally, we have $\tilde{g}_t = \text{Random-}M\big(g_t\big)$.

The motivation is to avoid biasing updates toward high-magnitude gradients. Enforcing stochastic sparsity prevents the model from over-relying on seemingly important local information. Our empirical results indicate that pruning the gradient by retaining only the largest values accelerates overfitting and significantly degrades performance.

\noindent \textbf{Importance in Moment Pruning.} To maintain efficiency, our goal is to construct sparse first and second moments, $\tilde{\mu}_t$ and $\tilde{\nu}_t$, each of size $M$. Since the sparse gradient changes its support dynamically, storing historical information becomes challenging. Without pruning, the first and second moments can become dense over time due to the accumulation of gradients with varying sparsity support. 

To address this issue, we introduce two fixed-size buffers that can store up to $2M$ values. This ensures that both new and past gradient values are considered. Let \(\mu_t\) and \(\nu_t\) be the temporary first and second moments, at iteration \(t\): 
\begin{align}
    \mu_t &= \beta_1 \, \tilde{\mu}_{t-1} + (1 - \beta_1) \, \tilde{g}_t, \\
    \nu_t &= \beta_2 \, \tilde{\nu}_{t-1} + (1 - \beta_2) \, \tilde{g}_t^2.
\end{align}

These moments may contain up to \(2M\) values in the worst-case scenario when there is no overlap between the newly selected sparse gradient $\tilde{g}_t$ and the previously retained sparse first moment $\tilde{\mu}_t$. To ensure stable updates, we employ a Top-\(M\) selection strategy for the first moment, where we rank the $2M$ values by magnitude and retain only the top \(M\) ones. This guarantees that long-term significant parameters are consistently updated. For the second moment, we enforce alignment by retaining elements corresponding to the same indices selected for the first moment. The sparse first and second moments are expressed as:
\begin{equation}
    \tilde{\mu}_{t} = \text{Top-}M\big(\mu_t\big), \:\:\:\:\: \tilde{\nu}_{t} = \nu_t[\mathcal{I}(\tilde{\mu}_{t})].
\end{equation}

Both sparse moments $\tilde{\mu}$ and $\tilde{\nu}$ are initialized at zero. Similar to Adam, we apply bias correction to $\tilde{\mu}_t$ and $\tilde{\nu}_t$ before parameter update. The corrected sparse first and second moments, denoted by \(\hat{\mu}_t\) and \(\hat{\nu}_t\), are expressed as follows:
\begin{equation}
    \hat{\mu}_t = \frac{\tilde{\mu}_t}{1 - \beta_1^t}, \:\:\:\:\: \hat{\nu}_t = \frac{\tilde{\nu}_t}{1 - \beta_2^t}.
\end{equation}

Finally, the parameters $\Theta$ are updated using the corrected sparse first and second moments, $\hat{\mu}_t$ and $\hat{\nu}_t$, as provided in:
\begin{equation}
    \Theta_{t+1} = \Theta_{t} - \frac{\eta}{\sqrt{\hat{\nu}_t} + \epsilon} \hat{\mu}_t.
\end{equation}
where $\eta$ is the learning rate, and $\beta_{1}$ and $\beta_{2}$ are decay rates.

\begin{table*}
\caption{Few-shot classification performance on 11 datasets with ViT-B/16 backbone. Top-1 accuracy averaged over 3 random seeds is reported. Highest value is highlighted in \textbf{bold}, and the second highest is \underline{underlined}.}
\label{tab:vitb16}
\centering
\resizebox{0.97\textwidth}{!}{
\begin{tabular}{llcccccccccccc}
\toprule
Shots & Method & ImageNet & SUN & Aircraft & EuroSAT & Cars & Food & Pets &  Flowers & Caltech & DTD & UCF & Average
\\ \midrule 
\multirow{1}{*}{0} & CLIP {\tiny \textbf{(ICML '21)}} & 66.7 & 62.6 & 24.7 & 47.5 & 65.3 & 86.1 & 89.1 & 71.4 & 92.9 & 43.6 & 66.7 & 65.1  \\

\midrule
\multirow{8}{*}{1} 
& Adam \scriptsize{(ICLR '15)} & 0.1 & 0.6 & 1.0 & 24.3 & 0.5 & 1.0 & 2.9 & 1.4 & 1.0 & 5.6 & 1.7 & 3.6 \\
& LoRA \scriptsize{(ICLR '22)} & 67.3 & 67.0 & 25.0 & 67.5 & 68.2 & 81.2 & 90.5 & \underline{85.7} & 92.3 & 52.4 & 72.9 & 70.0 \\
& ReLoRA \scriptsize{(ICLR '24)} & \textbf{70.3} & \underline{69.7} & \underline{28.8} & \underline{73.8} & \underline{70.5} & \underline{84.2} & \underline{91.9} & 85.1 & 93.4 & \underline{54.2} & \underline{76.0} & \underline{72.5} \\
& GaLoRE \scriptsize{(ICML '24)} & 65.1 & 65.5 & 21.1 & 65.2 & 63.6 & 74.1 & 85.2 & 80.5 & 91.4 & 48.9 & 69.9 & 66.4 \\
& PiSSA \scriptsize{(NeurIPS '24)} & 64.9 & 65.2 & 22.2 & 66.8 & 64.2 & 75.7 & 84.8 & 82.9 & 90.7 & 50.5 & 70.1 & 67.1 \\
& DoRA \scriptsize{(ICML '24)} & 66.9 & 66.9 & 25.3 & 71.4 & 68.3 & 81.3 & 89.3 & \textbf{86.2} & 92.3 & 51.9 & 72.9 & 70.2 \\
& VeRA \scriptsize{(ICLR '24)} & 68.4 & 67.3 & 27.5 & 69.3 & 67.4 & \textbf{86.2} & 91.0 & 73.1 & \underline{93.9} & 52.2 & 70.1 & 69.7 \\
\rowcolor{LightGray} & SO \scriptsize{(Ours)} & \underline{70.1} & \textbf{70.3} & \textbf{31.5} & \textbf{78.2} & \textbf{71.6} & \textbf{86.2} & \textbf{93.3} & 84.9 & \textbf{94.1} & \textbf{55.3} & \textbf{76.4} & \textbf{73.8} \\

\midrule
\multirow{8}{*}{2}
& Adam \scriptsize{(ICLR '15)} & 0.2 & 0.8 & 0.9 & 42.1 & 0.7 & 1.9 & 3.9 & 17.1 & 3.1 & 7.2 & 3.8 & 7.4 \\
& LoRA \scriptsize{(ICLR '22)} & 67.4 & 68.3 & 30.3 & 81.9 & 70.1 & 79.3 & 89.4 & 90.9 & 93.7 & 59.6 & 76.1 & 73.4 \\
& ReLoRA \scriptsize{(ICLR '24)} & \textbf{70.6} & \underline{70.5} & \underline{32.2} & \textbf{83.8} & \underline{72.9} & 82.6 & 90.3 & 90.7 & \underline{94.7} & \textbf{60.4} & \underline{79.1} & \underline{75.3} \\
& GaLoRE \scriptsize{(ICML '24)} & 66.1 & 67.0 & 25.8 & 82.2 & 67.2 & 72.6 & 81.5 & 85.6 & 91.5 & 56.1 & 73.4 & 69.9 \\
& PiSSA \scriptsize{(NeurIPS '24)} & 65.1 & 66.7 & 26.1 & 77.3 & 67.1 & 74.8 & 84.4 & 87.4 & 91.9 & 57.4 & 73.3 & 70.1 \\
& DoRA \scriptsize{(ICML '24)}  & 67.2 & 68.2 & 29.8 & \underline{83.4} & 70.3 & 79.4 & 89.6 & \underline{91.2} & 93.6 & 59.7 & 75.9 & 73.5 \\
& VeRA \scriptsize{(ICLR '24)} & 69.4 & 69.0 & 31.7 & 81.6 & 69.8 & \textbf{86.7} & \textbf{92.6} & 79.6 & 94.4 & 59.3 & 74.7 & 73.5 \\
\rowcolor{LightGray} & SO \scriptsize{(Ours)} & \underline{70.5} & \textbf{72.3} & \textbf{37.2} & 82.7 & \textbf{74.4} & \underline{85.3} & \underline{92.2} & \textbf{91.9} & \textbf{95.3} & \underline{60.2} & \textbf{80.4} & \textbf{76.6} \\

\midrule
\multirow{8}{*}{4}
& Adam \scriptsize{(ICLR '15)} & 0.1 & 2.1 & 2.4 & 47.9 & 0.7 & 3.1 & 6.3 & 30.5 & 12.5 & 9.6 & 9.2 & 11.3 \\
& LoRA \scriptsize{(ICLR '22)} & 68.5 & 69.7 & 35.2 & 85.2 & 74.8 & 78.7 & 87.9 & \underline{94.1} & 93.9 & 63.5 & 78.2 & 75.4 \\
& ReLoRA \scriptsize{(ICLR '24)} & \underline{71.2} & \underline{72.2} & \underline{37.6} & 84.7 & \underline{77.1} & 82.3 & 90.3 & \underline{94.1} & 94.9 & 64.0 & \underline{80.0} & \underline{77.1} \\
& GaLoRE \scriptsize{(ICML '24)} & 67.5 & 69.0 & 34.1 & 82.1 & 73.1 & 75.2 & 81.0 & 91.1 & 93.2 & 60.0 & 75.4 & 72.9 \\
& PiSSA \scriptsize{(NeurIPS '24)} & 66.4 & 68.8 & 32.4 & 82.8 & 73.2 & 75.6 & 80.6 & 92.0 & 92.9 & 61.1 & 75.7 & 72.9 \\
& DoRA \scriptsize{(ICML '24)}  & 68.4 & 69.7 & 34.9 & 85.5 & 75.1 & 79.0 & 89.5 & \underline{94.1} & 94.0 & 64.1 & 78.5 & 75.7 \\
& VeRA \scriptsize{(ICLR '24)} & 70.5 & 71.5 & 35.6 & \underline{85.6} & 73.7 & \textbf{86.7} & \textbf{93.1} & 89.8 & \underline{95.3} & \underline{64.5} & 79.5 & 76.9 \\
\rowcolor{LightGray} & SO \scriptsize{(Ours)} & \textbf{71.4} & \textbf{73.7} & \textbf{38.6} & \textbf{87.7} & \textbf{78.9} & \underline{85.3} & \underline{92.4} & \textbf{95.1} & \textbf{95.5} & \textbf{66.4} & \textbf{83.4} & \textbf{78.9} \\

\bottomrule
\end{tabular}}
\end{table*}
\begin{table*}
\caption{Impact of gradient and moment sparsification in few-shot classification (ViT-B/16).}
\label{tab:sparse_vs_dense}
\centering
\resizebox{0.97\textwidth}{!}{
\begin{tabular}{llcccccccccccccc}
\toprule
Shots & Method & \multicolumn{2}{c}{Gradient} & \multicolumn{2}{c}{Moments} & Aircraft & EuroSAT & Cars & Food & Pets & Flowers & Caltech & DTD & UCF & Average \\ 
\cmidrule(lr){3-4} \cmidrule(lr){5-6} 
& & Dense & Sparse & Dense & Sparse &  &  &  &  &  &  &  &  &   &  \\
\midrule 
\multirow{3}{*}{1}
& SO ($\equiv$ Adam)  & \checkmark & \ding{55} & \checkmark & \ding{55} & 1.0 & 24.3 & 0.5 & 1.0 & 2.9 & 1.4 & 1.0 & 5.6 & 1.7 & 4.4\\
& SO  & \ding{55} & \checkmark & \checkmark & \ding{55} & \underline{29.4} & \underline{73.9} & \underline{69.6} & \underline{83.9} & \underline{93.2} & \textbf{86.4} & \underline{93.7} & \textbf{55.7} & \textbf{77.0} & \underline{73.6}\\
\rowcolor{LightGray} & SO & \ding{55} & \checkmark & \ding{55} & \checkmark & \textbf{31.5} & \textbf{78.2} & \textbf{71.6} & \textbf{86.2} & \textbf{93.3} & \underline{84.9} & \textbf{94.1} & \underline{55.3} & \underline{76.4} & \textbf{74.6}\\
\midrule
\multirow{3}{*}{2}
& SO ($\equiv$ Adam)  & \checkmark & \ding{55} & \checkmark & \ding{55} & 0.9 & 42.1 & 0.7 & 1.9 & 3.9 & 17.1 & 3.1 & 7.2 & 3.8 & 9.0\\
& SO  & \ding{55} & \checkmark & \checkmark & \ding{55} & \underline{29.4} & \underline{82.6} & \underline{71.0} & \underline{81.5} & \underline{91.4} & \underline{90.1} & \underline{93.7} & \underline{59.7} & \underline{78.4} & \underline{75.3}\\
\rowcolor{LightGray} & SO & \ding{55} & \checkmark & \ding{55} & \checkmark & \textbf{37.2} & \textbf{82.7} & \textbf{74.4} & \textbf{85.3} & \textbf{92.2} & \textbf{91.9} & \textbf{95.3} & \textbf{60.2} & \textbf{80.4} & \textbf{77.7}\\
\midrule
\multirow{3}{*}{4}
& SO ($\equiv$ Adam)  & \checkmark & \ding{55} & \checkmark & \ding{55} & 2.4 & 47.9 & 0.7 & 3.1 & 6.3 & 30.5 & 12.5 & 9.6 & 9.2 & 13.6\\
& SO  & \ding{55} & \checkmark & \checkmark & \ding{55} & \underline{34.5} & \underline{85.4} & \underline{73.5} & \underline{78.9} & \underline{91.1} & \underline{92.7} & \underline{94.2} & \underline{66.2} & \underline{78.3} & \underline{77.2}\\
\rowcolor{LightGray} & SO & \ding{55} & \checkmark & \ding{55} & \checkmark & \textbf{38.6} & \textbf{87.7} & \textbf{78.9} & \textbf{85.3} & \textbf{92.4} & \textbf{95.1} & \textbf{95.5} & \textbf{66.4} & \textbf{83.4} & \textbf{80.4} \\
\bottomrule
\end{tabular}}
\end{table*}

\noindent \textbf{Algorithm.} Our algorithm is provided in Appendix B due to space limitations. We finetune CLIP until convergence using our optimization technique. To promote efficiency, we retain the gradient sparsity support fixed for \( T \) iterations before reselecting the trainable parameters. Overall, our approach introduces two hyperparameters: the density ratio \( \kappa \), which controls the fraction of retained gradient entries, and the update interval \( T \), which determines how frequently the sparsity support is refreshed. A smaller \( \kappa \) value improves generalization but results in slower convergence. In contrast, a larger \( \kappa \) accelerates optimization at the cost of reduced performance. Similarly, reducing \( T \) leads to better results by enabling more frequent updates to the sparsity support, although increasing \( T \) speeds up convergence.

The role of \( T \) in our method is conceptually similar to update intervals used in prior work such as GaLore~\cite{zhao2024galore} and ReLoRA~\cite{lialin2023relora}. In ReLoRA, the low-rank matrices are merged with the original parameters every \( T \) interval. GaLore also updates the low-rank projectors every \( T \) interval. Unlike the low-rank methods, where hyperparameters such as rank and number of training iterations involve a strong trade-off between overfitting and underfitting, our sparsity-based approach remains robust to overfitting.

\begin{table*}
\caption{Impact of randomness and importance in gradient and moment sparsification in few-shot classification (ViT-B/16).}
\label{tab:rand_vs_impt}
\centering
\resizebox{0.97\textwidth}{!}{
\begin{tabular}{llcccccccccccccc}
\toprule
Shots & Method & \multicolumn{2}{c}{Gradient} & \multicolumn{2}{c}{Moments} & Aircraft & EuroSAT & Cars & Food & Pets & Flowers & Caltech & DTD & UCF & Average \\ 
\cmidrule(lr){3-4} \cmidrule(lr){5-6} 
& & Rand & Impt & Rand & Impt &  &  &  &  &  &  &  &  &   &  \\
\midrule 
\multirow{4}{*}{1}
& SO  & \ding{55} & \checkmark & \ding{55} & \checkmark & 17.6 & \underline{73.9} & 60.5 & 72.0 & 87.9 & 75.9 & 89.9 & 53.9 & 68.2 & 66.6\\
& SO  & \ding{55} & \checkmark & \checkmark & \ding{55} & 20.7 & 71.4 & 63.7 & 74.6 & \underline{92.0} & 79.9 & \underline{91.9} & 54.6 & 71.1 & 69.6\\
& SO  & \checkmark & \ding{55} & \checkmark & \ding{55} & \underline{31.2} & 71.7 & \underline{69.8} & \underline{85.6} & \textbf{93.3} & \underline{80.3} & \textbf{94.1} & \underline{55.0} & \underline{74.9} & \underline{72.9} \\
\rowcolor{LightGray} & SO  & \checkmark & \ding{55} & \ding{55} & \checkmark & \textbf{31.5} & \textbf{78.2} & \textbf{71.6} & \textbf{86.2} & \textbf{93.3} & \textbf{84.9} & \textbf{94.1} & \textbf{55.3} & \textbf{76.4} & \textbf{74.6}\\

\midrule
\multirow{4}{*}{2}
& SO  & \ding{55} & \checkmark & \ding{55} & \checkmark & 21.4 & 80.8 & 62.6 & 67.7 & 82.8 & 84.6 & 88.7 & 57.6 & 68.2 & 68.3 \\
& SO  & \ding{55} & \checkmark & \checkmark & \ding{55} & 22.8 & \underline{82.4} & 65.3 & 73.4 & 87.1 & 84.2 & 91.3 & \underline{60.3} & 74.5 & 71.3 \\
& SO  & \checkmark & \ding{55} & \checkmark & \ding{55} & \underline{33.2} & 82.1 & \underline{72.8} & \textbf{85.7} & \underline{92.1} & \underline{91.2} & \underline{94.8} & \textbf{60.9} & \underline{79.6} & \underline{76.9} \\
\rowcolor{LightGray} & SO  & \checkmark & \ding{55} & \ding{55} & \checkmark & \textbf{37.2} & \textbf{82.7} & \textbf{74.4} & \underline{85.3} & \textbf{92.2} & \textbf{91.9} & \textbf{95.3} & 60.2 & \textbf{80.4} & \textbf{77.7}\\
\midrule
\multirow{4}{*}{4}
& SO  & \ding{55} & \checkmark & \ding{55} & \checkmark & 28.2 & 83.2 & 67.9 & 66.9 & 69.8 & 86.4 & 89.0 & 55.9 & 69.9 & 62.5 \\
& SO  & \ding{55} & \checkmark & \checkmark & \ding{55} & 31.0 & 81.0 & 71.5 & 73.0 & 77.7 & 86.9 & 91.4 & 58.9 & 74.1 & 71.6 \\
& SO  & \checkmark & \ding{55} & \checkmark & \ding{55} & \textbf{39.5} & \underline{84.7} & \underline{77.6} & \textbf{86.2} & \textbf{92.7} & \underline{93.3} & \textbf{95.8} & \underline{66.0} & \underline{82.5} & \underline{79.8}\\
\rowcolor{LightGray} & SO  & \checkmark & \ding{55} & \ding{55} & \checkmark & \underline{38.6} & \textbf{87.7} & \textbf{78.9} & \underline{85.3} & \underline{92.4} & \textbf{95.1} & \underline{95.5} & \textbf{66.4} & \textbf{83.4} & \textbf{80.4} \\
\bottomrule
\end{tabular}}
\end{table*}

\begin{table*}
\caption{Impact of dynamic sparsity support for gradient pruning in few-shot classification (ViT-B/16).}
\label{tab:static_vs_dynamic}
\centering
\resizebox{0.97\textwidth}{!}{
\begin{tabular}{llcccccccccccc}
\toprule
Shots & Method & \multicolumn{2}{c}{Sparsity support} & Aircraft & EuroSAT & Cars & Food & Pets & Flowers & Caltech & DTD & UCF & Average \\ 
\cmidrule(lr){3-4} \cmidrule(lr){4-4} 
& & Static & Dynamic &  &  &  &  &  &  &  &  &   &  \\
\midrule 
1 & SO  & \checkmark & \ding{55} & \underline{28.2} & \underline{72.4} & \underline{68.1} & \underline{85.9} & \underline{93.1} & \underline{73.4} & \textbf{94.5} & \underline{52.6} & \underline{72.6} & \underline{71.2}\\
\rowcolor{LightGray}  & SO & \ding{55} & \checkmark & \textbf{31.5} & \textbf{78.2} & \textbf{71.6} & \textbf{86.2} & \textbf{93.3} & \textbf{84.9} & \underline{94.1} & \textbf{55.3} & \textbf{76.4} & \textbf{74.6}\\
\midrule
2 & SO  & \checkmark & \ding{55} & \underline{31.0} & \textbf{83.0} & \underline{69.7} & \textbf{85.8} & \underline{91.6} & \underline{83.0} & \underline{94.3} & \underline{55.7} & \underline{75.8} & \underline{74.4}\\
\rowcolor{LightGray}  & SO & \ding{55} & \checkmark & \textbf{37.2} & \underline{82.7} & \textbf{74.4} & \underline{85.3} & \textbf{92.2} & \textbf{91.9} & \textbf{95.3} & \textbf{60.2} & \textbf{80.4} & \textbf{77.7}\\
\midrule
4 & SO  & \checkmark & \ding{55} & \underline{34.8} & \underline{83.6} & \underline{72.4} & \textbf{86.2} & \textbf{92.5} & \underline{88.0} & \underline{95.1} & \underline{62.5} & \underline{77.8} & \underline{77.0}\\
\rowcolor{LightGray}  & SO & \ding{55} & \checkmark & \textbf{38.6} & \textbf{87.7} & \textbf{78.9} & \underline{85.3} & \underline{92.4} & \textbf{95.1} & \textbf{95.5} & \textbf{66.4} & \textbf{83.4} & \textbf{80.4} \\
\bottomrule
\end{tabular}}
\end{table*}

\begin{table*}
\caption{Impact of moments in few-shot classification (ViT-B/16).}
\label{tab:with_vs_without_moment}
\centering
\resizebox{0.97\textwidth}{!}{
\begin{tabular}{llcccccccccccc}
\toprule
Shots & Method & \multicolumn{2}{c}{Moments} & Aircraft & EuroSAT & Cars & Food & Pets & Flowers & Caltech & DTD & UCF & Average \\ 
\cmidrule(lr){3-4} \cmidrule(lr){4-4} 
& & without & with &  &  &  &  &  &  &  &  &   &  \\
\midrule 
1 & SO  & \checkmark & \ding{55} & \underline{20.9} & \underline{74.7} & \underline{51.9} & \underline{76.5} & \underline{85.7} & \underline{82.9} & \underline{92.5} & \underline{54.5} & \underline{70.2} & \underline{67.8} \\
\rowcolor{LightGray}  & SO & \ding{55} & \checkmark & \textbf{31.5} & \textbf{78.2} & \textbf{71.6} & \textbf{86.2} & \textbf{93.3} & \textbf{84.9} & \textbf{94.1} & \textbf{55.3} & \textbf{76.4} & \textbf{74.6}\\
\midrule
2 & SO  & \checkmark & \ding{55} & \underline{24.9} & \textbf{84.6} & \underline{60.2} & \underline{77.5} & \underline{89.5} & \underline{90.9} & \underline{92.9} & \textbf{61.1} & \underline{76.0} & \underline{73.0}\\
\rowcolor{LightGray}  & SO & \ding{55} & \checkmark & \textbf{37.2} & \underline{82.7} & \textbf{74.4} & \textbf{85.3} & \textbf{92.2} & \textbf{91.9} & \textbf{95.3} & \underline{60.2} & \textbf{80.4} & \textbf{77.7}\\
\midrule
4 & SO  & \checkmark & \ding{55} & \underline{32.2} & \textbf{89.1} & \underline{72.1} & \underline{79.0} & \underline{89.1} & \underline{93.5} & \underline{93.8} & \underline{64.4} & \underline{77.3} & \underline{76.7}\\
\rowcolor{LightGray}  & SO & \ding{55} & \checkmark & \textbf{38.6} & \underline{87.7} & \textbf{78.9} & \textbf{85.3} &\textbf{92.4} & \textbf{95.1} & \textbf{95.5} & \textbf{66.4} & \textbf{83.4} & \textbf{80.4} \\
\bottomrule
\end{tabular}}
\end{table*}

\section{Results and Discussion}
\label{sec:experiment}

\noindent \textbf{(a) Experimental Methodology.} Our optimizer is evaluated on 11 datasets commonly used in few-shot VLM adaptation \cite{zanella2024low, zhou2022learning}, including ImageNet \cite{imagenet}, SUN397 \cite{sun397}, Aircraft \cite{aircraft}, EuroSAT \cite{eurosat}, Stanford-Cars \cite{stanford_cars}, Food101 \cite{food101}, OxfordPets \cite{oxford_pets}, Flowers102 \cite{flowers102}, Caltech101 \cite{caltech101}, DTD \cite{dtd}, and UCF101 \cite{ucf101}. These datasets provide a comprehensive assessment across a variety of image domains.

We evaluate SO against state-of-the-art low-rank methods, which have shown strong performance in fine-tuning Large Language Models and Vision Language models. Specifically, we compare our method against LoRA, ReLoRA, VeRA, DoRA, PiSSA, and GaLoRE. Our comparison also includes full fine-tuning using Adam to showcase the effect of severe overfitting. A discussion of these baselines, including their limitations, is provided in Sec.~\ref{sec:related_work}.

For a fair comparison, we follow the same hyperparameter settings as CLIP-LoRA \cite{zanella2024low} for the low-rank baselines. In particular, we set the rank to 2, and we train for 500 iterations. Due to severe overfitting, we introduce an early stopping criterion for the low-rank methods. If the training loss reaches \(0.01\), training is halted. This is particularly necessary to obtain competitive results using the low-rank approaches for some datasets where the test accuracy peaks often within a few iterations before rapidly overfitting. For our method, we set the density ratio to $\kappa = 0.05\%$, and we update the gradient sparsity support every $T=10$ iterations. We train until convergence, defined as reaching a loss threshold of \(0.01\). If convergence is not achieved, we stop training at a maximum limit of 2000 iterations.

Following CLIP-LoRA \cite{zanella2024low}, we adopt a learning rate of \(2 \times 10^{-4}\) with a cosine learning rate scheduler and a batch size of 32. We use the simple prompt template ``a photo of a [class name]" without complex manual prompt engineering. All experiments are conducted using a ViT-B/16 backbone, and finetuning is applied to all layers of the 12 vision and text Transformer blocks. These settings were suggested by CLIP-LoRA and are kept the same for all methods, including ours. All hyperparameters remain fixed across all datasets for our approach and the baselines. 

All experiments are conducted under identical hardware and software conditions. The details of these environments are provided in Appendix E. Additional experimental results showing CLIP’s few-shot adaptation performance are presented in Appendix F. A comparative analysis of the memory efficiency of SO and state-of-the-art low-rank methods is presented in Appendix C. The sensitivity of our approach to the hyperparameters \(\kappa\) and \(T\) is explored in Appendix D. Moreover, to further assess the effectiveness of our optimizer, we conduct both standard classification and few-shot learning experiments using a simple two-layer fully connected network. The results obtained with this small architecture are detailed in Appendix G.


\noindent \textbf{(b) Sparsity Outperforms Low-Rank Projections.} The results in Table~\ref{tab:vitb16} show the effectiveness of SO compared to state-of-the-art low-rank techniques. Across all shots (1, 2, and 4), our method achieves the highest average performance. Furthermore, SO yields the highest accuracy in most datasets, including EuroSAT, Cars, Flowers, and UCF101. Notably, it outperforms LoRA and ReLoRA on Aircraft by significant margins. Low-rank reparametrization methods constrain updates to a fixed subspace. The only exception is ReLoRA, which periodically merges and reinitializes the auxiliary low-rank matrices; this might explain why ReLoRA outperforms the other low-rank reparametrization methods. Moreover, these methods suffer from severe overfitting and require careful tuning of the rank and number of training iterations to achieve their highest performance. Unlike these methods, 
SO employs dynamic sparsity to update only a minimal subset of parameters, thereby mitigating overfitting while preserving the model's full representational capacity.


\noindent \textbf{(d) Effectiveness of Pruning in Few-Shot Optimization.} Table~\ref{tab:sparse_vs_dense} shows the impact of gradient and moment pruning on few-shot classification. Dense optimization (Adam) performs poorly in the few-shot setting due to severe overfitting. In contrast, gradient pruning leads to substantial performance improvement. Moment pruning further enhances performance by filtering out updates related to parameters that are only locally important. These findings validate that sparsity can improve generalization in few-shot learning.

\noindent \textbf{(e) Local Randomness Global Importance.} Table~\ref{tab:rand_vs_impt} presents the performance of four sparse optimization variants based on randomness and importance in both gradient and moment pruning. Importance-based pruning is achieved by retaining the values with the highest magnitude. The results indicate that applying random selection for gradient pruning and importance-based selection for moment pruning achieves the highest average performance across all few-shot settings. Results in Appendix F confirm that importance-based gradient selection leads to rapid overfitting. Furthermore, Table~\ref{tab:rand_vs_impt} shows that importance-based moment pruning outperforms random moment pruning. This is because significant parameters over the long term are consistently updated, while locally influential ones are filtered out from the moment updates. Overall, our results validate the local randomness and global importance paradigm. Stochastic gradient selection prevents bias toward high-magnitude local updates and importance-based pruning preserves influential parameters over the long term.



\noindent \textbf{(f) Local Sparsity Global Density} Table~\ref{tab:static_vs_dynamic} shows that dynamic sparsity support consistently outperforms fixed sparsity support in gradient pruning across all few-shot settings. This improvement validates the local sparsity and global density paradigm. On the one hand, dynamic sparsity ensures that few parameters contribute to the updates to reduce the local learning capacity and thus mitigate overfitting. On the other hand, dynamic sparsity ensures a broader exploration of the parameter space while making the learning process benefit from the model’s overall expressiveness. 

\noindent \textbf{(g) Role of Moments in Few-Shot Adaptation.} Table~\ref{tab:with_vs_without_moment} shows that moment-based updates significantly improve performance. The advantage is more pronounced in datasets with higher complexity, such as Aircraft and Cars. Without moments, updates rely solely on randomly selected gradient values. These results provide evidence that gradient pruning is not sufficient. Random gradient pruning is more effective when combined with moments, particularly sparse ones.

\section{Conclusion}
\label{sec:conclusion}

In this paper, we shed light on sparsity as an effective mechanism to mitigate overfitting in few-shot adaptation of VLMs. Conventional low-rank methods suffer from severe overfitting and require extensive hyperparameter tuning. To address these limitations, we introduce two key paradigms. The first paradigm consists of dynamically selecting a minimal subset of parameters to reduce overfitting while preserving the overall model expressiveness. The second paradigm aims to prevent bias in gradient updates while retaining critical long-term information. Our experiments confirm that sparsity outperforms low-rank projections in few-shot learning and constitutes a strong mechanism against overfitting. As future work, we plan to evaluate our optimizer on large language models. Memory efficiency in this setting is even more critical.




\section*{Acknowledgement}
This work was supported by the Natural Sciences and Engineering Research Council of Canada (NSERC), and the Digital Research Alliance of Canada.




\clearpage
\onecolumn
\appendix

\begin{center}
    \vspace*{2em}
    {\Large \textbf{Supplementary Material for:}}\\[0.5em]
    {\Large \textbf{``Sparsity Outperforms Low-Rank Projections in Few-Shot Adaptation''}}\\[2em]
\end{center}

\section{Few-shot Adaptation of VLMs}
\label{sec:preliminaries}

VLMs are designed to learn a joint representation space for images and text. They are typically trained on large-scale image-text pairs. By aligning visual and textual information, VLMs can perform tasks such as zero-shot classification and cross-modal retrieval. 
CLIP is a type of VLM that relies on contrastive learning to associate images with their corresponding textual descriptions in the latent space. Given a batch of $N$ image-text pairs $\{(x_i, \, t_i)\}_{i=1}^{N}$, where $x_i$ represents an image and $t_i$ its associated text, CLIP maximizes the similarity between matching pairs while minimizing it for non-matching ones. 

Formally, let \( f_{\theta}(\cdot) \) and \( g_{\phi}(\cdot) \) be the vision and text encoders, respectively, where \( \theta \) and \( \phi \) denote their learnable parameters. Given an image \( x_i \) and its corresponding text \( t_i \), the encoders project them into a shared normalized embedding space. The image embedding is given by \( v_i = f_{\theta}(x_i) \), and the text embedding is given by \( z_i = g_{\phi}(t_i) \). Both embeddings lie in the joint representation space. The VLM model is then trained using a symmetric InfoNCE loss.

\noindent \textbf{Zero-shot Inference.} Once pretrained, CLIP classifies images without additional training on the target task. Classification is performed by comparing the image embedding to predefined text embeddings representing class labels. For a classification task with \( C \) categories, each class \( c \) is associated with a set of $n$ text prompts \( \{t_{j,c}\}_{j=1}^{n} \). The prototype for class \( c \) is computed by averaging the embeddings of its prompts \( z_c = \frac{1}{n} \sum_{j=1}^{n} g_{\phi}(t_{j,c}) \). The model predicts the class using the softmax over the cosine similarities between the image embedding \( v_{i} \) and the class prototypes \( z_{c} \):
\begin{equation}
\hat{y}_{ic} = \frac{\exp\big((v_{i}^{\top} \cdot z_c)/\tau\big)}{\sum_{j=1}^{C} \exp\big((v_{i}^{\top} \cdot z_j)/\tau\big)},
\end{equation}
\noindent where \( \hat{y}_{ic} \) represents the probability of the image \(x_{i}\) belonging to class \( c \), and \( \tau \) is a temperature parameter that scales the logits to control the sharpness of the probability distribution. Since the embeddings \( v_{i} \) and \( z_c \) are \( \ell_2 \)-normalized, the cosine similarity simplifies to the dot product operation.

\noindent \textbf{Few-shot Learning.} Few-shot learning addresses the challenge of adapting VLMs to new tasks using only a limited number of labeled examples per class. Formally, let \( S = \{(x_i, \, y_i)\}_{i=1}^{K \times C} \) be the support set, where \( K \) is the number of examples per class. \( K \) typically takes small values, such as \( K \in \{1, \, 2, \, 4 \} \). Each label \( y \in \{0,1\}^C \) is represented as a one-hot vector, where only one dimension is active to indicate the correct class. The objective is to adapt the pretrained VLM efficiently by leveraging this small support set \( S \) while preserving its generalization ability.

To optimize the VLM model under the few-shot setting, the cross-entropy loss function is typically used. Given the support set \( S \), the objective is to minimize:
\begin{equation}
\mathcal{L}_{CE} = -\frac{1}{K} \sum_{i=1}^{K} \sum_{j=1}^{C} y_{ij} \, \ln \hat{y}_{ij}.
\end{equation}
This loss function maximizes the likelihood of the correct class labels and minimizes incorrect predictions.

\section{Algorithm}

Our optimization strategy is described in Algorithm \ref{alg:SO}. By promoting random gradient selection and importance-based moment pruning, SO balances the short-term updates with the long-term significance of parameters.

\begin{algorithm}
\caption{SO: Sparse Optimization Algorithm}
\label{alg:SO}
\begin{algorithmic}[1]
\Require \( \eta \) (learning rate), \( \beta_1, \beta_2 \in [0,1] \) (exponential decay rates for moment estimates), \( \kappa \) (density ratio), \( T \) (number of iterations before updating sparsity support),
\( \epsilon \) (numerical stability constant), \( \tau \) (convergence rate)
\Require \( \Theta_0 \) 
\State \( \tilde{\mu}_0 \gets 0 \) 
\State \( \tilde{\nu}_0 \gets 0 \) 
\State \( t \gets 0 \) 
\State $\mathfrak{I} \gets \emptyset $ ($\mathfrak{I}$ denotes gradient sparsity support) 
\While {$ \left| \mathcal{L}(\Theta_{t-1}) \right| > \tau $}
    \State \( t \gets t + 1 \)
    \State \( g_t \gets \nabla_{\Theta} \mathcal{L}(\Theta_{t-1}) \)
    \State \( M \gets \lfloor \kappa d \rfloor \) 
    
    \If {$(t-1) \mod T == 0$} 
        \State \( \tilde{g}_t \gets \text{Random-}M(g_t) \) 
        \State \( \mathfrak{I} \gets \mathcal{I}(\tilde{g}_t ) \) 
    \Else
        \State \( \tilde{g}_t \gets g_t[\mathfrak{I}] \)
    \EndIf
    
    \State \( \mu_t \gets \beta_1 \tilde{\mu}_{t-1} + (1 - \beta_1) \tilde{g}_t \) 
    \State \( \nu_t \gets \beta_2 \tilde{\nu}_{t-1} + (1 - \beta_2) \tilde{g}_t^2 \) 

    \If {$(t-1) \mod T == 0$} 
        \State \( \tilde{\mu}_t \gets \text{Top-}M(\mu_t) \) 
        \State \( \tilde{\nu}_t \gets \nu_t[\mathcal{I}(\tilde{\mu}_t)] \) 
    \Else
        \State \( \tilde{\mu}_t \gets \mu_t[\mathfrak{I}] \)
        \State \( \tilde{\nu}_t \gets \nu_t[\mathfrak{I}] \) 
    \EndIf
    
    \State \( \hat{\mu}_t \gets \frac{\tilde{\mu}_t}{1 - \beta_1^t} \) 
    \State \( \hat{\nu}_t \gets \frac{\tilde{\nu}_t}{1 - \beta_2^t} \) 
    
    \State \( \Theta_t \gets \Theta_{t-1} - \frac{\eta}{\sqrt{\hat{\nu}_t} + \epsilon} \hat{\mu}_t \) 
\EndWhile
\State \textbf{return} \( \Theta_t \) 
\end{algorithmic}
\end{algorithm}
 
\section{Memory Consumption for CLIP}
\label{sec:clip_memory}

\begin{table*}
\centering
\caption{Comparison of SO, GaLoRE, LoRA, PiSSA, DoRA, ReLoRA, VeRA, and Adam in memory requirements for a single fully-connected layer. Denote the weight of the layer $W \in \mathbb{R}^{m \times n}$, $n$ the input dimension, $m$ the output dimension, $(m \leq n)$, rank $r$, and sparsity ratio $1 - \kappa$.}
\resizebox{1\textwidth}{!}{\begin{tabular}{lcccccccc}
\toprule
                   & SO             & GaLoRE        & LoRA                & PiSSA               & DoRA                   & ReLoRA                 & VeRA                      & Adam \\
\midrule
Weight             & $mn$           & $mn$          & $mn + mr + nr$      & $mn + mr + nr$      & $mn + mr + nr + m$     & $mn + mr + nr$         & $mn + mr + nr + m + r$    & $mn$ \\
Gradient           & $2mn\kappa$           & $mn$          & $mr + nr$           & $mr + nr$           & $mr + nr + m$          & $mr + nr$              & $m + r$                   & $mn$ \\
Optimizer States   & $3mn\kappa$    & $mr + 2nr$    & $2mr + 2nr$         & $2mr + 2nr$         & $2mr + 2nr + 2m$       & $2mr + 2nr$            & $2m + 2r$                 & $2mn$ \\
\bottomrule
\end{tabular}}
\label{tab:conceptual_memory}
\end{table*}
\begin{table*}
\centering
\caption{Comparison of theoretical memory consumption for CLIP when adapting all vision and text transformer blocks. Biases and activations are excluded since they are shared across all methods. The table reports the total number of variables, overall memory usage, and trainable parameters.}

\scalebox{0.7}{\begin{tabular}{lccccc}
\toprule
Method & Weight (\#Vars, MB) & Gradient (\#Vars, MB) & Opt. States (\#Vars, MB) & \#Trainable & Total Mem. (MB) \\
\midrule
SO ($\kappa=0.05\%$) & $122683392\,(468\text{MB})$ & $122683\,(0.47\text{MB})$ & $184025\,(0.70\text{MB})$ & $61341$ & 469.17\text{MB} \\
SO ($\kappa=1\%$) & $122683392\,(468\text{MB})$ & $2453667\,(9.36\text{MB})$ & $3680501\,(14.04\text{MB})$ & $1226833$ & 491.40\text{MB} \\
SO ($\kappa=2\%$) & $122683392\,(468\text{MB})$ & $4907335\,(18.72\text{MB})$ & $7361003\,(28.08\text{MB})$ & $2453667$ & 514.80\text{MB} \\
SO ($\kappa=5\%$) & $122683392\,(468\text{MB})$ & $12268339\,(46.80\text{MB})$ & $18402508\,(70.20\text{MB})$ & $6134169$ & 585.00\text{MB} \\
SO ($\kappa=8\%$) & $122683392\,(468\text{MB})$ & $19629342\,(74.88\text{MB})$ & $29444014\,(112.32\text{MB})$ & $9814671$ & 655.20\text{MB} \\
SO ($\kappa=10\%$) & $122683392\,(468\text{MB})$ & $24536678\,(93.60\text{MB})$ & $36805017\,(140.40\text{MB})$ & $12268339$ & 702.00\text{MB} \\
\midrule
GaLoRE ($r=2$) & $122683392\,(468.00\text{MB})$ & $122683392\,(468.00\text{MB})$ & $706560\,(2.70\text{MB})$ & $122683392$ & 938.70\text{MB} \\
GaLoRE ($r=4$) & $122683392\,(468.00\text{MB})$ & $122683392\,(468.00\text{MB})$ & $1413120\,(5.39\text{MB})$ & $122683392$ & 941.39\text{MB} \\
GaLoRE ($r=8$) & $122683392\,(468.00\text{MB})$ & $122683392\,(468.00\text{MB})$ & $2826240\,(10.78\text{MB})$ & $122683392$ & 946.78\text{MB} \\
GaLoRE ($r=16$) & $122683392\,(468.00\text{MB})$ & $122683392\,(468.00\text{MB})$ & $5652480\,(21.56\text{MB})$ & $122683392$ & 957.56\text{MB} \\
LoRA ($r=2$) & $123174912\,(469.88\text{MB})$ & $491520\,(1.88\text{MB})$ & $983040\,(3.75\text{MB})$ & $491520$ & 475.50\text{MB} \\
LoRA ($r=4$) & $123666432\,(471.75\text{MB})$ & $983040\,(3.75\text{MB})$ & $1966080\,(7.50\text{MB})$ & $983040$ & 483.00\text{MB} \\
LoRA ($r=8$) & $124649472\,(475.50\text{MB})$ & $1966080\,(7.50\text{MB})$ & $3932160\,(15.00\text{MB})$ & $1966080$ & 498.00\text{MB} \\
LoRA ($r=16$) & $126615552\,(483.00\text{MB})$ & $3932160\,(15.00\text{MB})$ & $7864320\,(30.00\text{MB})$ & $3932160$ & 528.00\text{MB} \\
PiSSA ($r=2$) & $123174912\,(469.88\text{MB})$ & $491520\,(1.88\text{MB})$ & $983040\,(3.75\text{MB})$ & $491520$ & 475.50\text{MB} \\
PiSSA ($r=4$) & $123666432\,(471.75\text{MB})$ & $983040\,(3.75\text{MB})$ & $1966080\,(7.50\text{MB})$ & $983040$ & 483.00\text{MB} \\
PiSSA ($r=8$) & $124649472\,(475.50\text{MB})$ & $1966080\,(7.50\text{MB})$ & $3932160\,(15.00\text{MB})$ & $1966080$ & 498.00\text{MB} \\
PiSSA ($r=16$) & $126615552\,(483.00\text{MB})$ & $3932160\,(15.00\text{MB})$ & $7864320\,(30.00\text{MB})$ & $3932160$ & 528.00\text{MB} \\
DoRA ($r=2$) & $123313152\,(470.40\text{MB})$ & $629760\,(2.40\text{MB})$ & $1259520\,(4.80\text{MB})$ & $629760$ & 477.61\text{MB} \\
DoRA ($r=4$) & $123804672\,(472.28\text{MB})$ & $1121280\,(4.28\text{MB})$ & $2242560\,(8.55\text{MB})$ & $1121280$ & 485.11\text{MB} \\
DoRA ($r=8$) & $124787712\,(476.03\text{MB})$ & $2104320\,(8.03\text{MB})$ & $4208640\,(16.05\text{MB})$ & $2104320$ & 500.11\text{MB} \\
DoRA ($r=16$) & $126753792\,(483.53\text{MB})$ & $4070400\,(15.53\text{MB})$ & $8140800\,(31.05\text{MB})$ & $4070400$ & 530.11\text{MB} \\
ReLoRA ($r=2$) & $123174912\,(469.88\text{MB})$ & $491520\,(1.88\text{MB})$ & $983040\,(3.75\text{MB})$ & $491520$ & 475.50\text{MB} \\
ReLoRA ($r=4$) & $123666432\,(471.75\text{MB})$ & $983040\,(3.75\text{MB})$ & $1966080\,(7.50\text{MB})$ & $983040$ & 483.00\text{MB} \\
ReLoRA ($r=8$) & $124649472\,(475.50\text{MB})$ & $1966080\,(7.50\text{MB})$ & $3932160\,(15.00\text{MB})$ & $1966080$ & 498.00\text{MB} \\
ReLoRA ($r=16$) & $126615552\,(483.00\text{MB})$ & $3932160\,(15.00\text{MB})$ & $7864320\,(30.00\text{MB})$ & $3932160$ & 528.00\text{MB} \\
VeRA ($r=2$) & $123313344\,(470.40\text{MB})$ & $138432\,(0.53\text{MB})$ & $276864\,(1.06\text{MB})$ & $138432$ & 471.99\text{MB} \\
VeRA ($r=4$) & $123805056\,(472.28\text{MB})$ & $138624\,(0.53\text{MB})$ & $277248\,(1.06\text{MB})$ & $138624$ & 473.87\text{MB} \\
VeRA ($r=8$) & $124788480\,(476.03\text{MB})$ & $139008\,(0.53\text{MB})$ & $278016\,(1.06\text{MB})$ & $139008$ & 477.62\text{MB} \\
VeRA ($r=16$) & $126755328\,(483.53\text{MB})$ & $139776\,(0.53\text{MB})$ & $279552\,(1.07\text{MB})$ & $139776$ & 485.13\text{MB} \\
\midrule
Adam (Full Finetune) & $122683392\,(468.00\text{MB})$ & $122683392\,(468.00\text{MB})$ & $245366784\,(936.00\text{MB})$ & $122683392$ & 1872.00\text{MB} \\
\bottomrule
\end{tabular}}
\label{tab:memory_clip}
\end{table*}

Table~\ref{tab:conceptual_memory} compares SO to various low-rank methods and Adam in theoretical memory usage for a single linear layer \((W \in \mathbb{R}^{m \times n})\). It illustrates how each method’s 
weight, gradient, and optimizer states scale w.r.t.\ rank \(r\) or sparsity ratio \(1 - \kappa\). Notably, SO stores \(2mn\kappa\) and \(3mn\kappa\) parameters for the gradient and optimizer states, 
respectively, offering significant savings when \(\kappa\) is small.

Table~\ref{tab:memory_clip} extends this analysis to the full CLIP model, covering all 12 blocks of both the text and vision encoders. We exclude biases and activations since they are shared across all approaches. Our experiments use a default sparsity ratio of \(1 - \kappa\) with 
\(\kappa = 0.05\%\). This extremely sparse update leads to minimal overheads in the gradient and optimizer states (\(\approx 0.47\,\text{MB}\) and \(0.70\,\text{MB}\), respectively). As a result, the total memory 
grows only slightly beyond the baseline weight storage. Rank-based techniques, by contrast, rely on separate low-rank matrices and typically require more memory than SO at extreme sparsities. Adam imposes the highest overhead due to storing a full gradient and two full optimizer states for every parameter.

Hence, even at very low \(\kappa\) (i.e., \(0.05\%\)), SO preserves adaption flexibility while significantly reducing memory consumption, which is particularly advantageous in few-shot or resource-constrained settings.

\section{Sensitivity}

\begin{figure}[!h]
    \centering
    \begin{subfigure}[b]{0.49\linewidth}
        \centering
        \includegraphics[width=\textwidth]{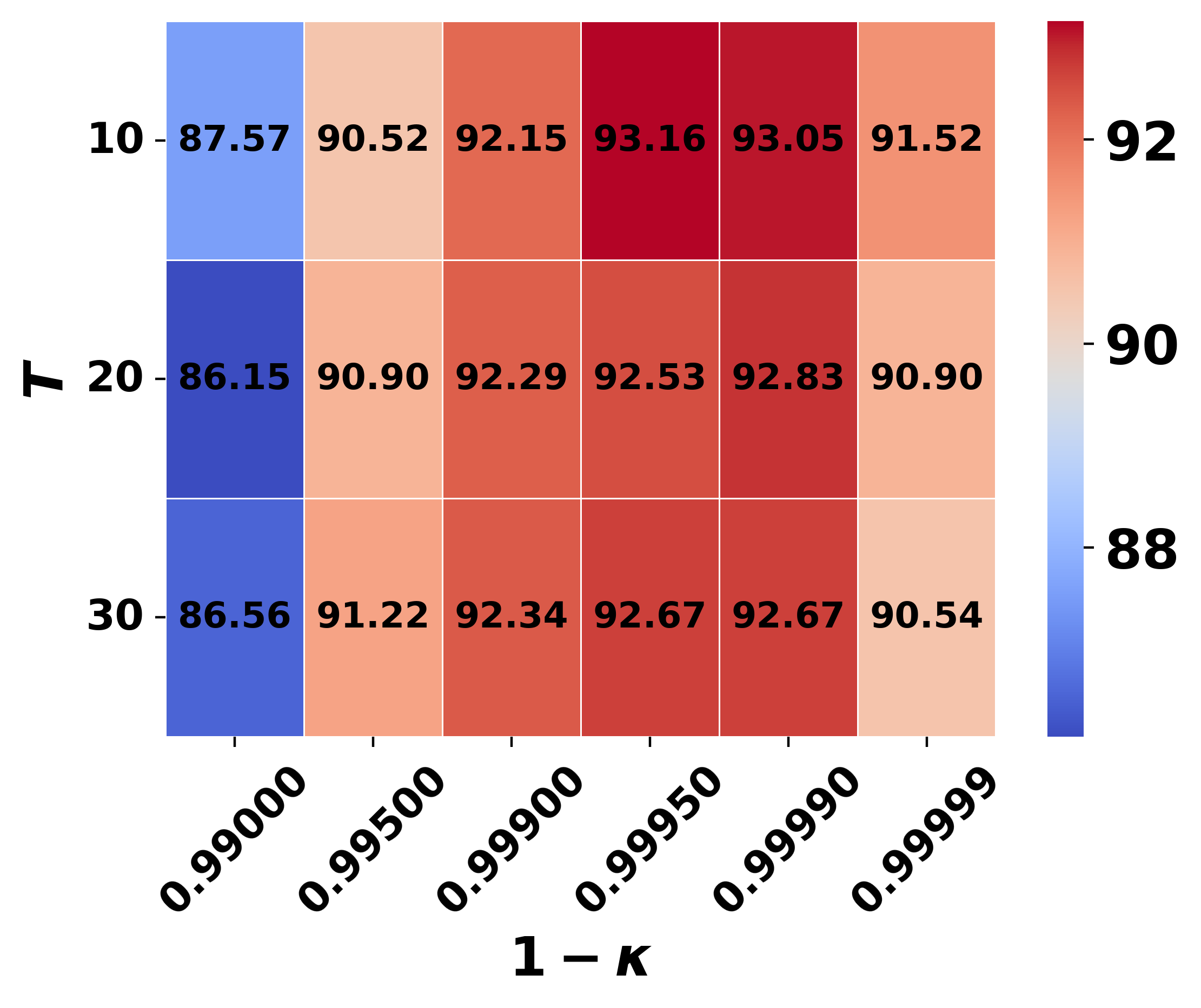}
        \caption{Pets}
        \label{fig:heatmap_oxford_pets}
    \end{subfigure}
    \begin{subfigure}[b]{0.49\linewidth}
        \centering
        \includegraphics[width=\textwidth]{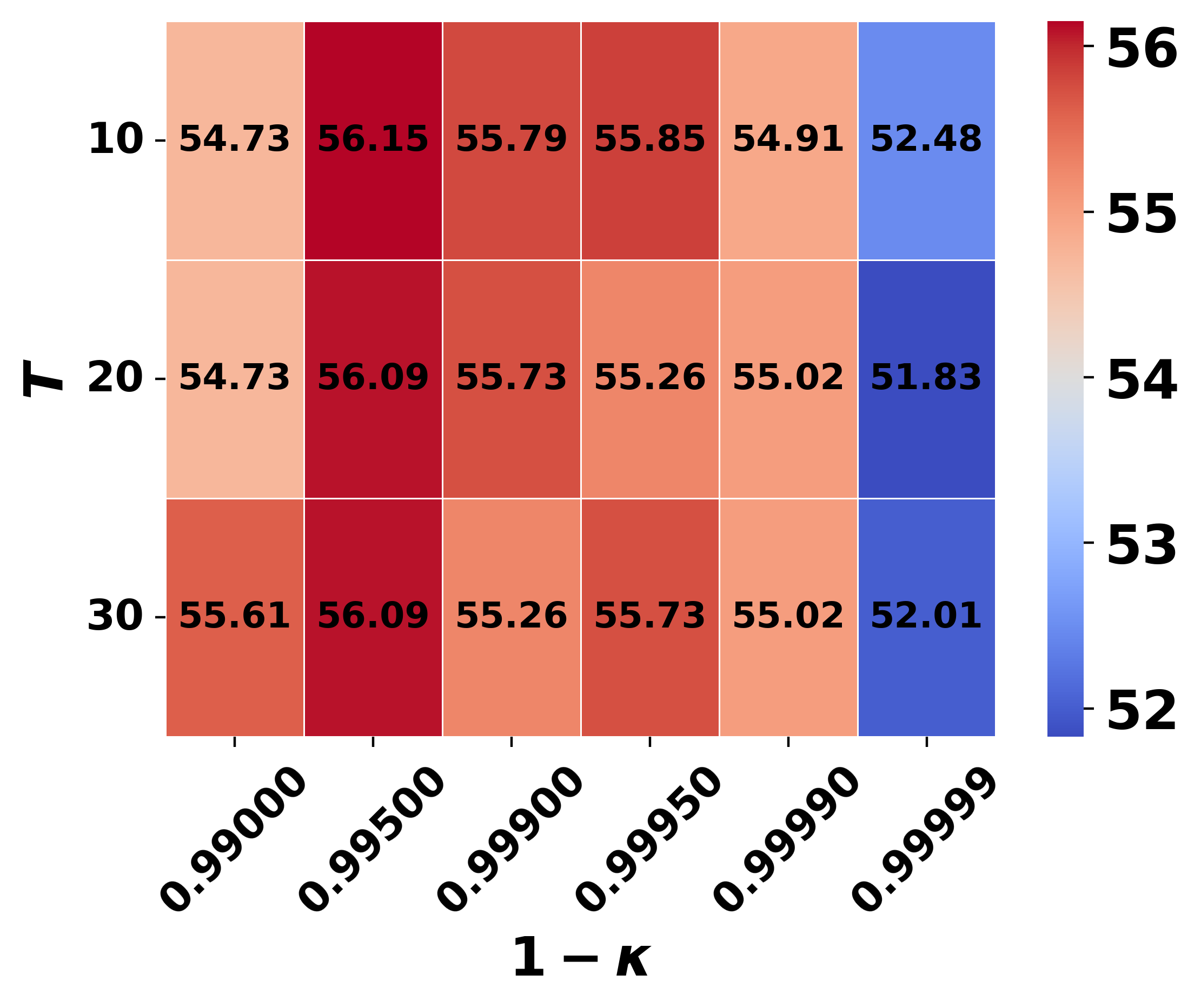}
        \caption{DTD}
        \label{fig:heatmap_dtd}
    \end{subfigure}
    \caption{Test accuracy of SO for different density ratios $\kappa$ and update intervals $T$ (1-shot setting).}
    \label{fig:heatmaps}
\end{figure}

Fig. \ref{fig:heatmaps} illustrates the impact of the density ratio \( \kappa \) and the update interval \( T \) on one-shot adaptation. In these experiments, we vary \(\kappa\) over a set of values \( \{0.99, \, 0.995, \, 0.999, \, 0.9995, \, 0.9999, \, 0.9999\} \) and choose update intervals \(T \in \{10, \, 20, \, 30\}\). 

A smaller \( \kappa \) generally leads to improved performance as it prevents overfitting by retaining fewer parameter updates per step. However, overly low \(\kappa\) reduces the model’s effective capacity and degrades accuracy. Similarly, smaller \( T \) values lead to better performance by promoting a more dynamic selection of trainable parameters. 
 
\section{Hardware and Software}
\label{hardware_softaware}

All experiments are conducted on a Linux server with a consistent hardware and software environment. Table \ref{tab:hardware_software} provides details on the hardware and software used.

\begin{table}[!h]
  \caption{Hardware and software.}
  \begin{center}
  \begin{small}
  \scalebox{0.9}{\begin{tabular}{|p{2.7cm}|c|} \hline 
    \multicolumn{2}{|c|}{\textbf{Hardware}}\\ \hline
    \textbf{RAM} & 504 GB \\ \hline
    \textbf{CPU model} & Intel(R) Xeon(R) Silver 4310 CPU @ 2.10GHz \\ \hline
    \textbf{\# of CPUs} & 48 \\ \hline
    \textbf{GPU model} & NVIDIA RTX A6000 \\ \hline
    \textbf{GPU memory} & 48 GB \\ \hline
    \textbf{\# of GPUs} & 4 \\ \hline
    \multicolumn{2}{|c|}{\textbf{Software}}\\ \hline
    \textbf{Operating System} & Ubuntu 18.04.6 LTS \\ \hline
    \textbf{Python} & 3.10.16 \\ \hline
    \textbf{PyTorch} & 2.5.1 \\ \hline
    \textbf{CUDA} & 12.4 \\ \hline
  \end{tabular}}
  \end{small}
  \end{center}
  \label{tab:hardware_software}
\end{table} 
\section{Additional Results with CLIP (ViT-B/16)}

\begin{figure*}
\centering
\renewcommand{\arraystretch}{1.3} 

\begin{tabular}{c@{\hskip 10pt} c c c c} 
    & Rank 2 & Rank 3 & Rank 4 & Rank 5 \\

    \rotatebox{90}{\usebox1} & 
    \begin{subfigure}[b]{0.22\textwidth}
        \centering
        \includegraphics[width=\linewidth]{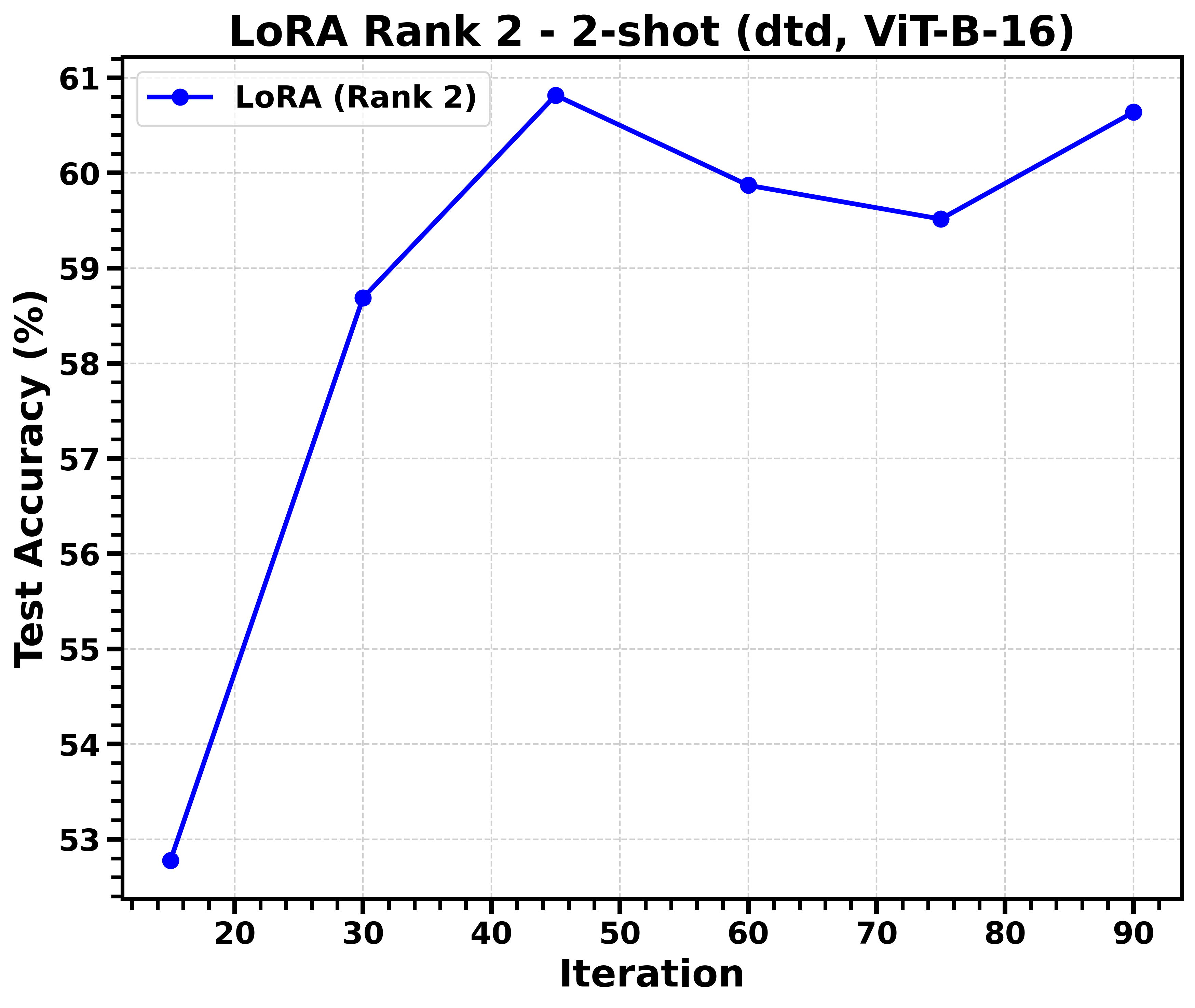}
    \end{subfigure} &
    \begin{subfigure}[b]{0.22\textwidth}
        \centering
        \includegraphics[width=\linewidth]{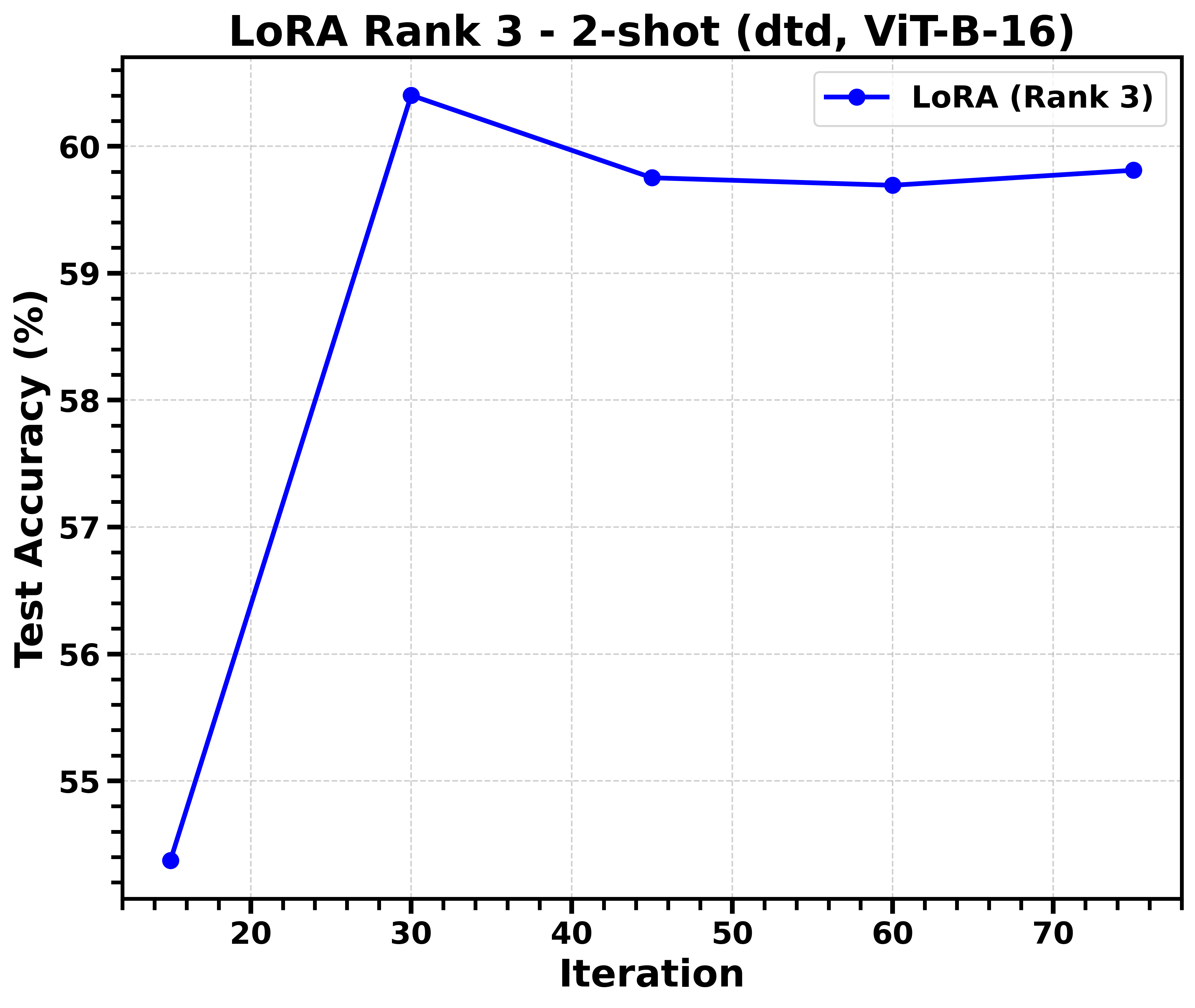}
    \end{subfigure} &
    \begin{subfigure}[b]{0.22\textwidth}
        \centering
        \includegraphics[width=\linewidth]{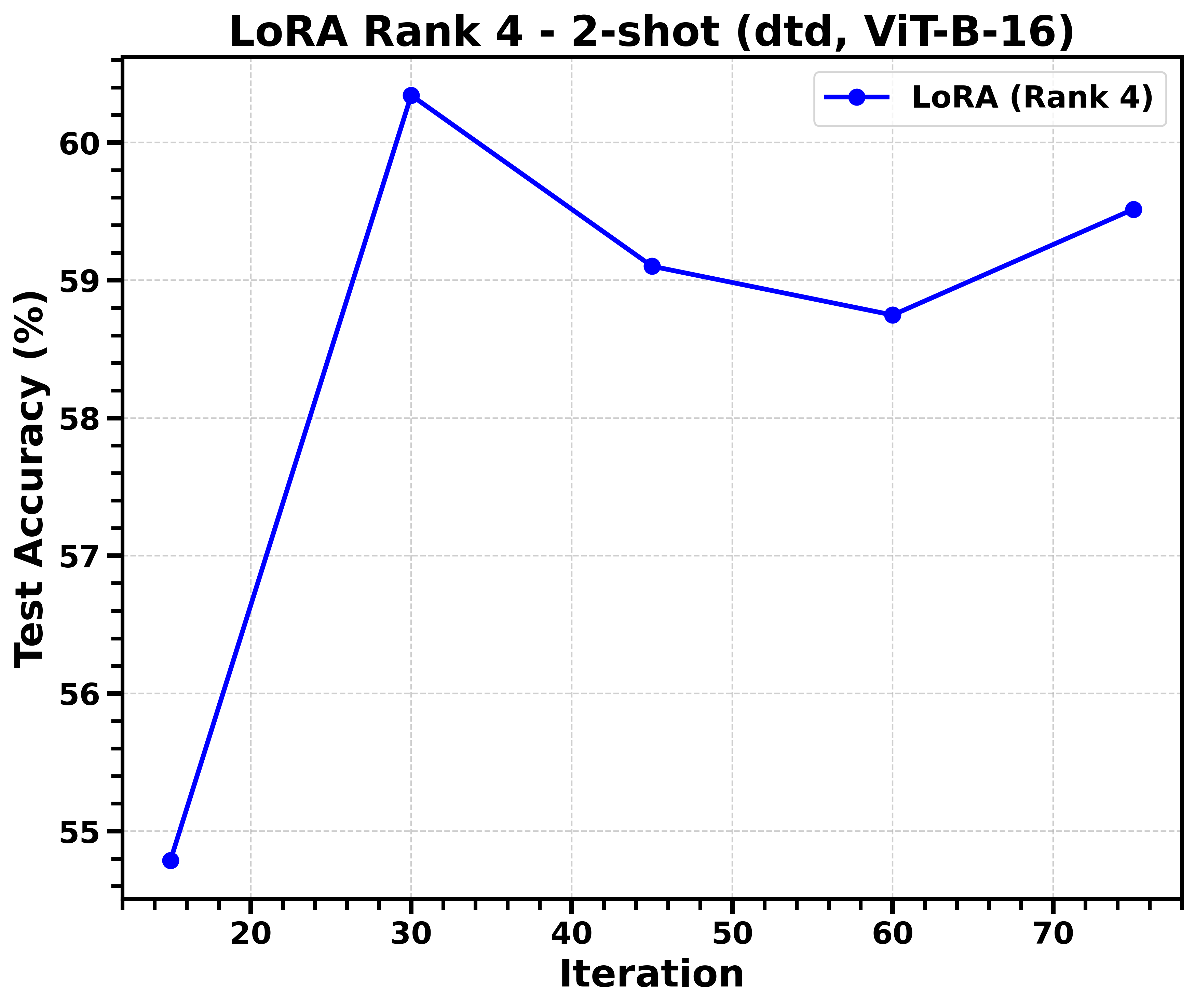}
    \end{subfigure} &
    \begin{subfigure}[b]{0.22\textwidth}
        \centering
        \includegraphics[width=\linewidth]{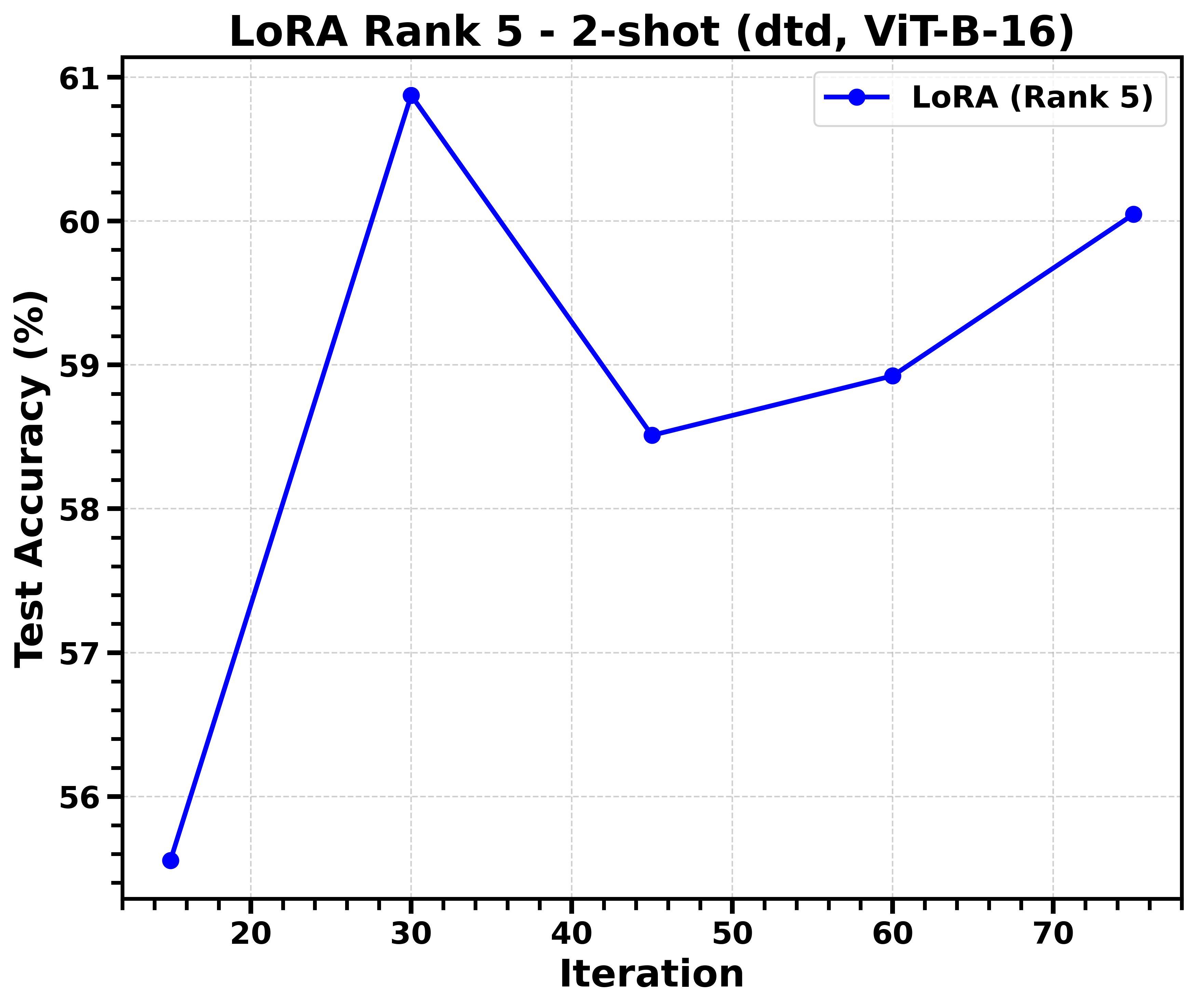}
    \end{subfigure} \\

    \rotatebox{90}{\usebox2} & 
    \begin{subfigure}[b]{0.22\textwidth}
        \centering
        \includegraphics[width=\linewidth]{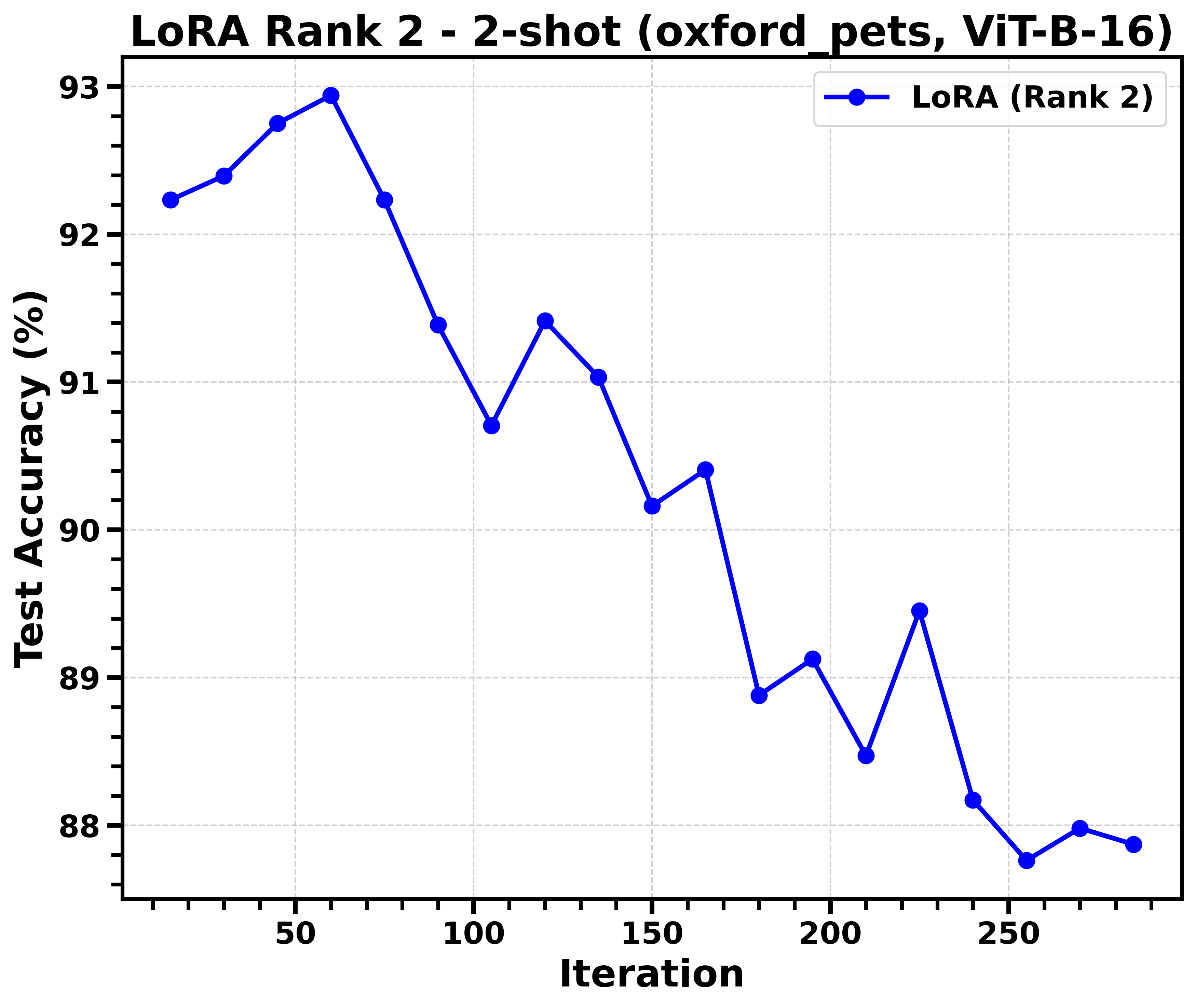}
    \end{subfigure} &
    \begin{subfigure}[b]{0.22\textwidth}
        \centering
        \includegraphics[width=\linewidth]{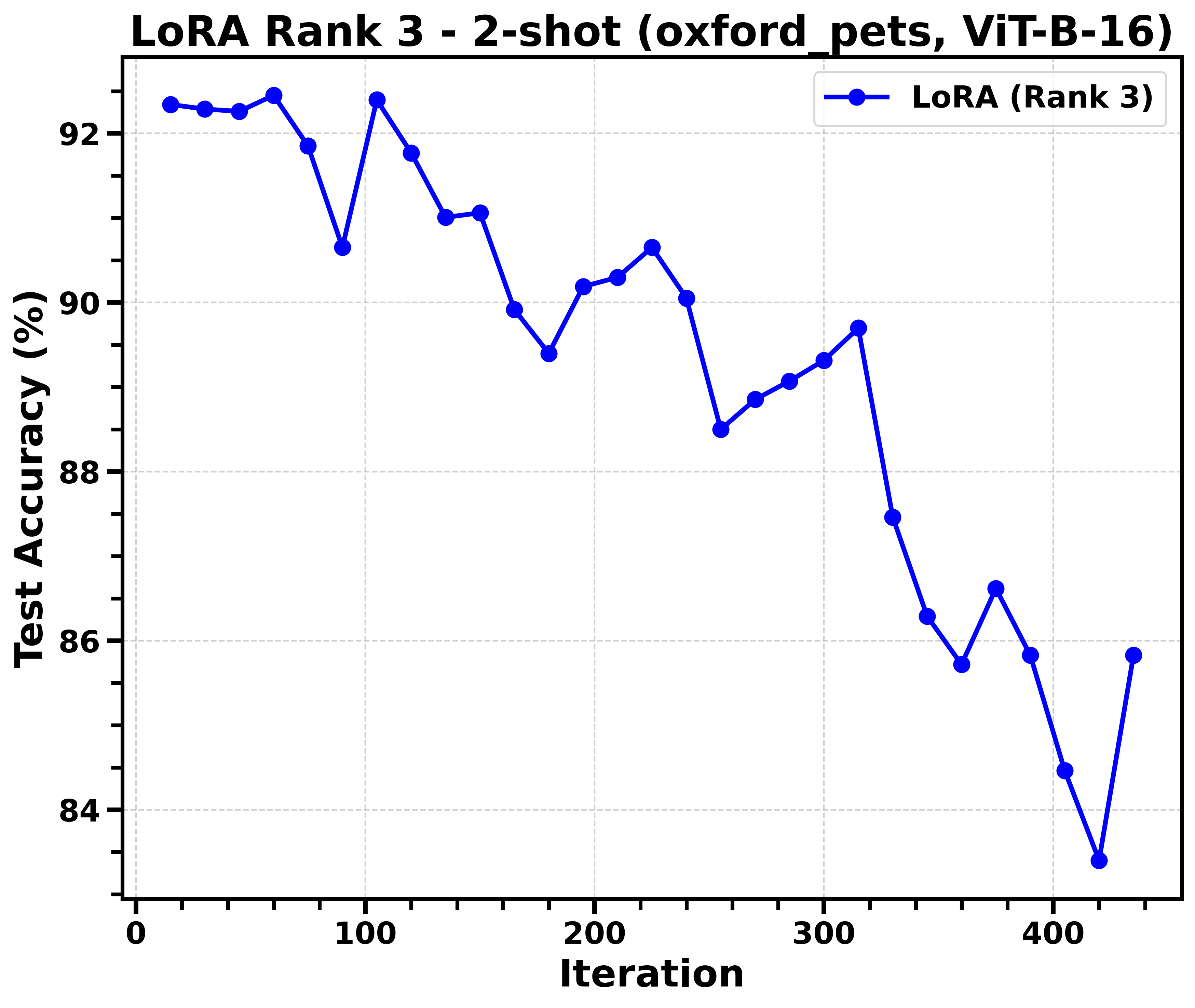}
    \end{subfigure} &
    \begin{subfigure}[b]{0.22\textwidth}
        \centering
        \includegraphics[width=\linewidth]{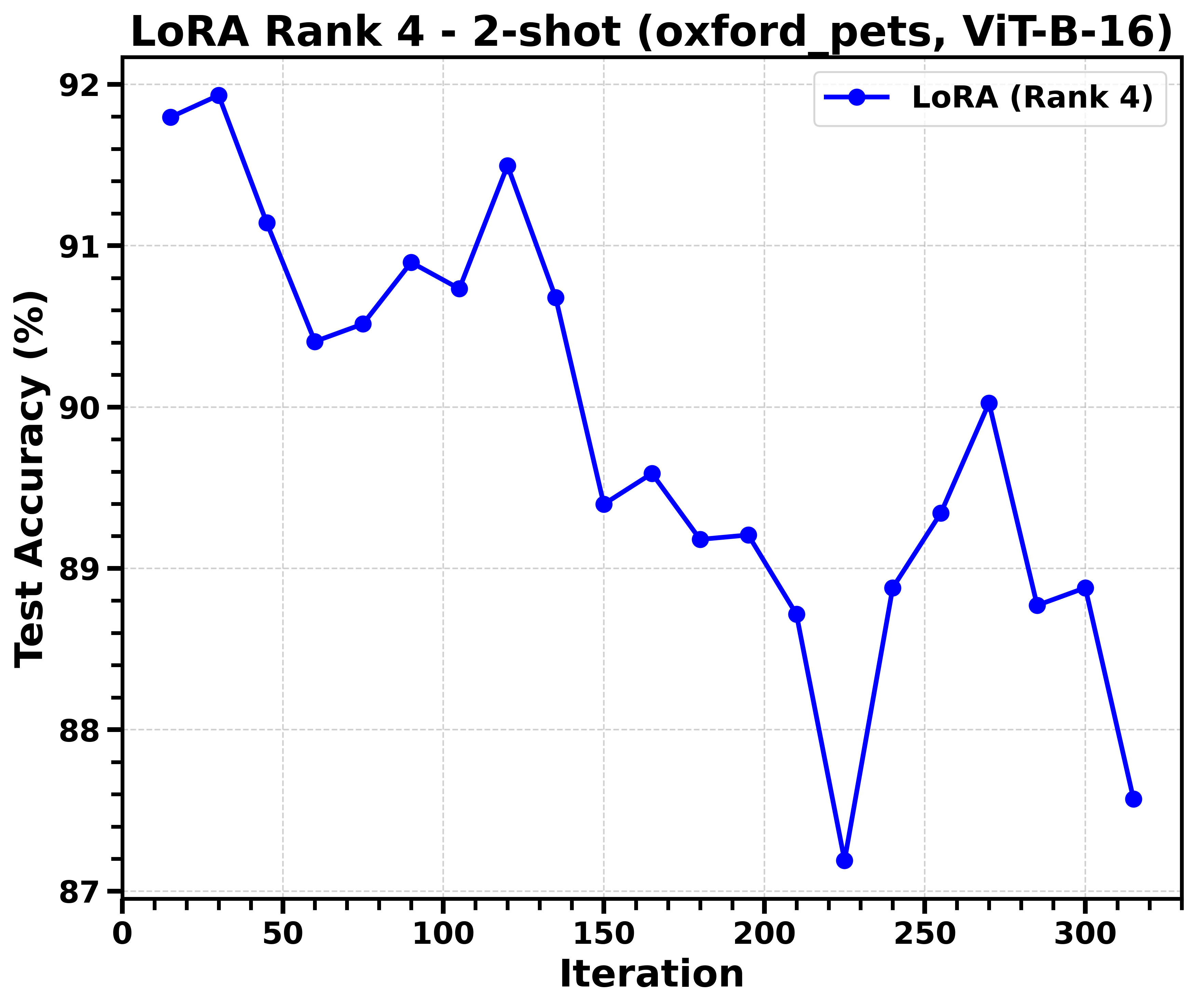}
    \end{subfigure} &
    \begin{subfigure}[b]{0.22\textwidth}
        \centering
        \includegraphics[width=\linewidth]{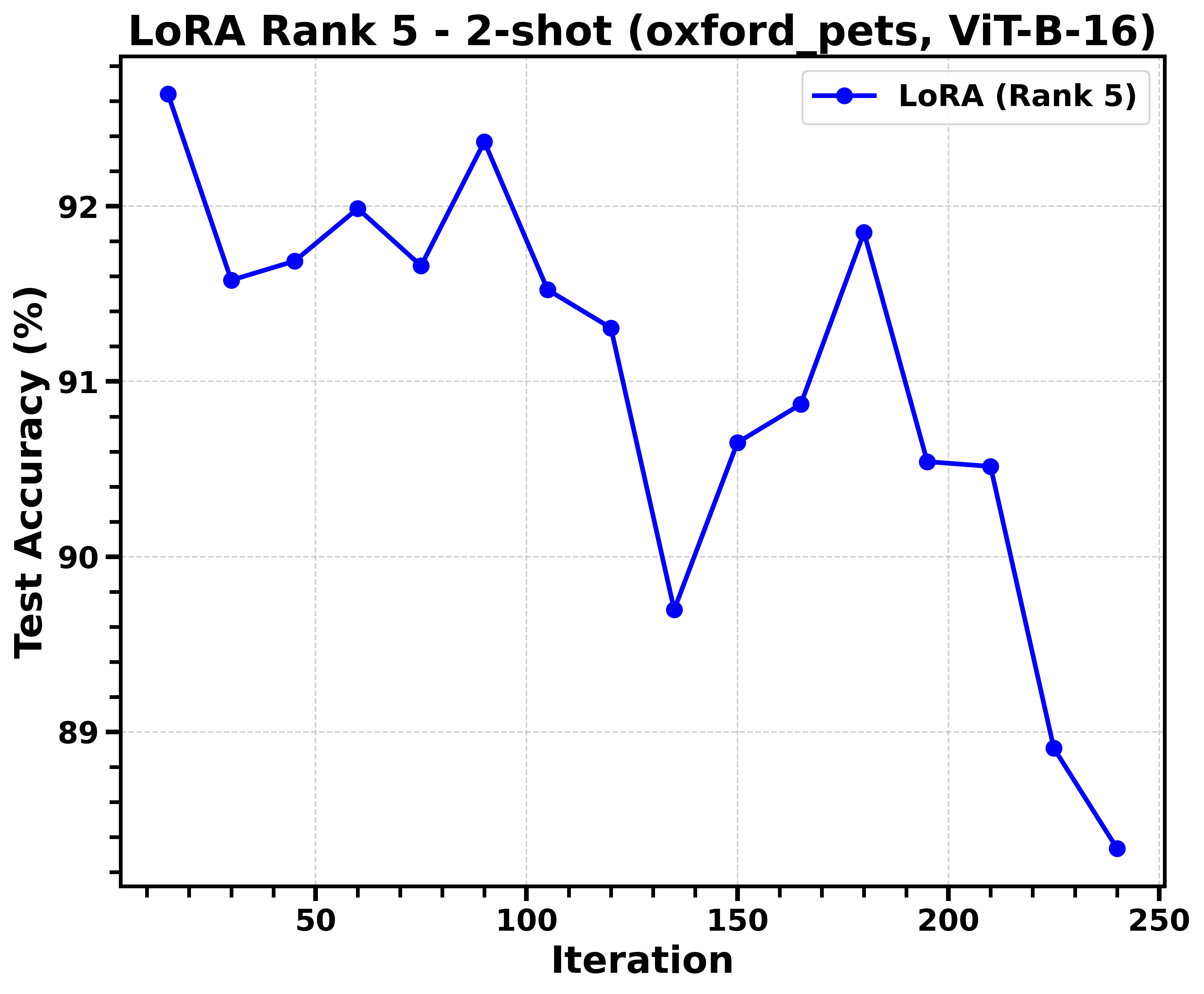}
    \end{subfigure} \\

    \rotatebox{90}{\usebox3} & 
    \begin{subfigure}[b]{0.22\textwidth}
        \centering
        \includegraphics[width=\linewidth]{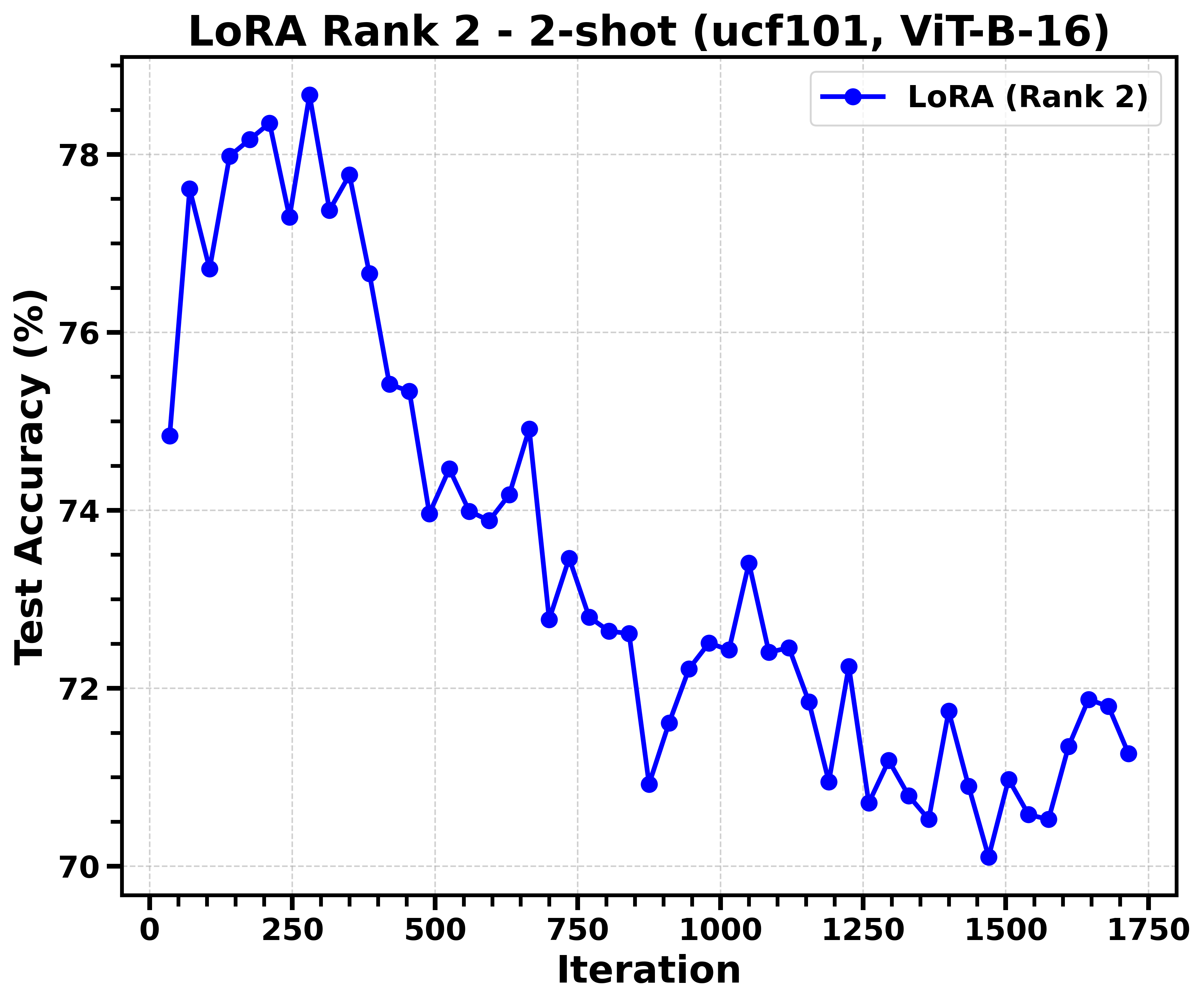}
    \end{subfigure} &
    \begin{subfigure}[b]{0.22\textwidth}
        \centering
        \includegraphics[width=\linewidth]{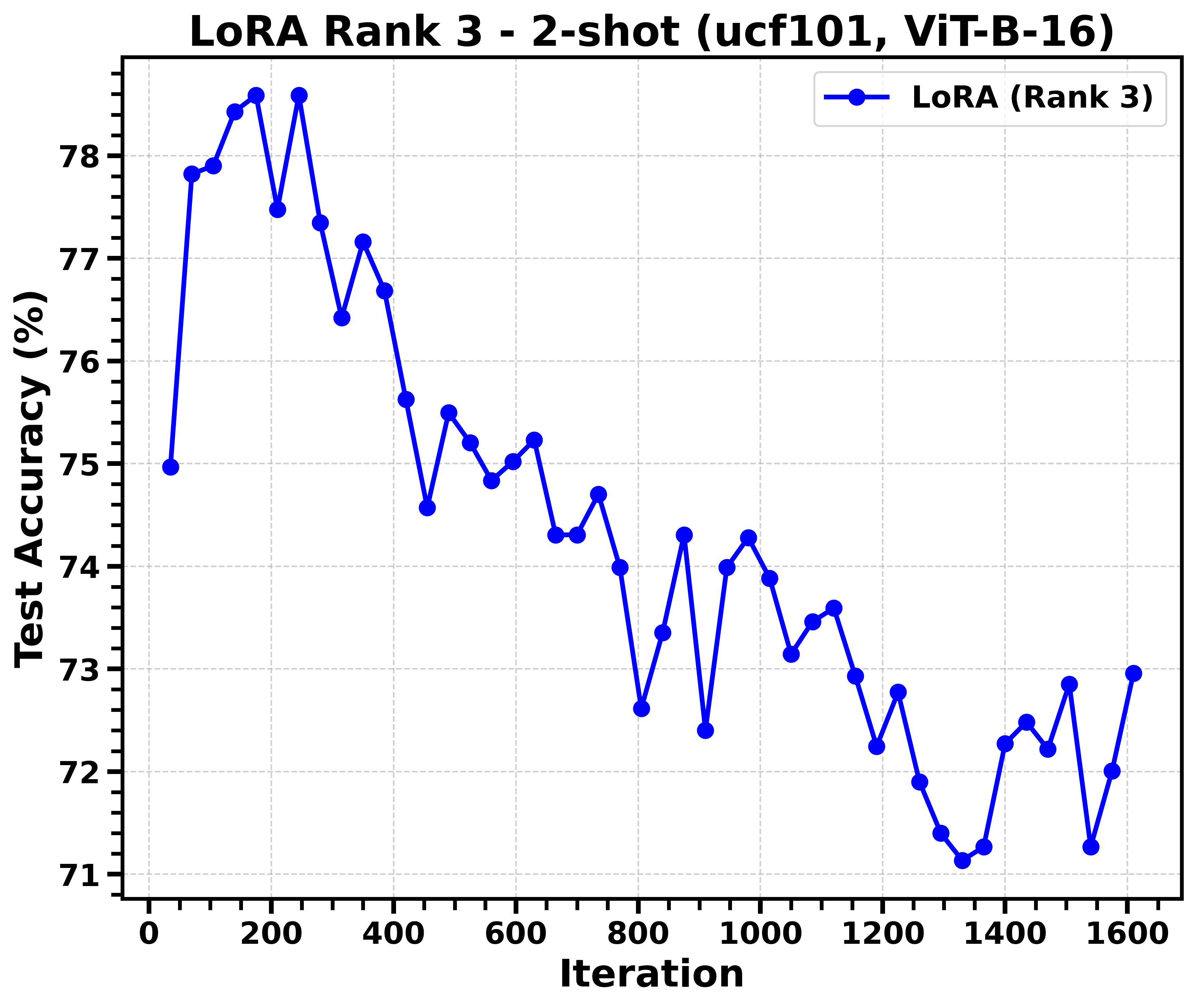}
    \end{subfigure} &
    \begin{subfigure}[b]{0.22\textwidth}
        \centering
        \includegraphics[width=\linewidth]{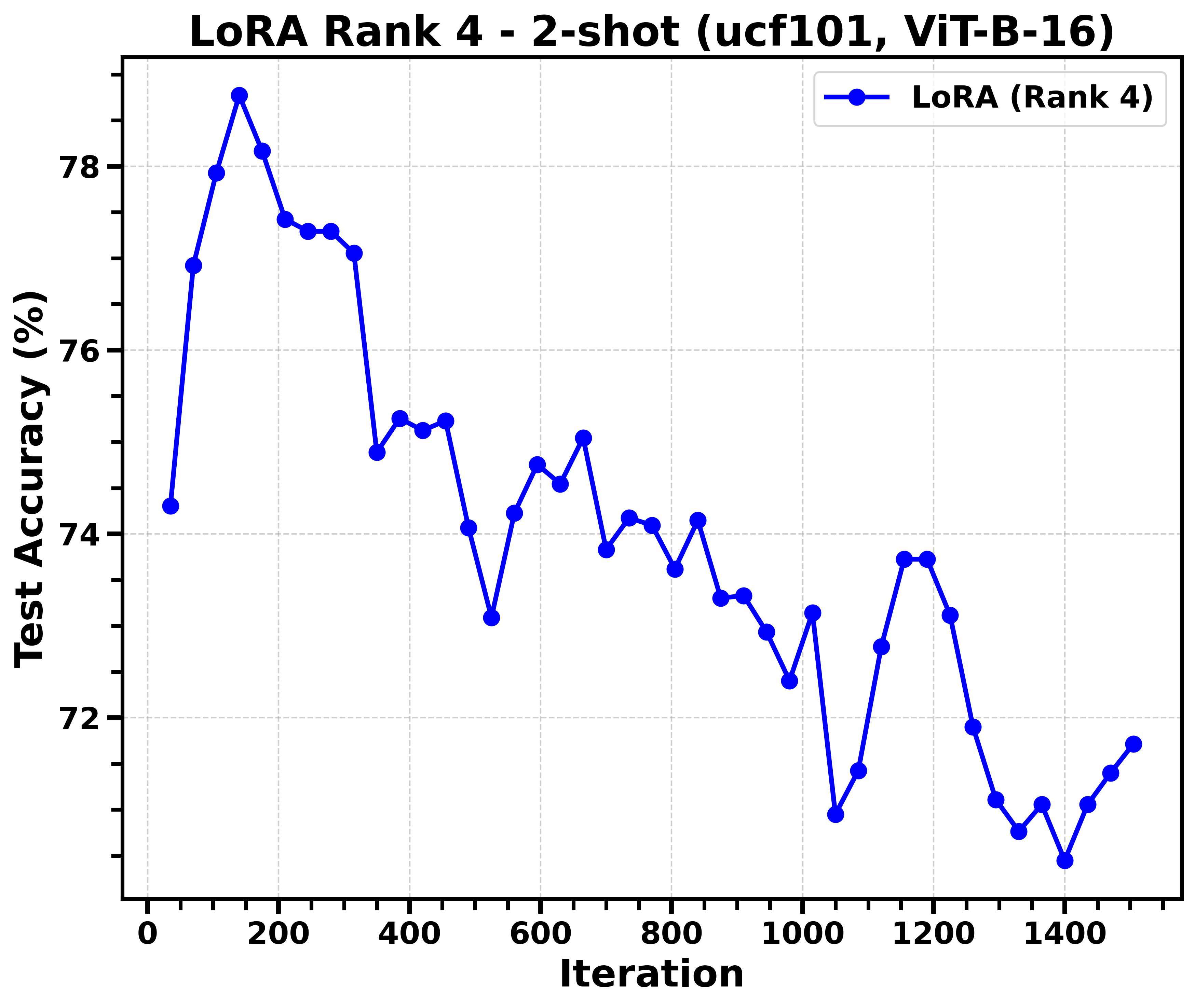}
    \end{subfigure} &
    \begin{subfigure}[b]{0.22\textwidth}
        \centering
        \includegraphics[width=\linewidth]{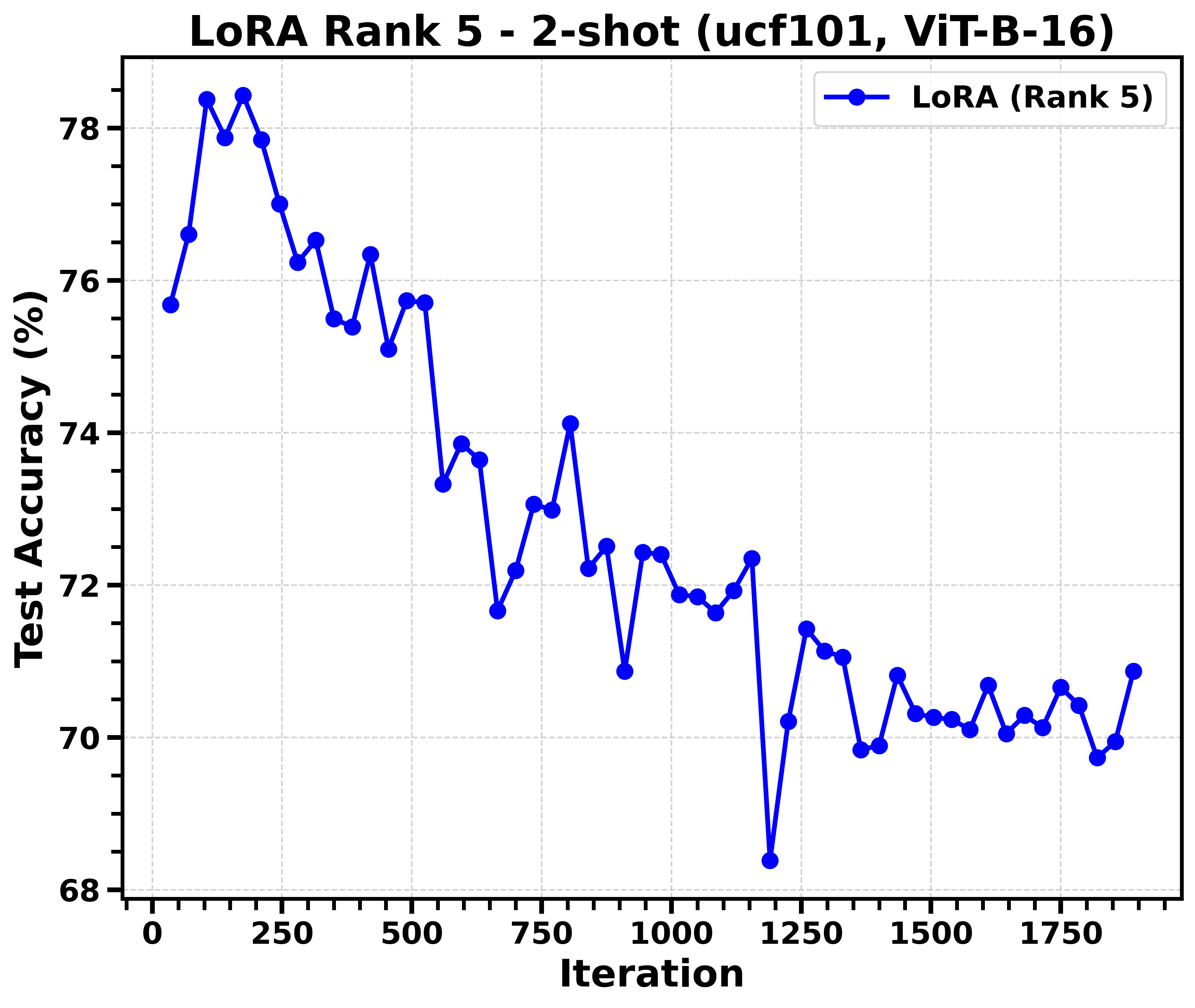}
    \end{subfigure} \\

\end{tabular}

\caption{LoRA performance in a 2-shot setting on three datasets—DTD, Oxford Pets, and UCF101— using a pretrained CLIP with ViT-B/16 backbone. The model is trained for at most 2000 iterations or until the loss \(\le 0.01\).}
\label{fig:matrix_of_figures_2shots}
\end{figure*}

\begin{figure*}
\centering
\renewcommand{\arraystretch}{1.3} 

\begin{tabular}{c@{\hskip 10pt} c c c c} 
    & Rank 2 & Rank 3 & Rank 4 & Rank 5 \\

    \rotatebox{90}{\usebox1} & 
    \begin{subfigure}[b]{0.22\textwidth}
        \centering
        \includegraphics[width=\linewidth]{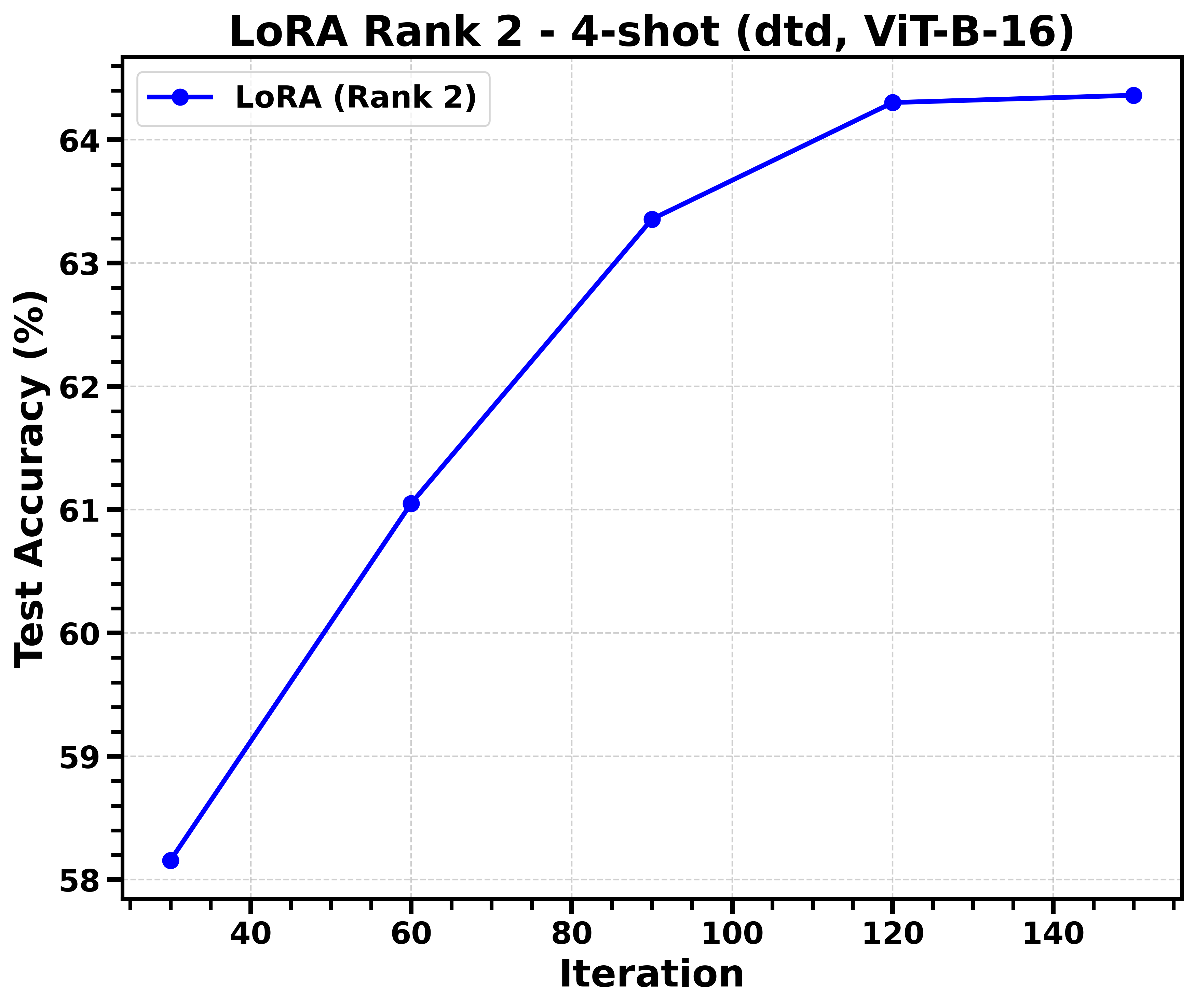}
    \end{subfigure} &
    \begin{subfigure}[b]{0.22\textwidth}
        \centering
        \includegraphics[width=\linewidth]{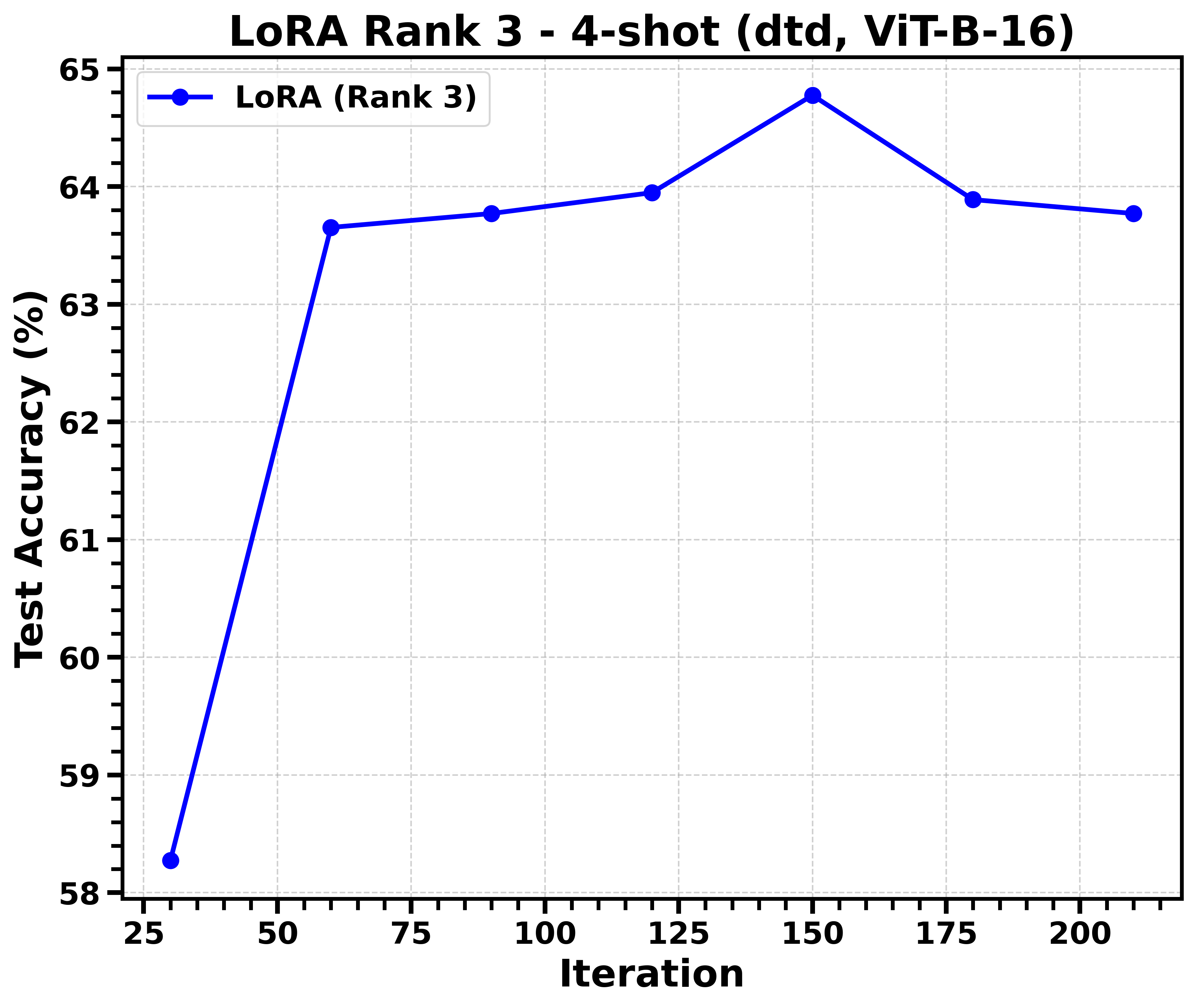}
    \end{subfigure} &
    \begin{subfigure}[b]{0.22\textwidth}
        \centering
        \includegraphics[width=\linewidth]{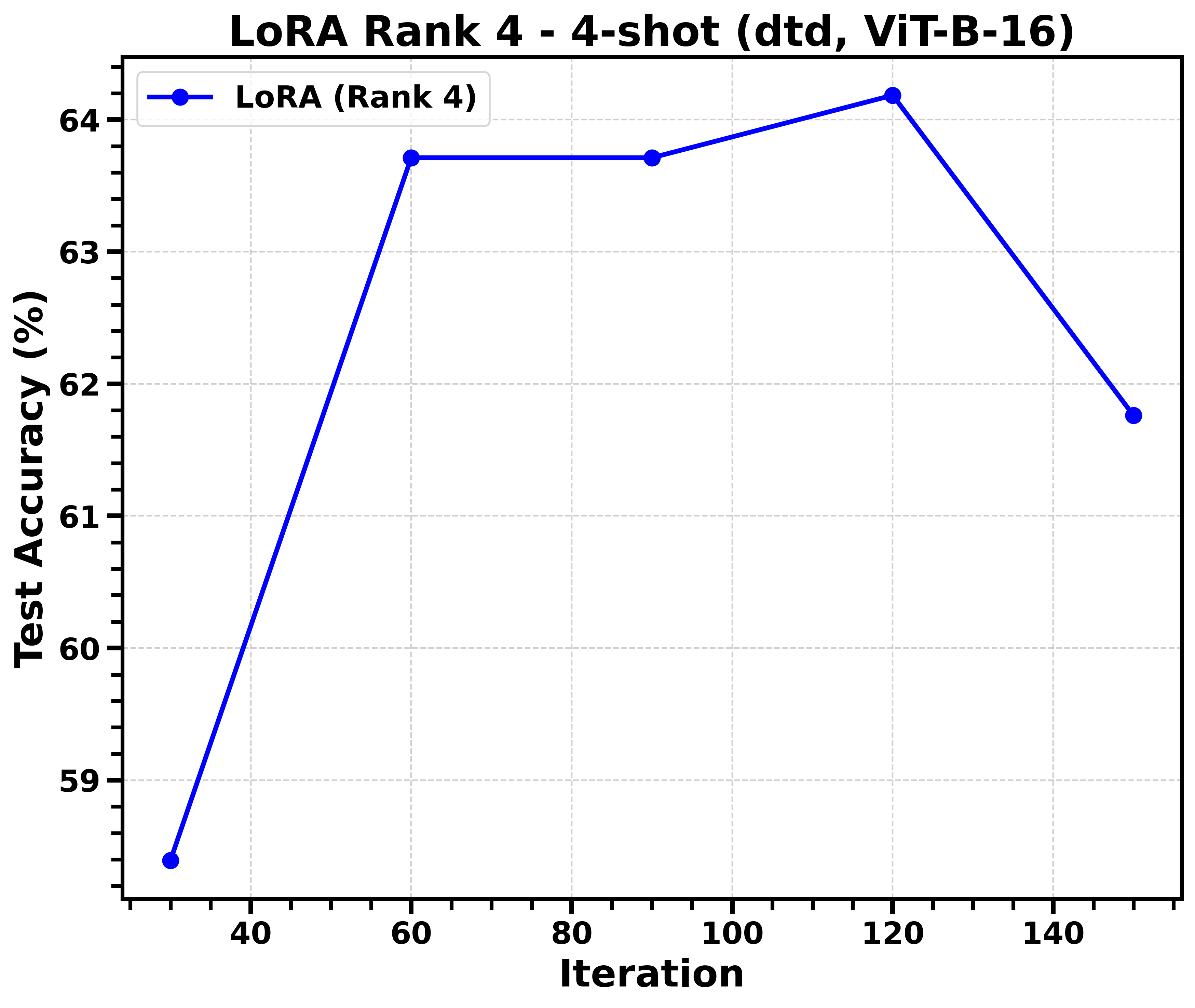}
    \end{subfigure} &
    \begin{subfigure}[b]{0.22\textwidth}
        \centering
        \includegraphics[width=\linewidth]{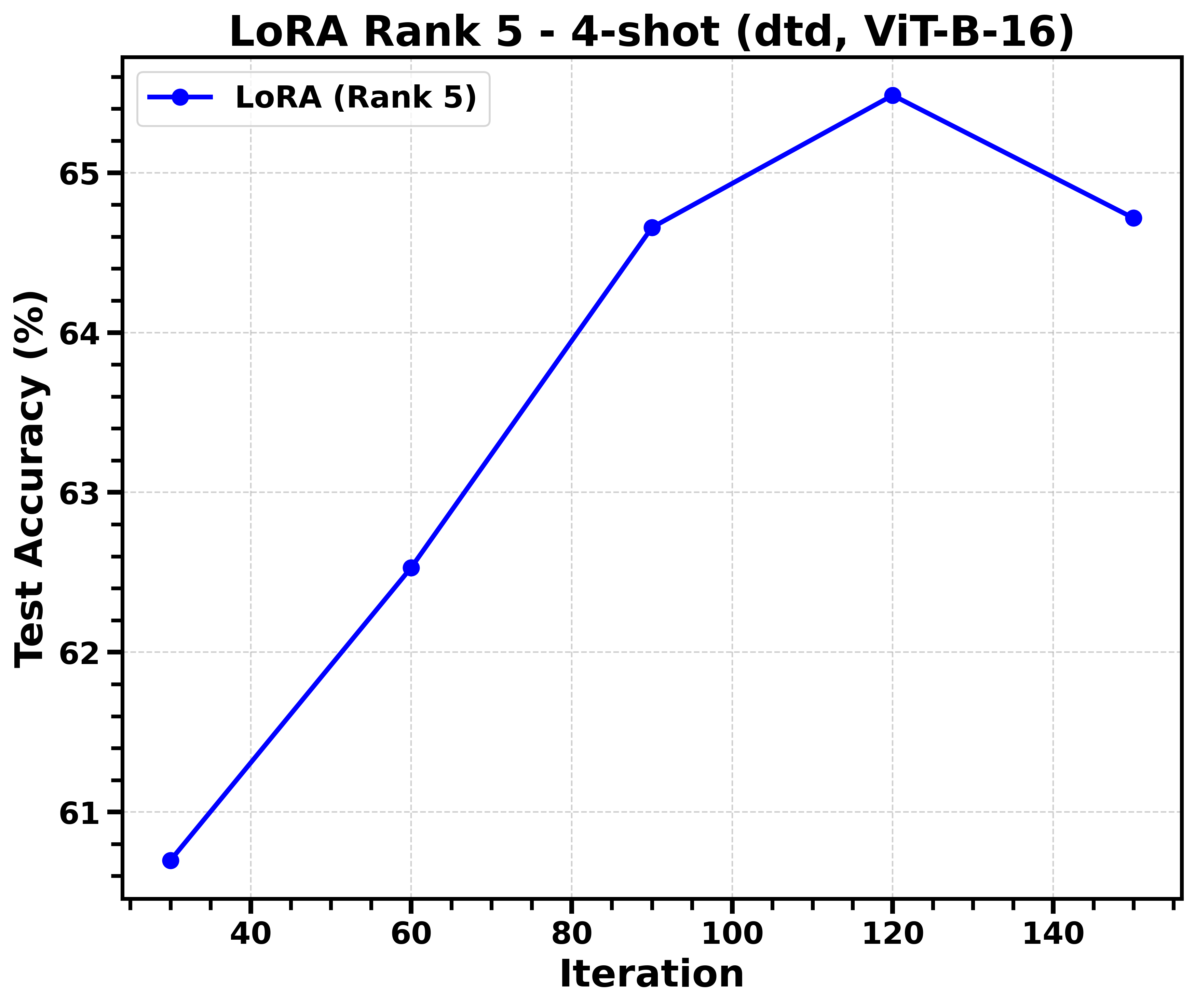}
    \end{subfigure} \\

    \rotatebox{90}{\usebox2} & 
    \begin{subfigure}[b]{0.22\textwidth}
        \centering
        \includegraphics[width=\linewidth]{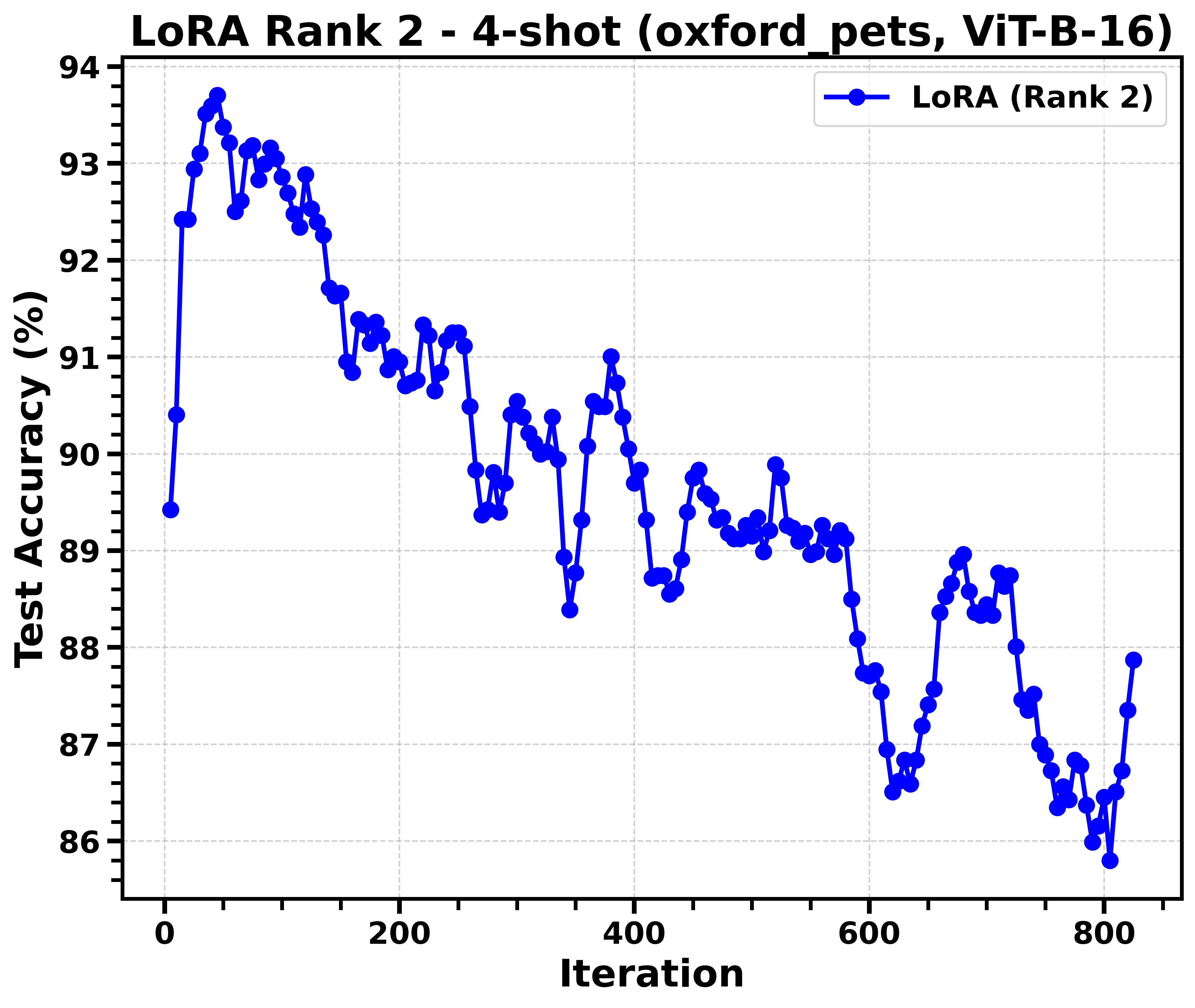}
    \end{subfigure} &
    \begin{subfigure}[b]{0.22\textwidth}
        \centering
        \includegraphics[width=\linewidth]{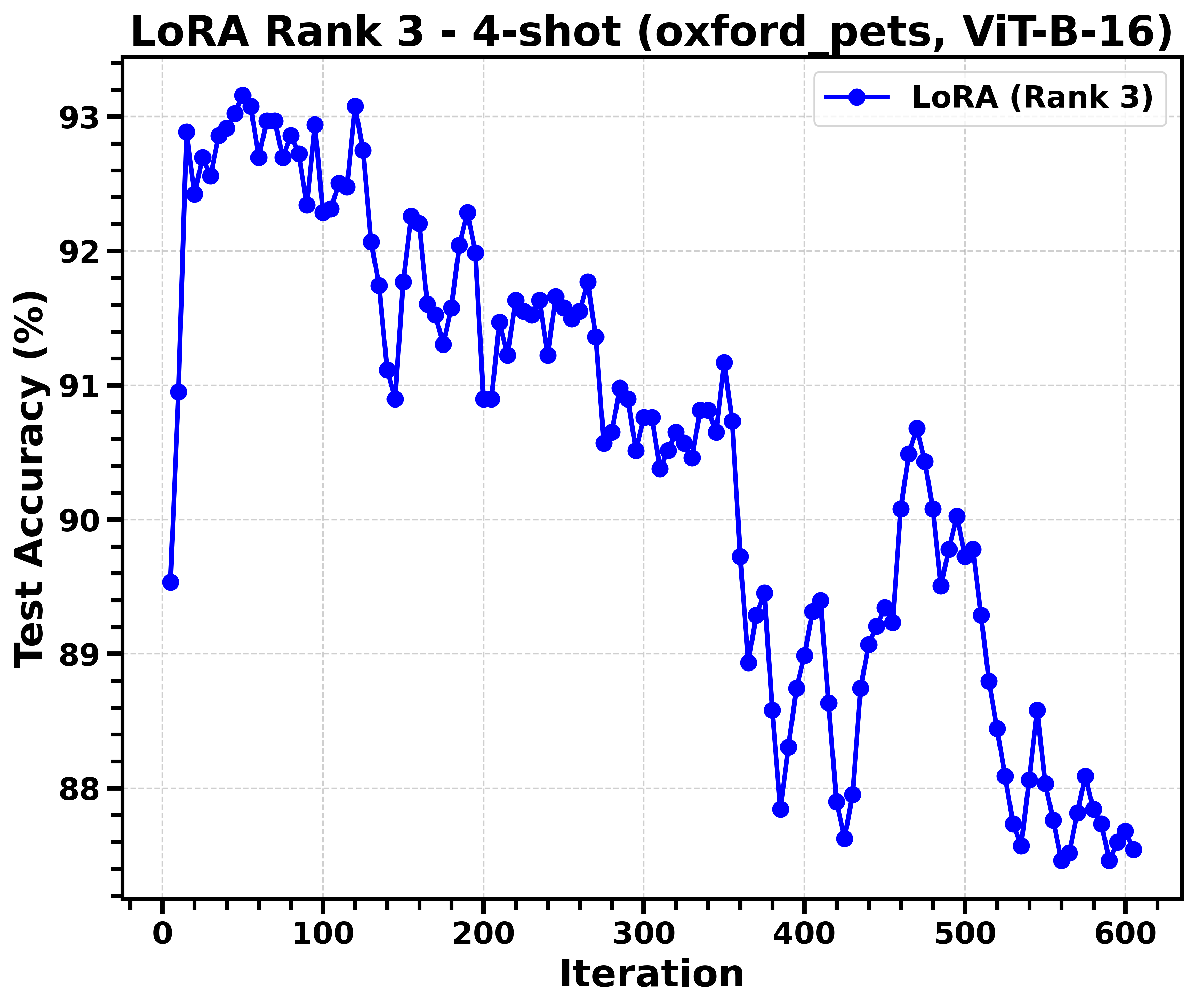}
    \end{subfigure} &
    \begin{subfigure}[b]{0.22\textwidth}
        \centering
        \includegraphics[width=\linewidth]{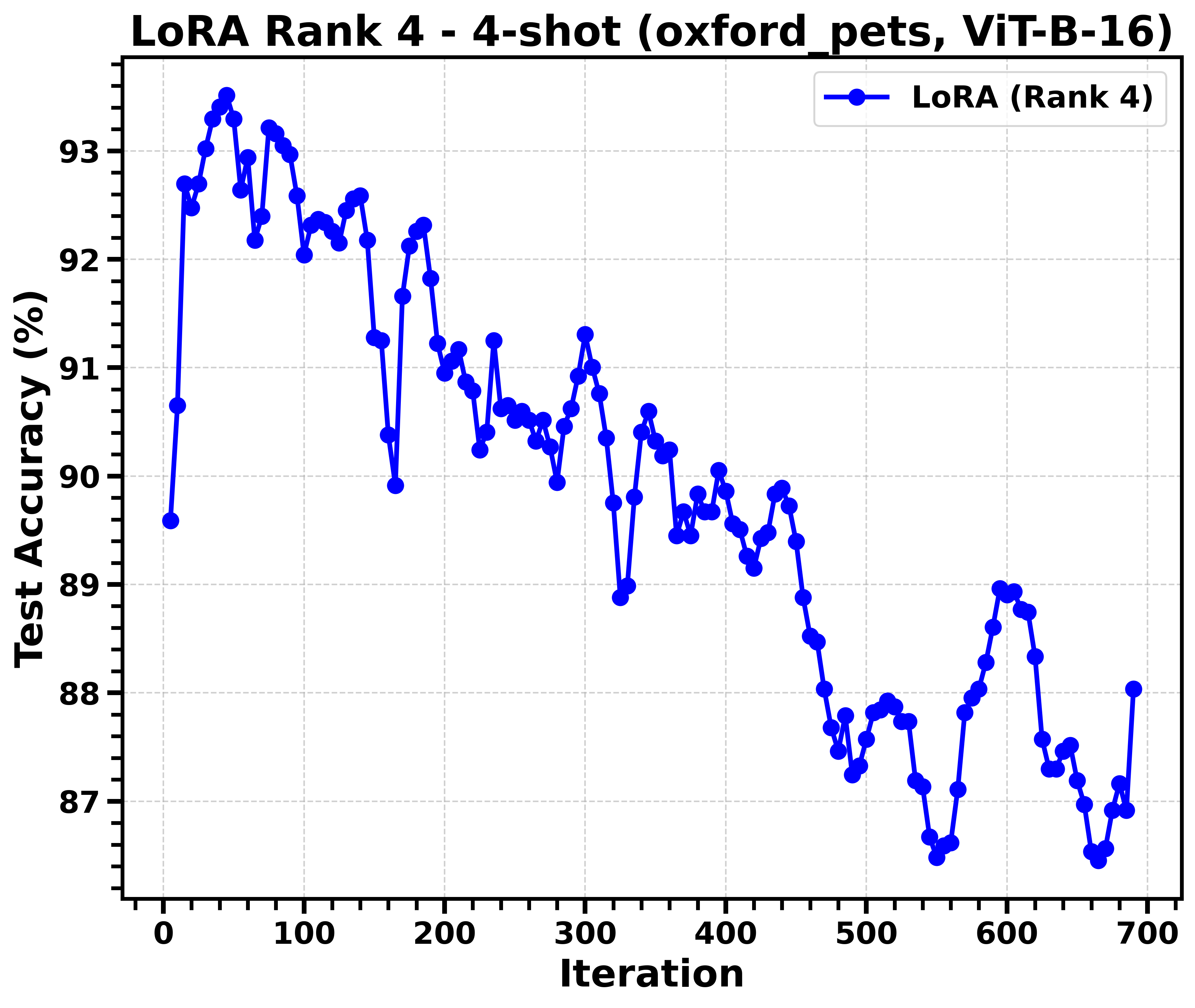}
    \end{subfigure} &
    \begin{subfigure}[b]{0.22\textwidth}
        \centering
        \includegraphics[width=\linewidth]{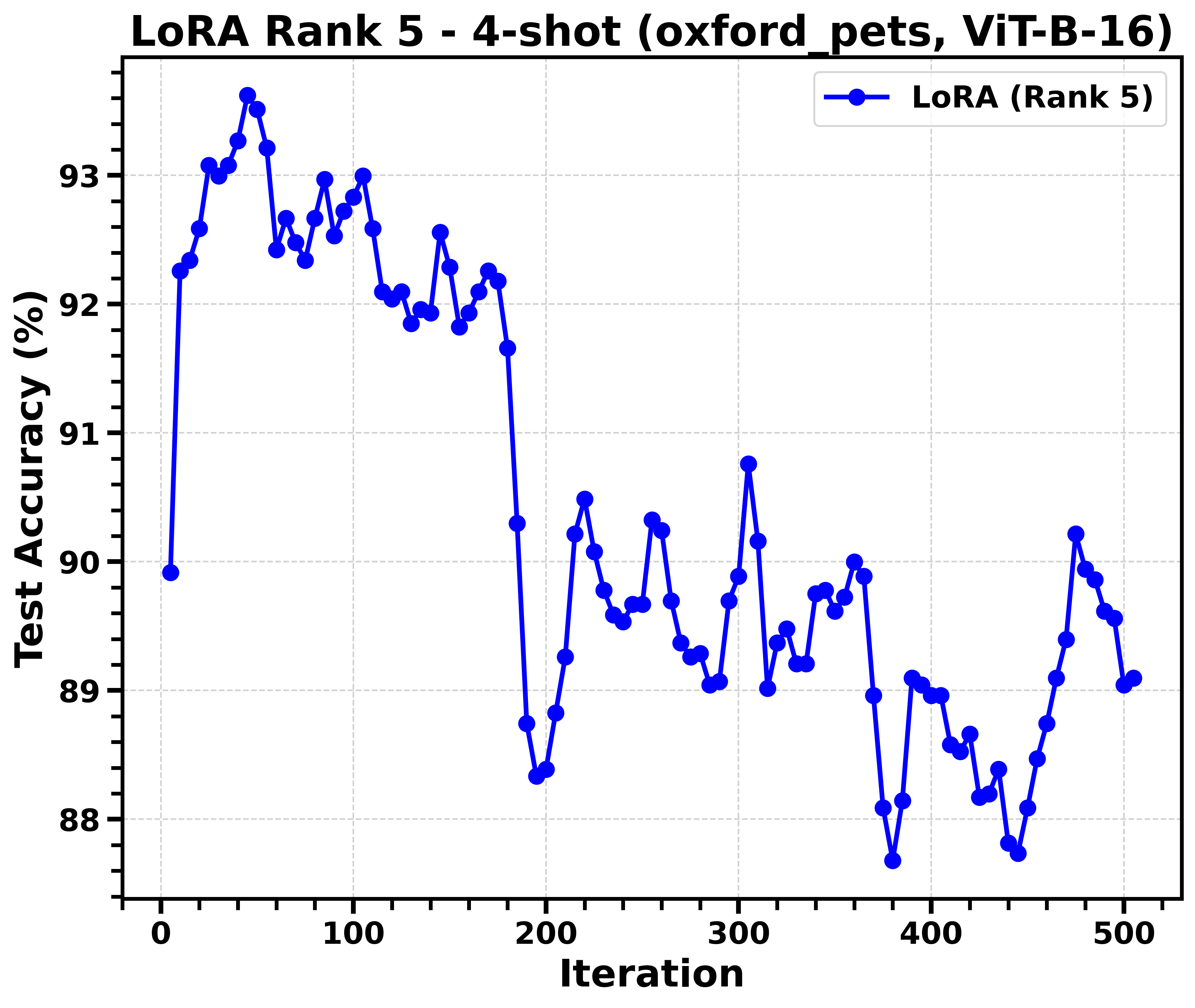}
    \end{subfigure} \\

    \rotatebox{90}{\usebox3} & 
    \begin{subfigure}[b]{0.22\textwidth}
        \centering
        \includegraphics[width=\linewidth]{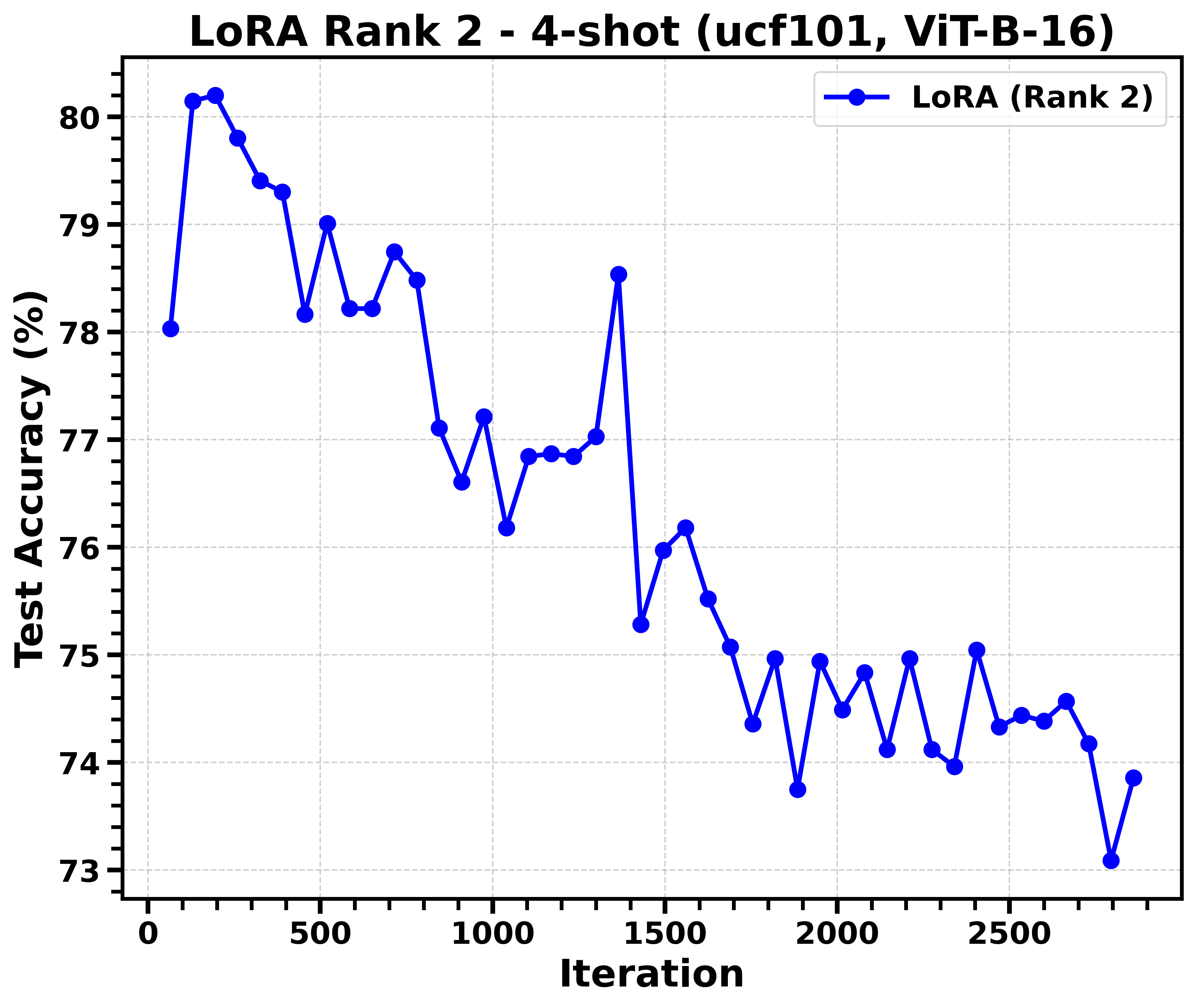}
    \end{subfigure} &
    \begin{subfigure}[b]{0.22\textwidth}
        \centering
        \includegraphics[width=\linewidth]{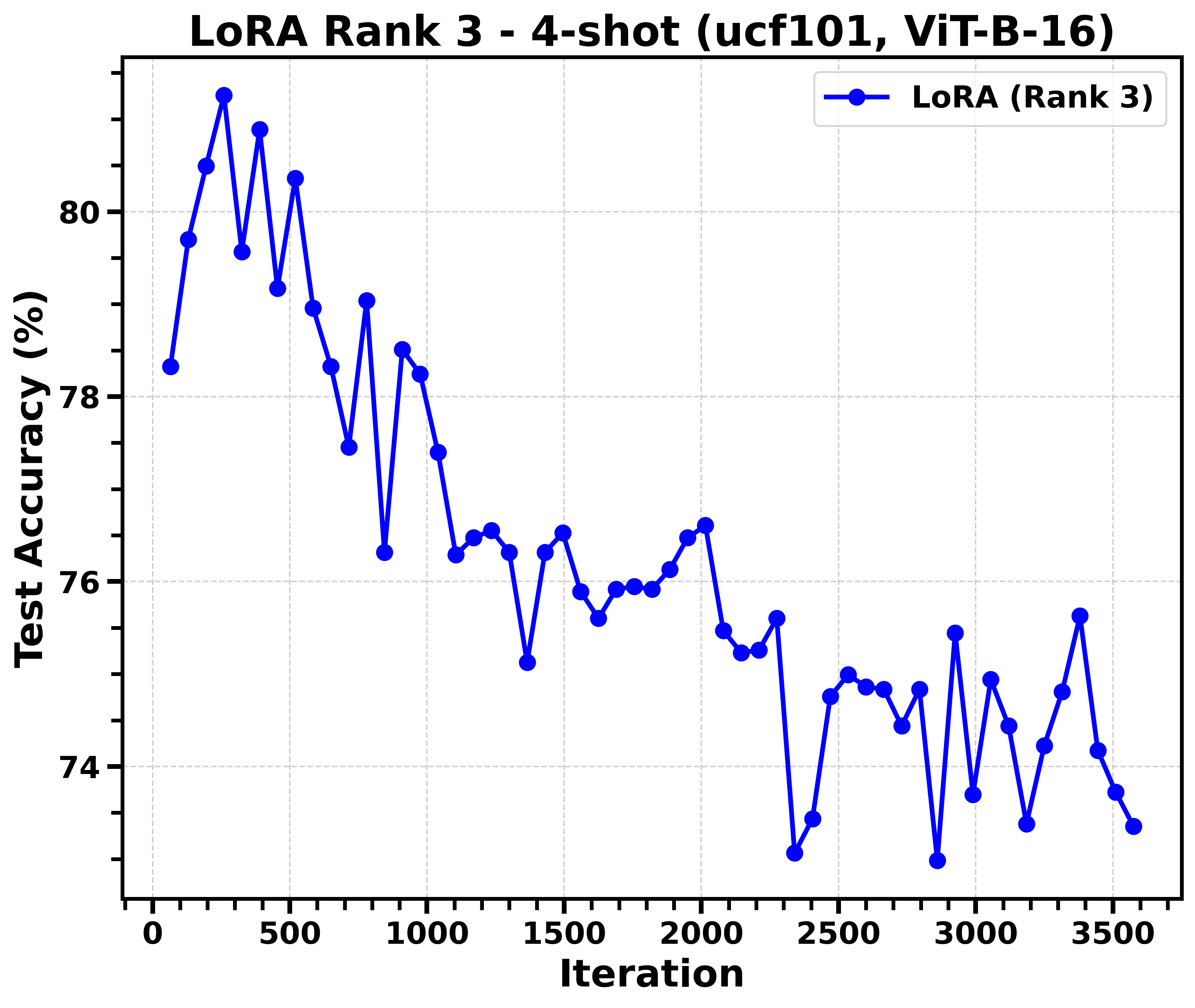}
    \end{subfigure} &
    \begin{subfigure}[b]{0.22\textwidth}
        \centering
        \includegraphics[width=\linewidth]{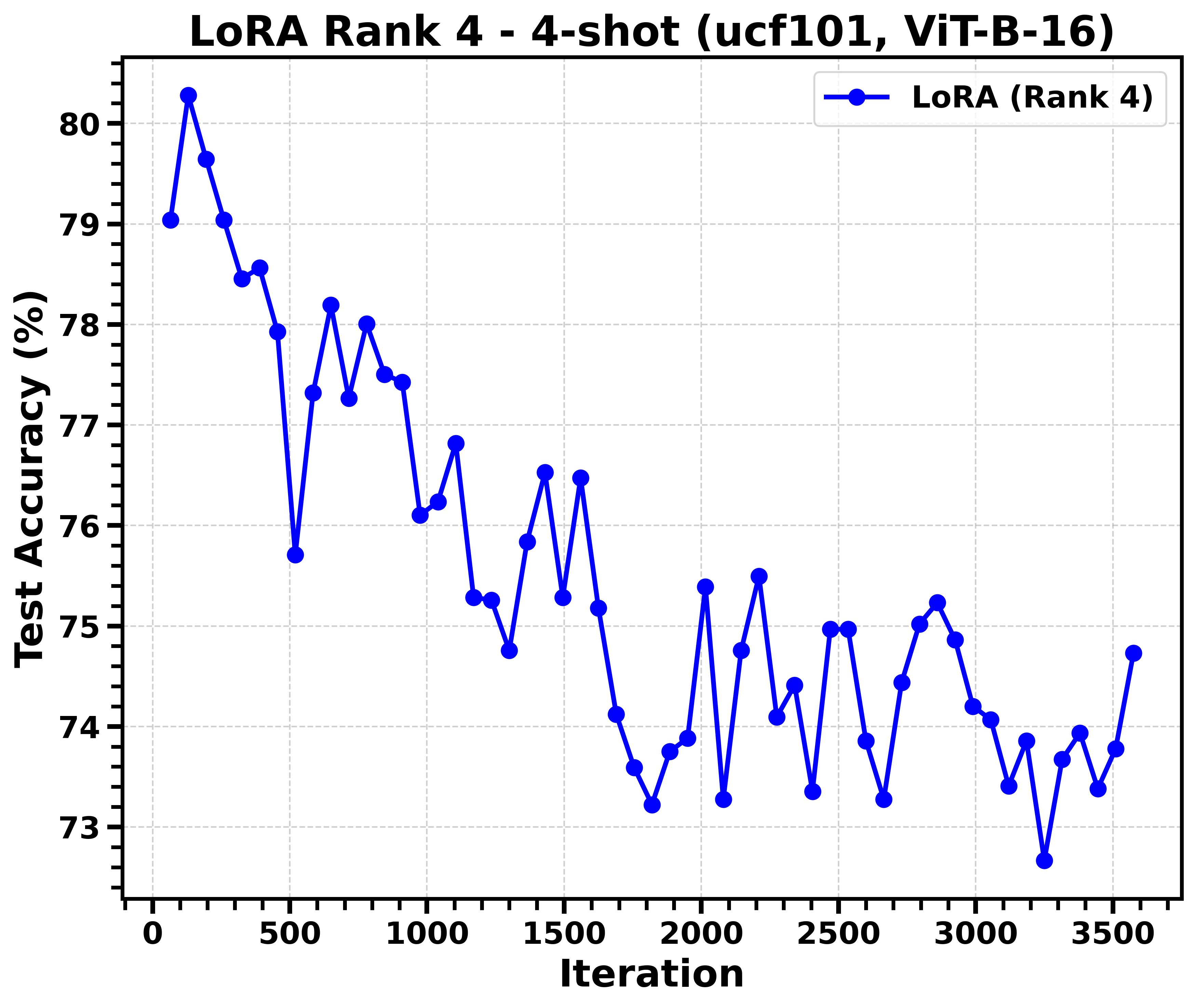}
    \end{subfigure} &
    \begin{subfigure}[b]{0.22\textwidth}
        \centering
        \includegraphics[width=\linewidth]{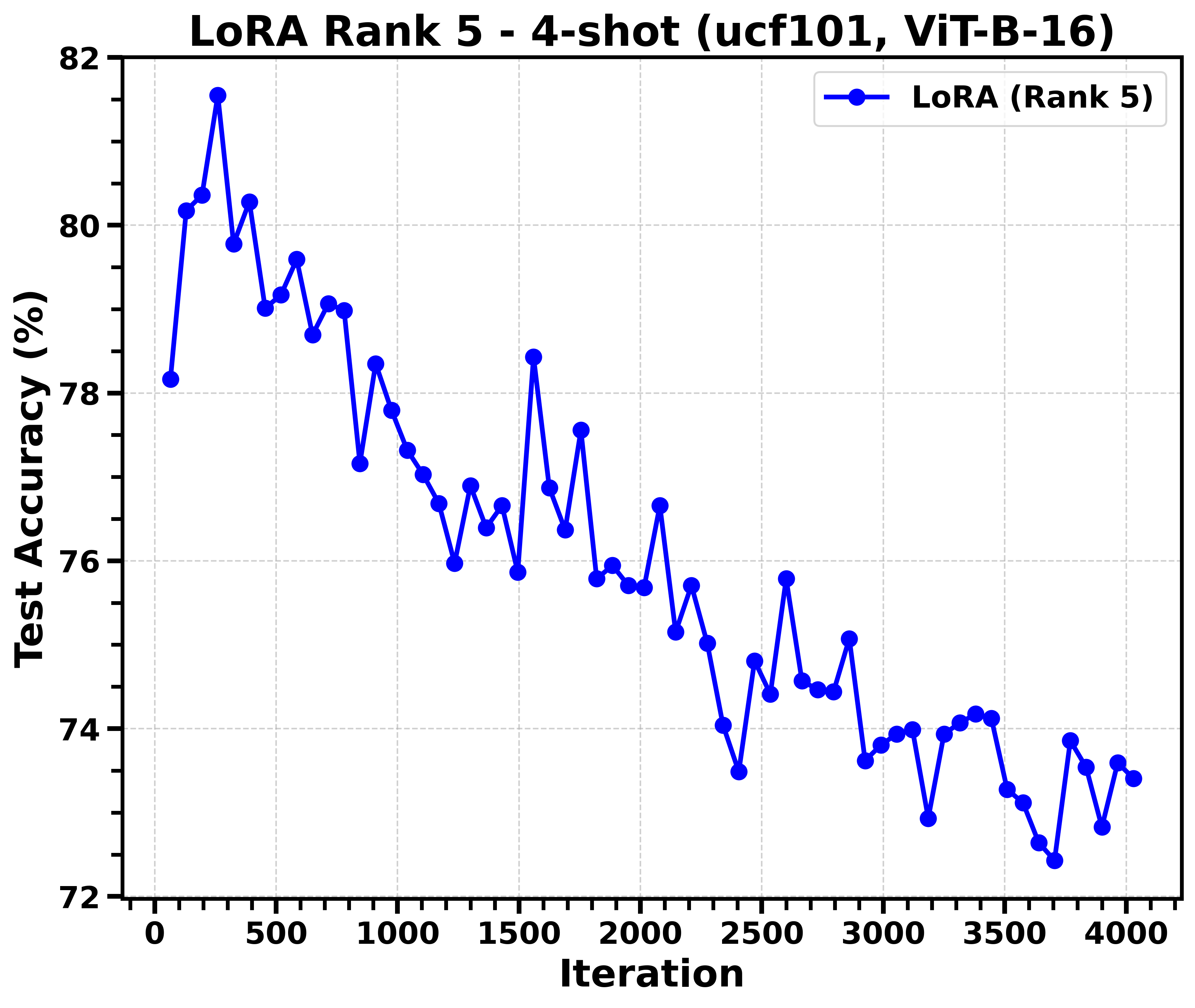}
    \end{subfigure} \\
\end{tabular}

\caption{LoRA performance in a 4-shot setting on three datasets—DTD, Oxford Pets, and UCF101— using a pretrained CLIP with ViT-B/16 backbone. The model is trained for at most 2000 iterations or until the loss \(\le 0.01\).}
\label{fig:matrix_of_figures_4shots}
\end{figure*}

Figures~\ref{fig:matrix_of_figures_2shots} and~\ref{fig:matrix_of_figures_4shots} illustrate LoRA’s performance under few-shot settings, specifically with 2-shot and 4-shot. We finetune CLIP with a ViT-B/16 backbone on three benchmark datasets—DTD, Oxford Pets, and UCF101—and vary the LoRA rank in \(\{2,3,4,5\}\). We set the LoRA rank in \(\{2,3,4,5\}\) and train for a maximum of 2000 iterations or until the training loss reaches or falls below 0.01, whichever occurs first. At each iteration, we measure test-set accuracy to assess overfitting and convergence.

First, LoRA generally shows an improvement in test accuracy at the beginning of the training process, but results deteriorate noticeably as training proceeds. For instance, in the 2-shot Pets case at rank 2, accuracy peaks around $200$ iterations before declining by over $5\%$, indicating overfitting. In contrast, the 4-shot DTD setting at rank 4 peaks around $300$ iterations and then loses several points of accuracy once training proceeds. 

Second, the optimal rank differs by dataset and shot setting. While rank 2 seems sufficient for DTD, it is not associated with the best results in Pets or UCF101. The results of LoRA in 2-shot learning on UCF101 at rank 5 initially surpasses rank 2 but soon drop below it once the model overfits. These behaviors illustrate how peak accuracy depends strongly on the dataset, the number of shots, 
and the chosen rank. 

Finally, the oscillatory accuracy trends underscore LoRA’s sensitivity to both the rank parameter and the number of training iterations. Such fluctuations align with our main critique of LoRA. This method can be unstable in few-shot scenarios, which makes it difficult to choose a single hyperparameter setup that generalizes well across tasks.

\begin{figure}
    \centering
    \begin{subfigure}[b]{0.49\linewidth}
        \centering
        \includegraphics[width=\textwidth]{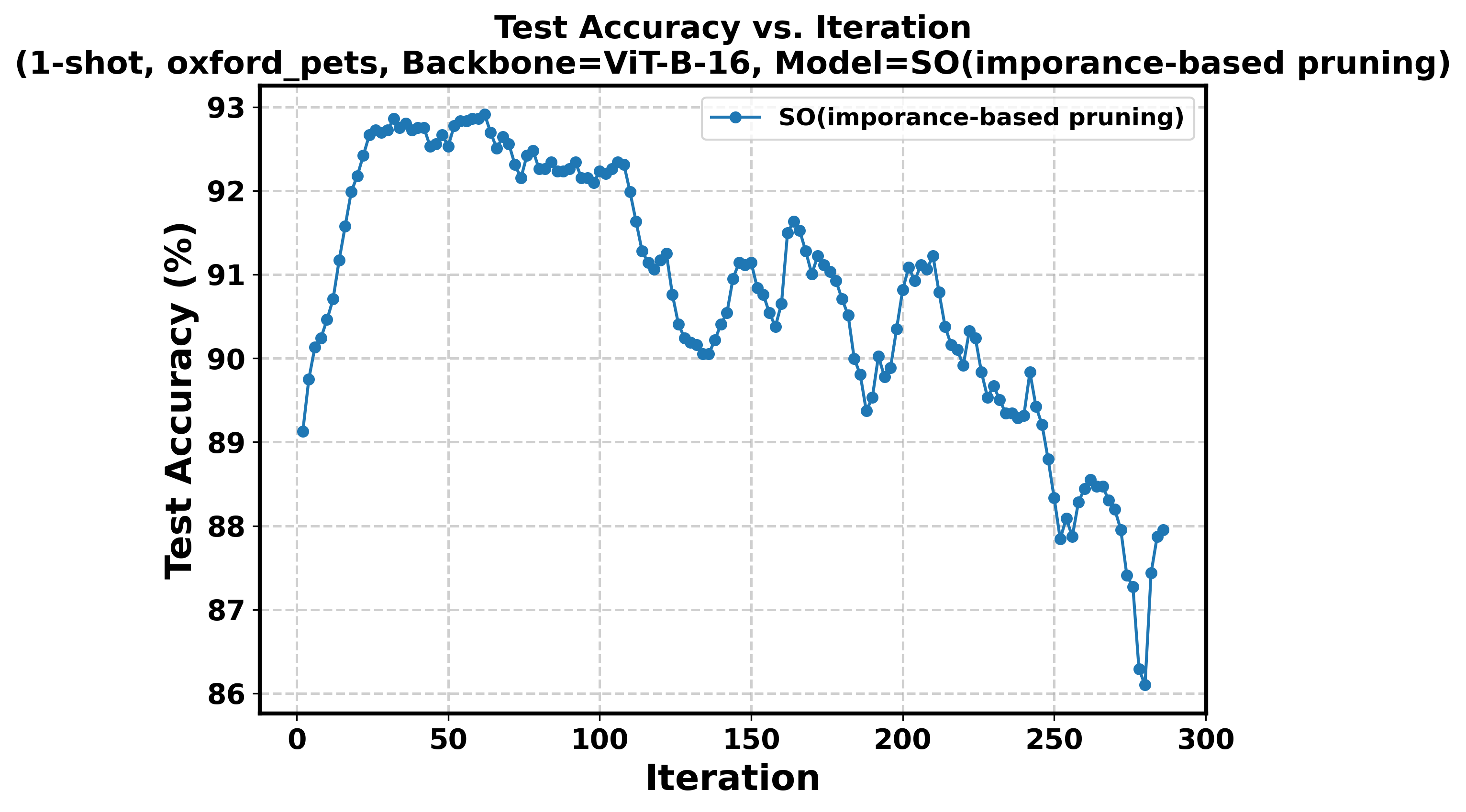}
        \caption{Pets}
        \label{fig:SIGIM_pets}
    \end{subfigure}
    \begin{subfigure}[b]{0.49\linewidth}
        \centering
        \includegraphics[width=\textwidth]{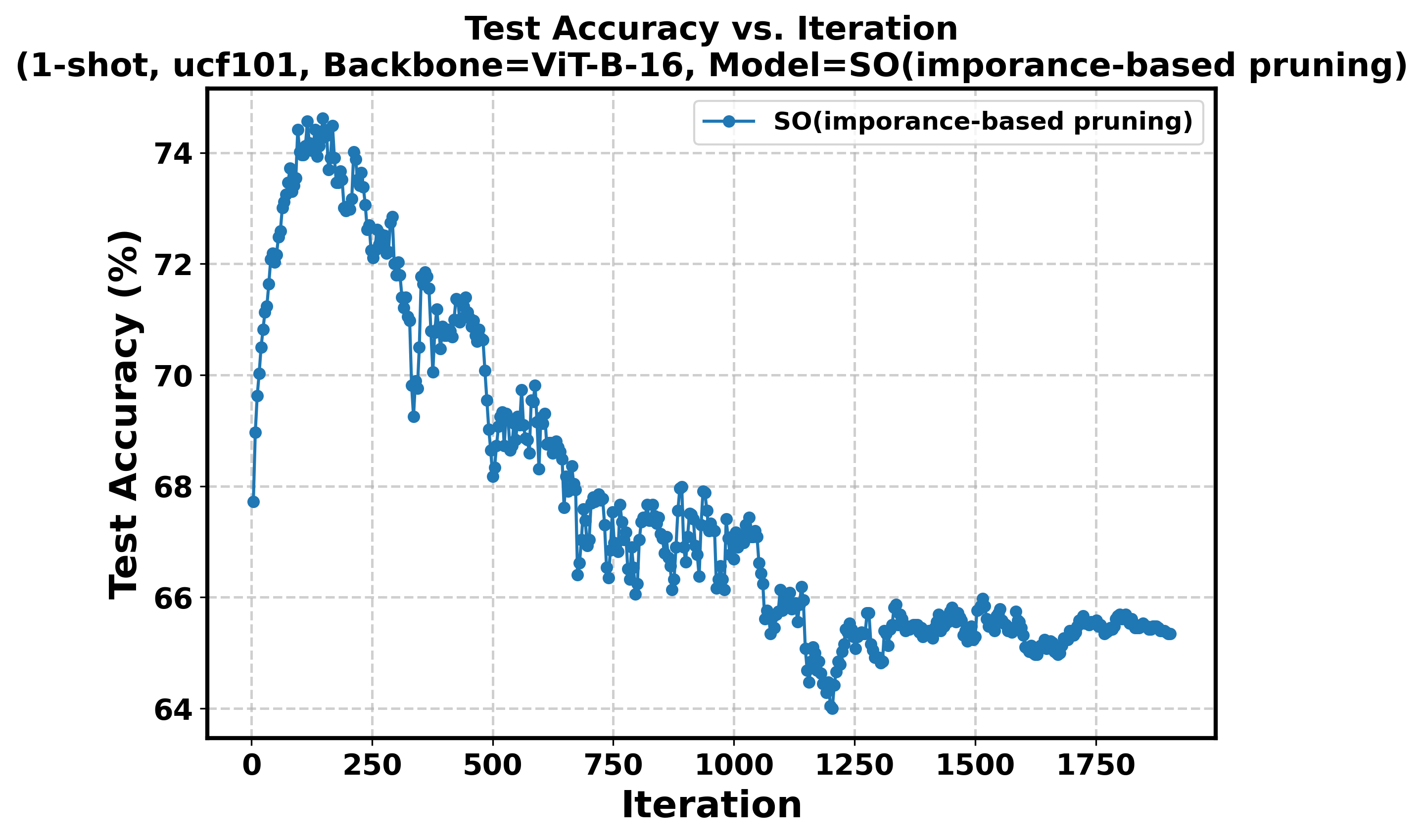}
        \caption{UCF}
        \label{fig:SIGIM_ucf}
    \end{subfigure}
    \caption{Test accuracy of SO (importance-based gradient pruning) with density ratio \(\kappa = 0.05\%\) and update interval \(T=10\), applied to a pretrained CLIP (ViT-B/16) backbone. We train on a 1-shot setting for Pets and UCF101 until the loss \(<0.01\) or a maximum of 2000 iterations.}
    \label{fig:SIGIM}
\end{figure}

Figures~\ref{fig:SIGIM_pets} and \ref{fig:SIGIM_ucf} illustrate SO’s test accuracy across training iterations when using importance-based gradient pruning for 1-shot adaptation of a pretrained CLIP (ViT-B/16). 
We set the density ratio to \(\kappa = 0.05\%\) and refresh the sparsity support every \(T=10\) iterations, training either until the loss falls below 0.01 or until 2000 iterations are reached. 

On Oxford Pets (Fig.~\ref{fig:SIGIM_pets}), the model briefly attains nearly 93\% accuracy before declining by about $7\%$ due to overfitting. In UCF101 (Fig.~\ref{fig:SIGIM_ucf}), accuracy rises above 74\% but steadily drops and stabilizes near $64\%$. These trends confirm that, despite high initial gains, importance-based updates can still overfit in few-shot scenarios.

\section{Results with a Two-Layer Fully-Connected Architecture}

In addition to the main experiments with CLIP, we evaluate our SO optimizer on a simpler two-layer fully-connected network. 

\subsection{Experimental Methodology}
\label{app:experiments}

The two-layer fully connected network has an input layer of size \(28\times28\), a single hidden layer of size 128 with \(\mathrm{ReLU}\) activations, and an output layer matching the number of classes. 

We conduct two types of experiments: \textbf{(i)} pretraining the model on MNIST or FMNIST, then fine-tuning on a target dataset (adaptation), and \textbf{(ii)} training from randomly initialized weights on the target dataset (no adaptation).

We explore both standard classification (full data) and few-shot learning (limited labeled data) under these scenarios. Finally, we include ablation studies that assess the influence of random-gradient pruning compared with importance-based gradient pruning.

All weights and biases are trainable, and we apply the same hyperparameters across all experiments for all methods. We train until convergence or for a maximum of $3000$ iterations, whichever is reached first. In all two-layer network experiments, we use our SO optimizer and the low-rank baselines with the following default hyperparameters:

\begin{table}
    \centering
    \begin{tabular}{ll}
    \toprule
    \textbf{Hyperparameter} & \textbf{Value} \\
    \midrule
    Learning Rate          & \(1 \times 10^{-3}\) \\
    Adam Betas             & \((\beta_1, \beta_2) = (0.9,\,0.999)\) \\
    Epsilon ($\epsilon$)    & \(1 \times 10^{-8}\) \\
    Sparsity Ratio \((\kappa)\) & \( [1\%, 2\%, 5\%, 8\%, 10\%]\) \\
    Update Interval \((T)\) & 30 (iterations) \\
    Target Loss           & \(10^{-4}\) (early stopping) \\
    \bottomrule
    \end{tabular}
    \caption{Default hyperparameters for two-layer experiments.}
    \label{tab:hyperparams}
\end{table}

We use the following six datasets for all experiments:

\begin{itemize}[leftmargin=1.5em]
    \item \textbf{MNIST}: A canonical dataset of handwritten digits (0--9). Each sample is a \(28\times 28\) grayscale image, comprising 60k training and 10k test examples.
    \item \textbf{FashionMNIST (FMNIST)}: Contains \(28\times 28\) grayscale images of ten clothing and footwear categories. It serves as a harder drop-in replacement for MNIST in benchmarking.
    \item \textbf{EMNIST (Balanced Split)}: Extends MNIST to letters and digits, covering 47 classes of handwritten alphanumeric characters. It includes both uppercase and lowercase letters.
    \item \textbf{PathMNIST}: A medical image dataset with histopathological images of colorectal tissue, each labeled among nine classes (normal tissue, tumor tissue, etc.).
    \item \textbf{OrganMNISTAxial}: Features axial-view organ scans across eleven abdominal classes (e.g., spleen, kidney, aorta). Images are grayscale and resized to \(28\times 28\).
    \item \textbf{BloodMNIST}: Comprises microscopic blood cell images from eight categories. Samples are also converted to a standardized \(28\times 28\) resolution.
\end{itemize}

For adaptation, we first pretrain the models on MNIST or FMNIST. Then, we adapt this pretrained model to several target datasets. In each scenario, we compare SO to state-of-the-art low-rank methods (LoRA, ReLoRA, GaLoRE, etc.) and a fully finetuned baseline (Adam).  

We measure classification accuracy on each target dataset. We report the mean and standard deviation over 10 runs with different random seeds. By presenting results for multiple tasks, we show the robustness and versatility of our sparse optimization approach compared to low-rank and dense baselines.

\begin{table*}
\centering
\caption{Comparison of theoretical memory usage, number of variables, and trainable parameters 
for the first layer \((W_1 \in \mathbb{R}^{128 \times 784})\) of the two-layer fully connected architecture. Activations and bias terms are excluded because they remain consistent across all methods.}
\scalebox{0.7}{
\begin{tabular}{lccccc}
\toprule
Method & Weight (\#Vars, Mem.) & Gradient (\#Vars, Mem.) & Opt. States (\#Vars, Mem.) & \#Trainable & Total Mem. (MB) \\
\midrule
SO ($\kappa=1\%$) & $100352\,(0.38MB)$ & $2007\,(0.01MB)$ & $3010\,(0.01MB)$ & $1003$ & 0.40MB \\
SO ($\kappa=2\%$) & $100352\,(0.38MB)$ & $4014\,(0.02MB)$ & $6021\,(0.02MB)$ & $2007$ & 0.42MB \\
SO ($\kappa=5\%$) & $100352\,(0.38MB)$ & $10035\,(0.04MB)$ & $15052\,(0.06MB)$ & $5017$ & 0.48MB \\
SO ($\kappa=8\%$) & $100352\,(0.38MB)$ & $16056\,(0.06MB)$ & $24084\,(0.09MB)$ & $8028$ & 0.54MB \\
SO ($\kappa=10\%$) & $100352\,(0.38MB)$ & $20070\,(0.08MB)$ & $30105\,(0.11MB)$ & $10035$ & 0.57MB \\
\midrule
GaLoRE ($r=2$) & $100352\,(0.38MB)$ & $100352\,(0.38MB)$ & $3392\,(0.01MB)$ & $100352$ & 0.78MB \\
GaLoRE ($r=4$) & $100352\,(0.38MB)$ & $100352\,(0.38MB)$ & $6784\,(0.03MB)$ & $100352$ & 0.79MB \\
GaLoRE ($r=8$) & $100352\,(0.38MB)$ & $100352\,(0.38MB)$ & $13568\,(0.05MB)$ & $100352$ & 0.82MB \\
GaLoRE ($r=16$) & $100352\,(0.38MB)$ & $100352\,(0.38MB)$ & $27136\,(0.10MB)$ & $100352$ & 0.87MB \\
LoRA ($r=2$) & $102176\,(0.39MB)$ & $1824\,(0.01MB)$ & $3648\,(0.01MB)$ & $1824$ & 0.41MB \\
LoRA ($r=4$) & $104000\,(0.40MB)$ & $3648\,(0.01MB)$ & $7296\,(0.03MB)$ & $3648$ & 0.44MB \\
LoRA ($r=8$) & $107648\,(0.41MB)$ & $7296\,(0.03MB)$ & $14592\,(0.06MB)$ & $7296$ & 0.49MB \\
LoRA ($r=16$) & $114944\,(0.44MB)$ & $14592\,(0.06MB)$ & $29184\,(0.11MB)$ & $14592$ & 0.61MB \\
PiSSA ($r=2$) & $102176\,(0.39MB)$ & $1824\,(0.01MB)$ & $3648\,(0.01MB)$ & $1824$ & 0.41MB \\
PiSSA ($r=4$) & $104000\,(0.40MB)$ & $3648\,(0.01MB)$ & $7296\,(0.03MB)$ & $3648$ & 0.44MB \\
PiSSA ($r=8$) & $107648\,(0.41MB)$ & $7296\,(0.03MB)$ & $14592\,(0.06MB)$ & $7296$ & 0.49MB \\
PiSSA ($r=16$) & $114944\,(0.44MB)$ & $14592\,(0.06MB)$ & $29184\,(0.11MB)$ & $14592$ & 0.61MB \\
DoRA ($r=2$) & $102304\,(0.39MB)$ & $1952\,(0.01MB)$ & $3904\,(0.01MB)$ & $1952$ & 0.41MB \\
DoRA ($r=4$) & $104128\,(0.40MB)$ & $3776\,(0.01MB)$ & $7552\,(0.03MB)$ & $3776$ & 0.44MB \\
DoRA ($r=8$) & $107776\,(0.41MB)$ & $7424\,(0.03MB)$ & $14848\,(0.06MB)$ & $7424$ & 0.50MB \\
DoRA ($r=16$) & $115072\,(0.44MB)$ & $14720\,(0.06MB)$ & $29440\,(0.11MB)$ & $14720$ & 0.61MB \\
ReLoRA ($r=2$) & $102176\,(0.39MB)$ & $1824\,(0.01MB)$ & $3648\,(0.01MB)$ & $1824$ & 0.41MB \\
ReLoRA ($r=4$) & $104000\,(0.40MB)$ & $3648\,(0.01MB)$ & $7296\,(0.03MB)$ & $3648$ & 0.44MB \\
ReLoRA ($r=8$) & $107648\,(0.41MB)$ & $7296\,(0.03MB)$ & $14592\,(0.06MB)$ & $7296$ & 0.49MB \\
ReLoRA ($r=16$) & $114944\,(0.44MB)$ & $14592\,(0.06MB)$ & $29184\,(0.11MB)$ & $14592$ & 0.61MB \\
VeRA ($r=2$) & $102306\,(0.39MB)$ & $130\,(0.00MB)$ & $260\,(0.00MB)$ & $130$ & 0.39MB \\
VeRA ($r=4$) & $104132\,(0.40MB)$ & $132\,(0.00MB)$ & $264\,(0.00MB)$ & $132$ & 0.40MB \\
VeRA ($r=8$) & $107784\,(0.41MB)$ & $136\,(0.00MB)$ & $272\,(0.00MB)$ & $136$ & 0.41MB \\
VeRA ($r=16$) & $115088\,(0.44MB)$ & $144\,(0.00MB)$ & $288\,(0.00MB)$ & $144$ & 0.44MB \\
\midrule
Adam & $100352\,(0.38MB)$ & $100352\,(0.38MB)$ & $200704\,(0.77MB)$ & $100352$ & 1.53MB \\
\bottomrule
\end{tabular}}
\label{tab:comparison_memory_layer1}
\end{table*}

\begin{table*}
\centering
\caption{Comparison of theoretical memory consumption, number of variables, and trainable parameters 
for the second layer \((W_2 \in \mathbb{R}^{128 \times 128})\) of the two-layer fully connected architecture. Activations and bias terms are excluded since they remain unchanged across methods.}
\scalebox{0.7}{\begin{tabular}{lccccc}
\toprule
Method & Weight (\#Vars, Mem.) & Gradient (\#Vars, Mem.) & Opt. States (\#Vars, Mem.) & \#Trainable & Total Mem. (MB) \\
\midrule
SO ($\kappa=0.01$) & $16384\,(0.06MB)$ & $327\,(0.00MB)$ & $491\,(0.00MB)$ & $163$ & 0.07MB \\
SO ($\kappa=0.02$) & $16384\,(0.06MB)$ & $655\,(0.00MB)$ & $983\,(0.00MB)$ & $327$ & 0.07MB \\
SO ($\kappa=0.05$) & $16384\,(0.06MB)$ & $1638\,(0.01MB)$ & $2457\,(0.01MB)$ & $819$ & 0.08MB \\
SO ($\kappa=0.08$) & $16384\,(0.06MB)$ & $2621\,(0.01MB)$ & $3932\,(0.01MB)$ & $1310$ & 0.09MB \\
SO ($\kappa=0.1$) & $16384\,(0.06MB)$ & $3276\,(0.01MB)$ & $4915\,(0.02MB)$ & $1638$ & 0.09MB \\
\midrule
GaLoRE ($r=2$) & $16384\,(0.06MB)$ & $16384\,(0.06MB)$ & $768\,(0.00MB)$ & $16384$ & 0.13MB \\
GaLoRE ($r=4$) & $16384\,(0.06MB)$ & $16384\,(0.06MB)$ & $1536\,(0.01MB)$ & $16384$ & 0.13MB \\
GaLoRE ($r=8$) & $16384\,(0.06MB)$ & $16384\,(0.06MB)$ & $3072\,(0.01MB)$ & $16384$ & 0.14MB \\
GaLoRE ($r=16$) & $16384\,(0.06MB)$ & $16384\,(0.06MB)$ & $6144\,(0.02MB)$ & $16384$ & 0.15MB \\
LoRA ($r=2$) & $16896\,(0.06MB)$ & $512\,(0.00MB)$ & $1024\,(0.00MB)$ & $512$ & 0.07MB \\
LoRA ($r=4$) & $17408\,(0.07MB)$ & $1024\,(0.00MB)$ & $2048\,(0.01MB)$ & $1024$ & 0.08MB \\
LoRA ($r=8$) & $18432\,(0.07MB)$ & $2048\,(0.01MB)$ & $4096\,(0.02MB)$ & $2048$ & 0.09MB \\
LoRA ($r=16$) & $20480\,(0.08MB)$ & $4096\,(0.02MB)$ & $8192\,(0.03MB)$ & $4096$ & 0.12MB \\
PiSSA ($r=2$) & $16896\,(0.06MB)$ & $512\,(0.00MB)$ & $1024\,(0.00MB)$ & $512$ & 0.07MB \\
PiSSA ($r=4$) & $17408\,(0.07MB)$ & $1024\,(0.00MB)$ & $2048\,(0.01MB)$ & $1024$ & 0.08MB \\
PiSSA ($r=8$) & $18432\,(0.07MB)$ & $2048\,(0.01MB)$ & $4096\,(0.02MB)$ & $2048$ & 0.09MB \\
PiSSA ($r=16$) & $20480\,(0.08MB)$ & $4096\,(0.02MB)$ & $8192\,(0.03MB)$ & $4096$ & 0.12MB \\
DoRA ($r=2$) & $17024\,(0.06MB)$ & $640\,(0.00MB)$ & $1280\,(0.00MB)$ & $640$ & 0.07MB \\
DoRA ($r=4$) & $17536\,(0.07MB)$ & $1152\,(0.00MB)$ & $2304\,(0.01MB)$ & $1152$ & 0.08MB \\
DoRA ($r=8$) & $18560\,(0.07MB)$ & $2176\,(0.01MB)$ & $4352\,(0.02MB)$ & $2176$ & 0.10MB \\
DoRA ($r=16$) & $20608\,(0.08MB)$ & $4224\,(0.02MB)$ & $8448\,(0.03MB)$ & $4224$ & 0.13MB \\
ReLoRA ($r=2$) & $16896\,(0.06MB)$ & $512\,(0.00MB)$ & $1024\,(0.00MB)$ & $512$ & 0.07MB \\
ReLoRA ($r=4$) & $17408\,(0.07MB)$ & $1024\,(0.00MB)$ & $2048\,(0.01MB)$ & $1024$ & 0.08MB \\
ReLoRA ($r=8$) & $18432\,(0.07MB)$ & $2048\,(0.01MB)$ & $4096\,(0.02MB)$ & $2048$ & 0.09MB \\
ReLoRA ($r=16$) & $20480\,(0.08MB)$ & $4096\,(0.02MB)$ & $8192\,(0.03MB)$ & $4096$ & 0.12MB \\
VeRA ($r=2$) & $17026\,(0.06MB)$ & $130\,(0.00MB)$ & $260\,(0.00MB)$ & $130$ & 0.07MB \\
VeRA ($r=4$) & $17540\,(0.07MB)$ & $132\,(0.00MB)$ & $264\,(0.00MB)$ & $132$ & 0.07MB \\
VeRA ($r=8$) & $18568\,(0.07MB)$ & $136\,(0.00MB)$ & $272\,(0.00MB)$ & $136$ & 0.07MB \\
VeRA ($r=16$) & $20624\,(0.08MB)$ & $144\,(0.00MB)$ & $288\,(0.00MB)$ & $144$ & 0.08MB \\
\midrule
Adam & $16384\,(0.06MB)$ & $16384\,(0.06MB)$ & $32768\,(0.12MB)$ & $16384$ & 0.25MB \\
\bottomrule
\end{tabular}}
\label{tab:comparison_memory_layer2}
\end{table*}

\subsection{Memory Consumption}
\label{app:app_memory}

We evaluate GPU memory usage in a two-layer fully connected network. Tables~\ref{tab:comparison_memory_layer1} and~\ref{tab:comparison_memory_layer2} compare theoretical memory consumption, number of variables, and trainable parameters for the first and second layers, respectively. Activations and bias terms are excluded since they are identical for all methods.

Table~\ref{tab:comparison_memory_layer1} reports results for the first layer \((W_1 \in \mathbb{R}^{128 \times 784})\). Sparse Optimization (SO) requires fewer additional variables than most low-rank methods at a similar sparsity ratio. Its gradient and optimizer states are smaller because of dynamic sparsity. 
In contrast, Adam consumes the largest memory due to storing full gradients and optimizer states. 
Overall, SO balances memory and flexibility by freezing the base weights and updating a small subset of parameters.

Table~\ref{tab:comparison_memory_layer2} presents results for the second layer \((W_2 \in \mathbb{R}^{128 \times 128})\). SO uses considerably fewer gradients and optimizer variables, especially for low sparsity. 
Low-rank methods show higher memory usage due to extra low-rank matrices per layer. In contrast, SO’s minimal updates mitigate memory overhead. Hence, the results confirm that SO reduces memory requirements.

\subsection{Results}
\label{app:results}

\textbf{Effectiveness in Classification.} Table~\ref{tab:comparison_classification} presents classification performance when training from scratch (no pretraining) on each target dataset. Adam achieves slightly higher accuracy on some tasks (e.g., EMNIST, MNIST), but it often performs comparably or worse on others. Meanwhile, SO consistently outperforms GaLoRE ($r=2$--16) on datasets like OrganMNISTAxial, BloodMNIST, and BreastMNIST. For instance, SO ($\kappa=1\%$ or $2\%$) generally improves upon the low-rank methods while retaining fewer trainable parameters. This highlights the ability of optimizer.

Tables~\ref{tab:comparison_acc_finetuning_classification_source_mnist} and~\ref{tab:comparison_acc_finetuning_classification_source_fmnist} report results after pretraining on MNIST or FMNIST, respectively. All methods benefit from pretraining, and Adam attains strong performance. However, our SO optimizer often yields higher or comparable accuracy to the low-rank baselines across most datasets, especially for moderate $\kappa$ values (e.g., 1\%--5\%). In OrganMNISTAxial or BloodMNIST, for instance, SO frequently exceeds or matches GaLoRE and ReLoRA, while also preserving memory efficiency (Sec.~\ref{app:app_memory}).

Tables~\ref{tab:ablation_so_classification}, \ref{tab:ablation_so_adaptation_classification_MNIST}, 
and~\ref{tab:ablation_so_adaptation_classification_FMNIST} compare two pruning strategies: 
importance-based (selecting the largest gradients) vs.\ randomness-based (selecting random gradients). When training from scratch or adapting a pretrained model, we observe that randomness-based pruning generally outperforms its importance-based counterpart, particularly at lower $\kappa$. This supports our hypothesis that sparse updates driven by random gradient selection mitigate overfitting more effectively than always choosing high-magnitude gradients. 

\textbf{Effectiveness in Few-Shot Learning.} Table~\ref{tab:comparison_acc_few_shot} shows results when training from scratch on just a few labeled samples per class (4, 8, and 16 shots). GaLoRE tends to underfit for low‐shot regimes, especially when \(r\) is small. In contrast, SO (\(\kappa\leq2\%\)) achieves higher accuracy on datasets such as EMNIST, MNIST, and BreastMNIST. For instance, at 4 shots, SO surpasses GaLoRE by up to 2–3\% in EMNIST and BreastMNIST. This performance is probably attributed to the capacity of SO to mitigate overfitting in limited data situations, even without pretraining.

Tables~\ref{tab:comparison_acc_finetuning_few_shot_source_mnist} and \ref{tab:comparison_acc_finetuning_few_shot_source_fmnist} provide few‐shot results after pretraining on MNIST or FMNIST, respectively. All methods improve considerably over the no‐adaptation scenario, as the pretrained backbone offers a strong initialization. Nevertheless, SO still outperforms many low‐rank baselines (LoRA, ReLoRA, GaLoRE) at 4, 8, or 16 shots, often matching or exceeding Adam. The benefits of sparsity persist in this setting, allowing SO to avoid overfitting.

Tables~\ref{tab:ablation_study_SO_shots}--\ref{tab:ablation_study_adaptation_FMNIST} compare importance‐based vs. randomness‐based gradient pruning in few‐shot scenarios. We observe that random gradient selection provides better accuracy, especially at lower \(\kappa\). By avoiding exclusive reliance on large‐magnitude updates, SO’s sparse updates reduce overfitting risk. 

\begin{figure*}
\centering
\captionsetup{justification=centering, singlelinecheck=false} 
\caption{Comparison of Random and Importance Gradient Pruning in Sparse Optimization (SO) – Gradient Rank Evolution in Few-Shot Learning (4 shots) on EMNIST Dataset.}
\label{fig:rank_EMNIST_4_shots}
\begin{adjustbox}{max width=0.92\textwidth} 
\begin{tabular}{|c|}
\hline
\begin{subfigure}[b]{0.9\textwidth}
    \centering
    \includegraphics[width=\linewidth]{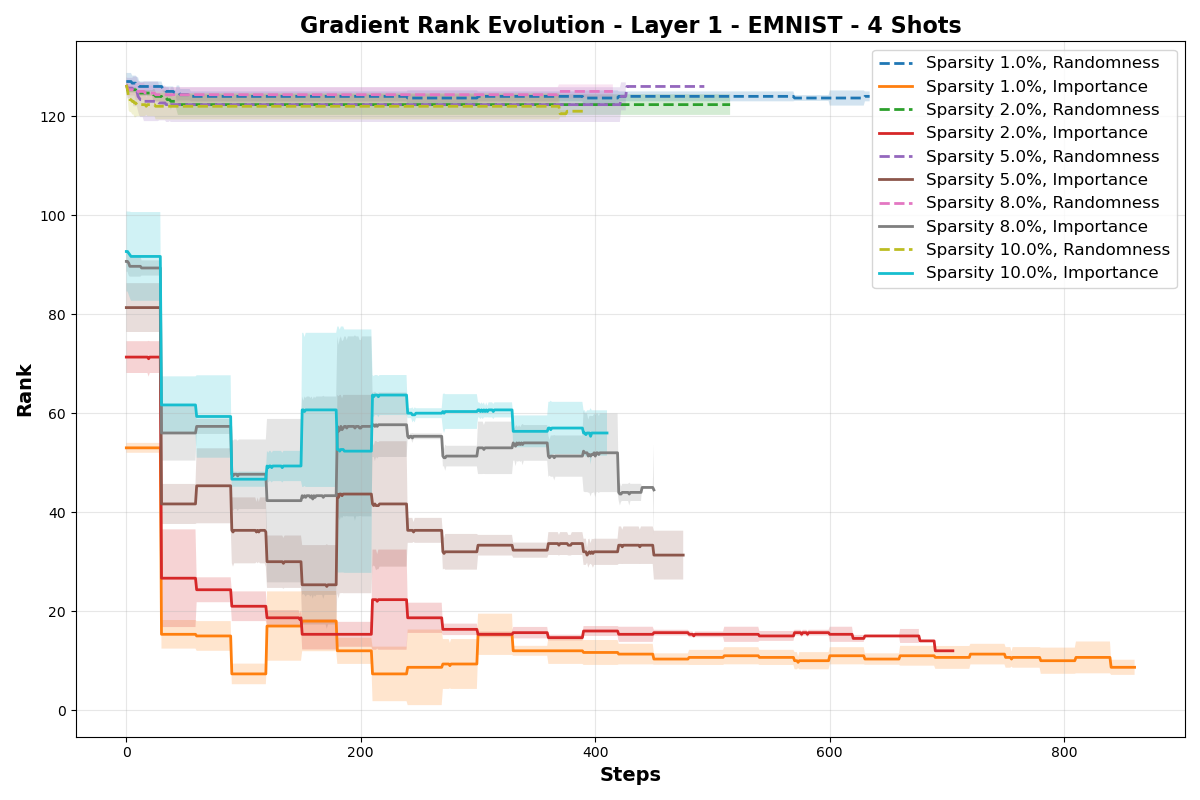}
\end{subfigure} \\ \hline
\begin{subfigure}[b]{0.9\textwidth}
    \centering
    \includegraphics[width=\linewidth]{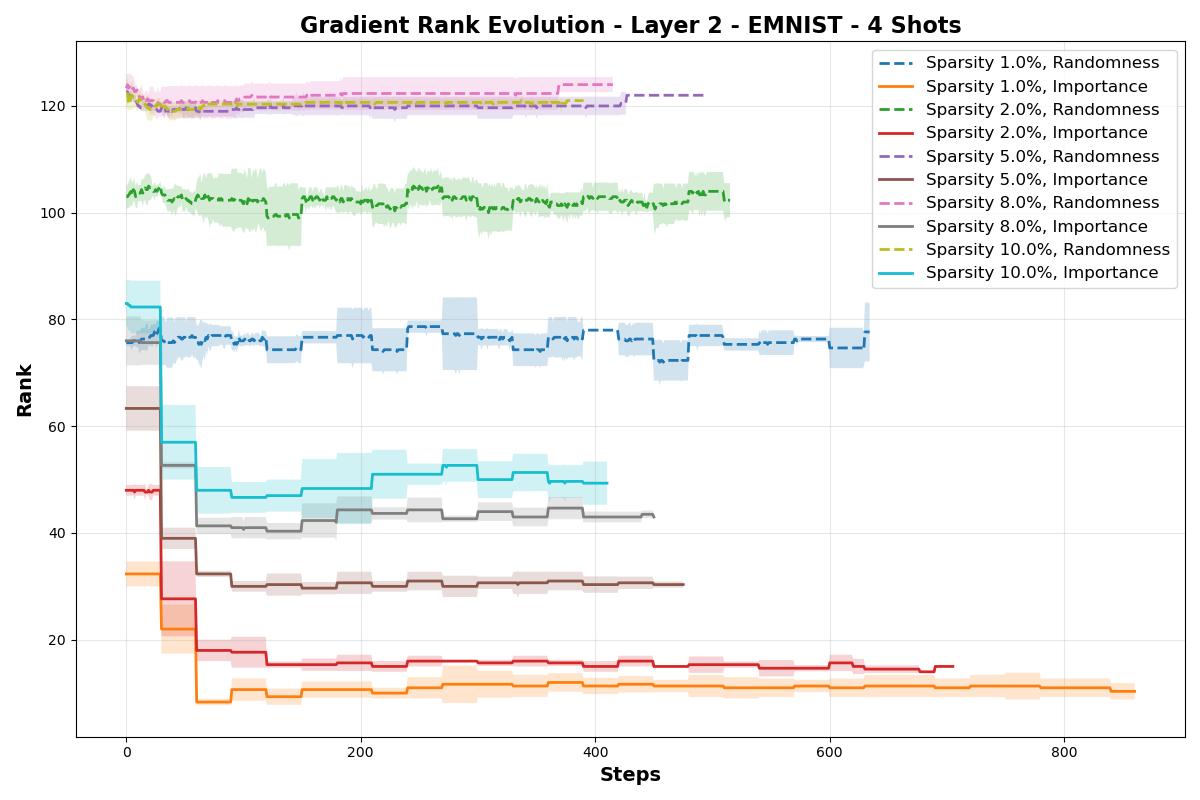}
\end{subfigure} \\ \hline
\end{tabular}
\end{adjustbox}
\end{figure*}

\begin{figure*}
\centering
\captionsetup{justification=centering, singlelinecheck=false} 
\caption{Comparison of Random and Importance Gradient Pruning in Sparse Optimization (SO) – Loss Evolution in Few-Shot Learning (4 shots) on EMNIST Dataset.}
\label{fig:loss_EMNIST_4_shots}
\begin{adjustbox}{max width=0.92\textwidth} 
\begin{tabular}{|c|}
\hline
\begin{subfigure}[b]{0.9\textwidth}
    \centering
    \includegraphics[width=\linewidth]{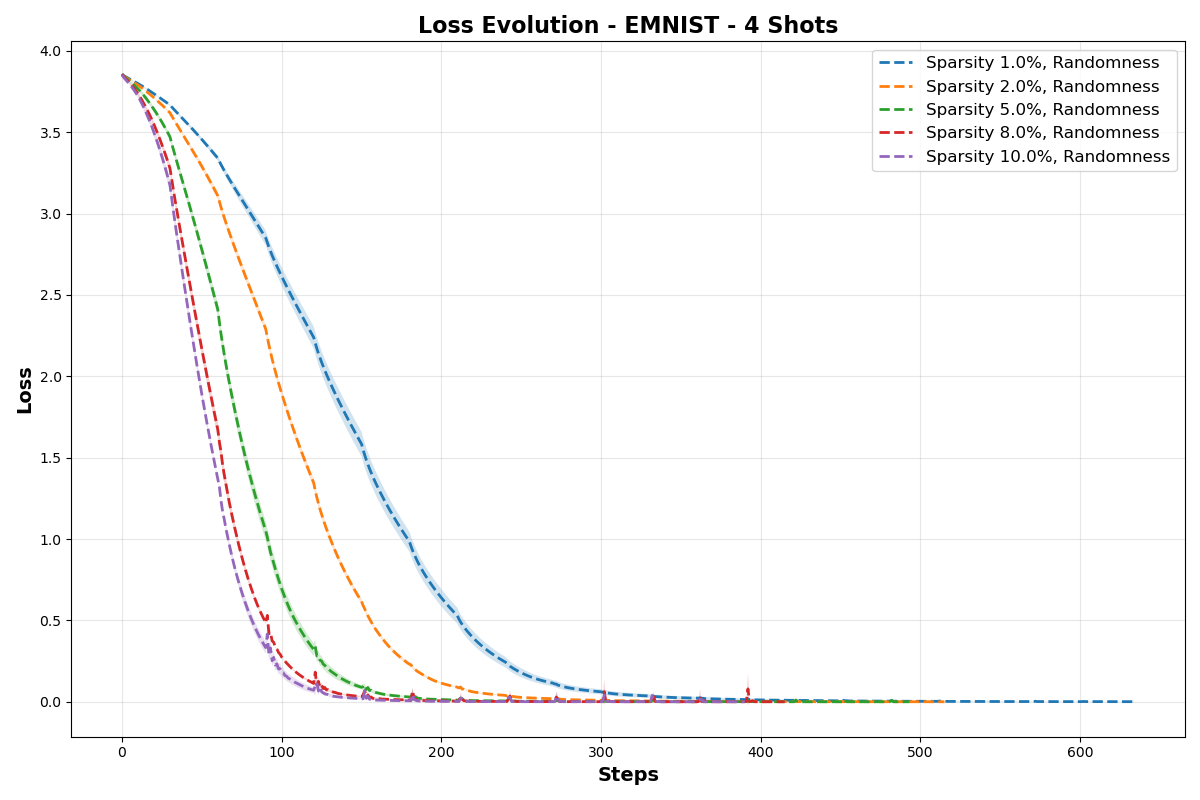}
\end{subfigure} \\ \hline
\begin{subfigure}[b]{0.9\textwidth}
    \centering
    \includegraphics[width=\linewidth]{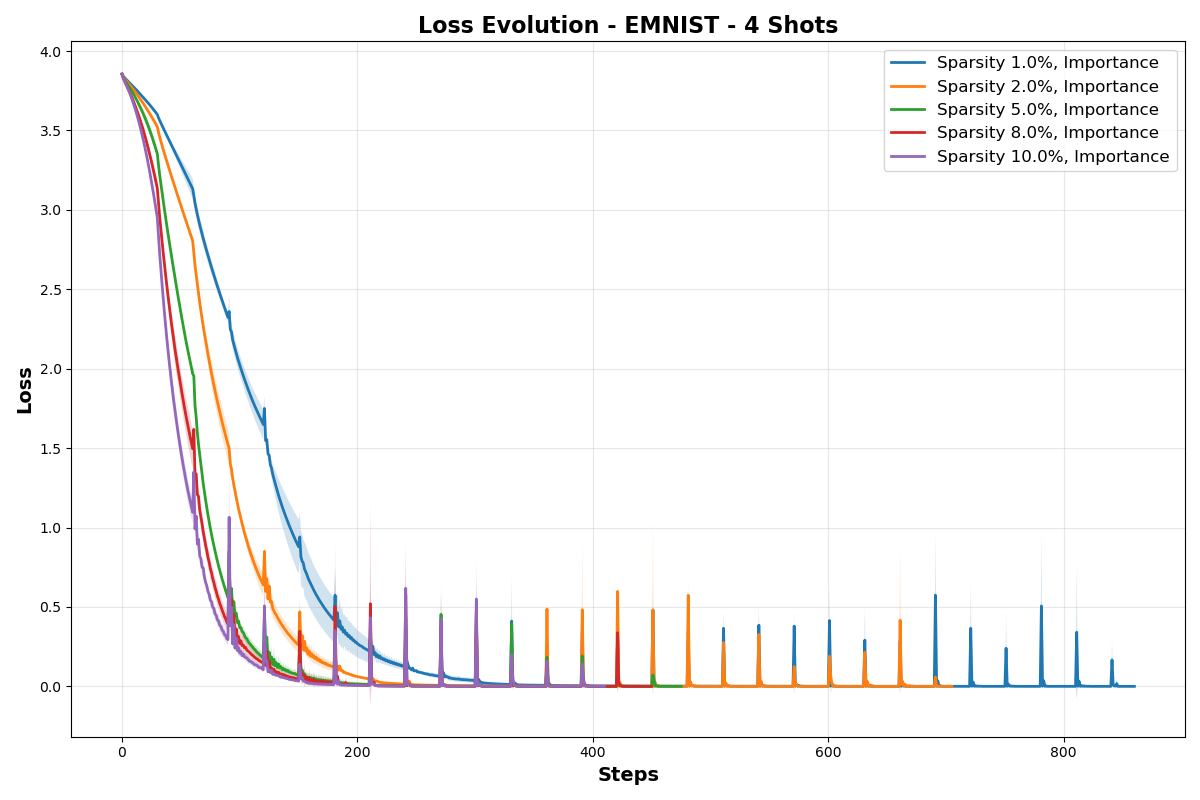}
\end{subfigure} \\ \hline
\end{tabular}
\end{adjustbox}
\end{figure*}

\begin{figure*}
\centering
\captionsetup{justification=centering, singlelinecheck=false} 
\caption{Comparison of Random and Importance Gradient Pruning in Sparse Optimization (SO) – Gradient Rank Evolution in Few-Shot Learning (8 shots) on EMNIST Dataset.}
\label{fig:rank_EMNIST_8_shots}
\begin{adjustbox}{max width=0.92\textwidth} 
\begin{tabular}{|c|}
\hline
\begin{subfigure}[b]{0.9\textwidth}
    \centering
    \includegraphics[width=\linewidth]{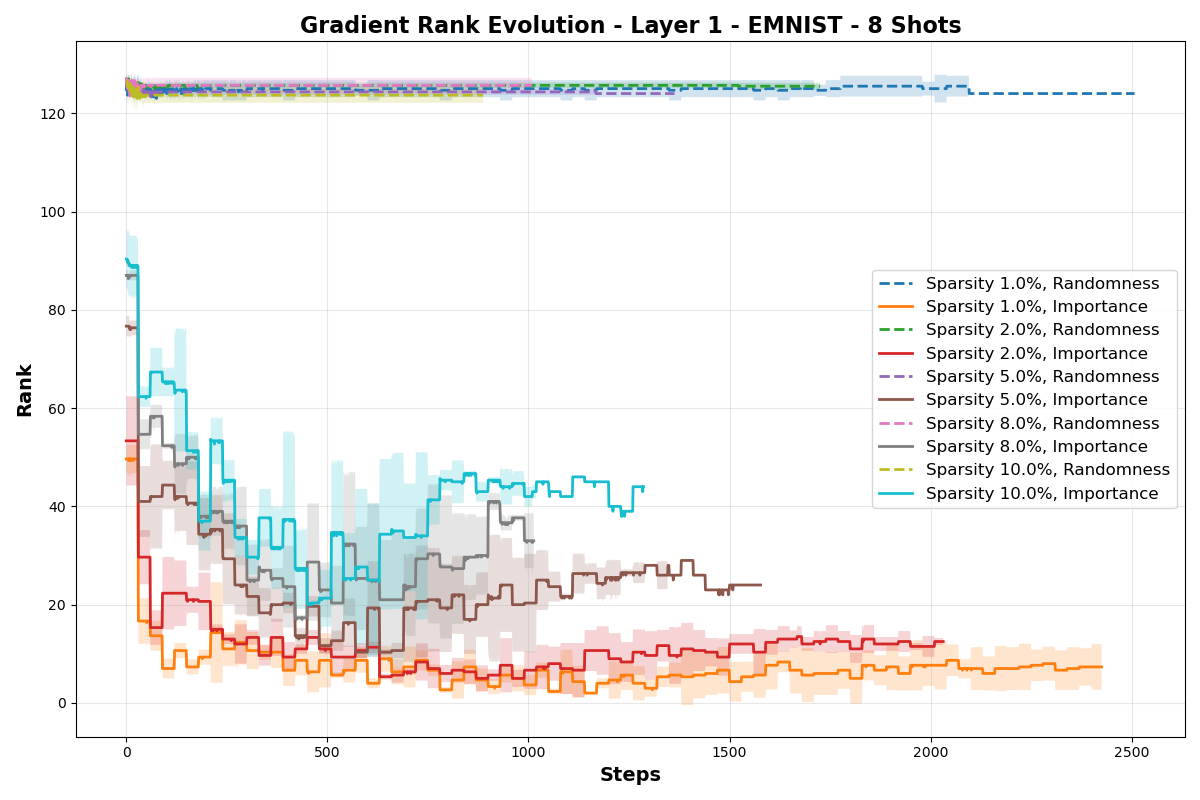}
\end{subfigure} \\ \hline
\begin{subfigure}[b]{0.9\textwidth}
    \centering
    \includegraphics[width=\linewidth]{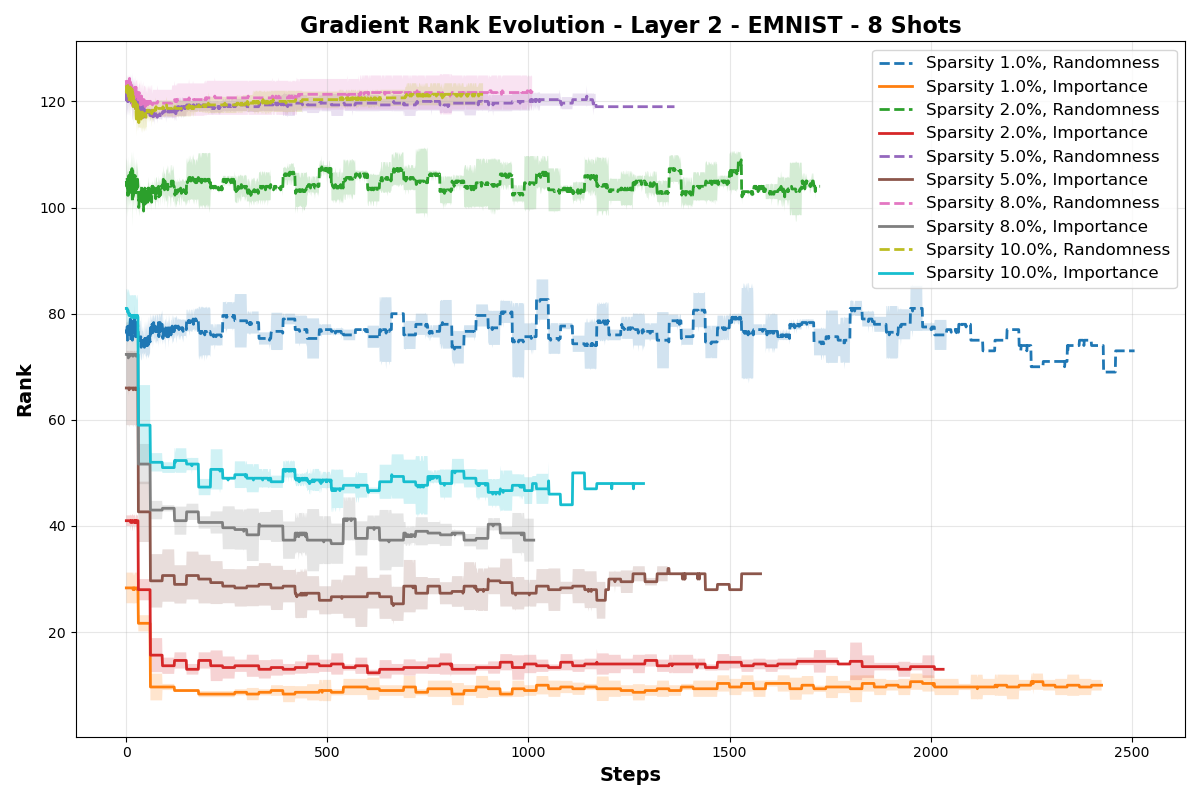}
\end{subfigure} \\ \hline
\end{tabular}
\end{adjustbox}
\end{figure*}

\begin{figure*}
\centering
\captionsetup{justification=centering, singlelinecheck=false} 
\caption{Comparison of Random and Importance Gradient Pruning in Sparse Optimization (SO) – Loss Evolution of SO in Few-Shot Learning (8 shots) on EMNIST Dataset.}
\label{fig:loss_EMNIST_8_shots}
\begin{adjustbox}{max width=0.92\textwidth} 
\begin{tabular}{|c|}
\hline
\begin{subfigure}[b]{0.9\textwidth}
    \centering
    \includegraphics[width=\linewidth]{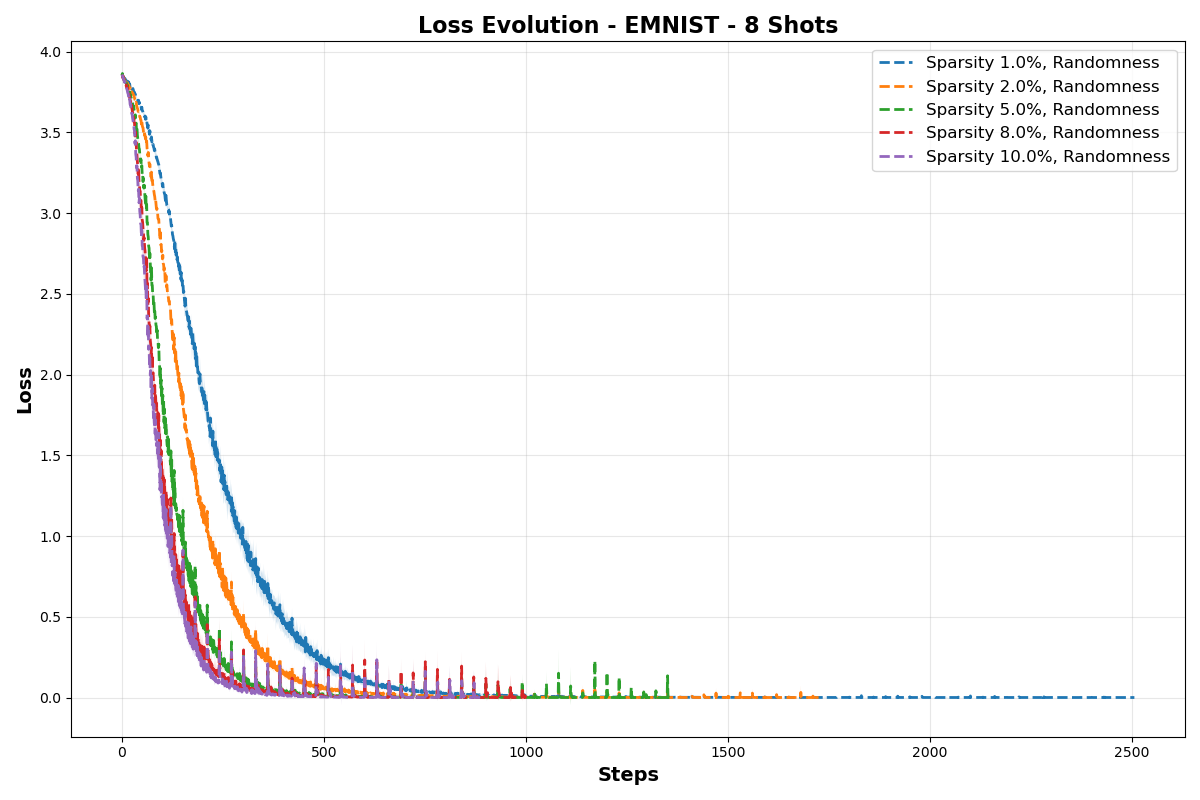}
\end{subfigure} \\ \hline
\begin{subfigure}[b]{0.9\textwidth}
    \centering
    \includegraphics[width=\linewidth]{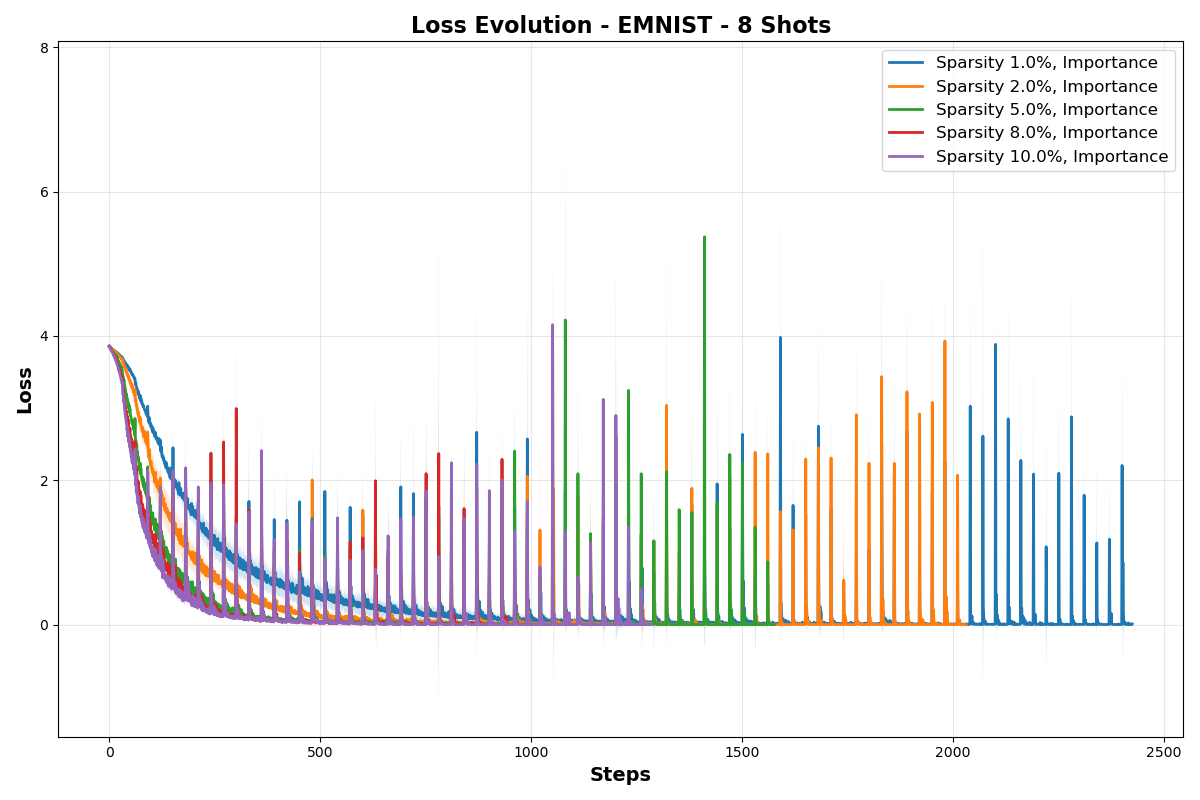}
\end{subfigure} \\ \hline
\end{tabular}
\end{adjustbox}
\end{figure*}

\begin{figure*}
\centering
\captionsetup{justification=centering, singlelinecheck=false} 
\caption{Comparison of Random and Importance Gradient Pruning in Sparse Optimization (SO) – Gradient Rank Evolution in Few-Shot Learning (16 shots) on EMNIST Dataset.}
\label{fig:rank_EMNIST_16_shots}
\begin{adjustbox}{max width=0.92\textwidth} 
\begin{tabular}{|c|}
\hline
\begin{subfigure}[b]{0.9\textwidth}
    \centering
    \includegraphics[width=\linewidth]{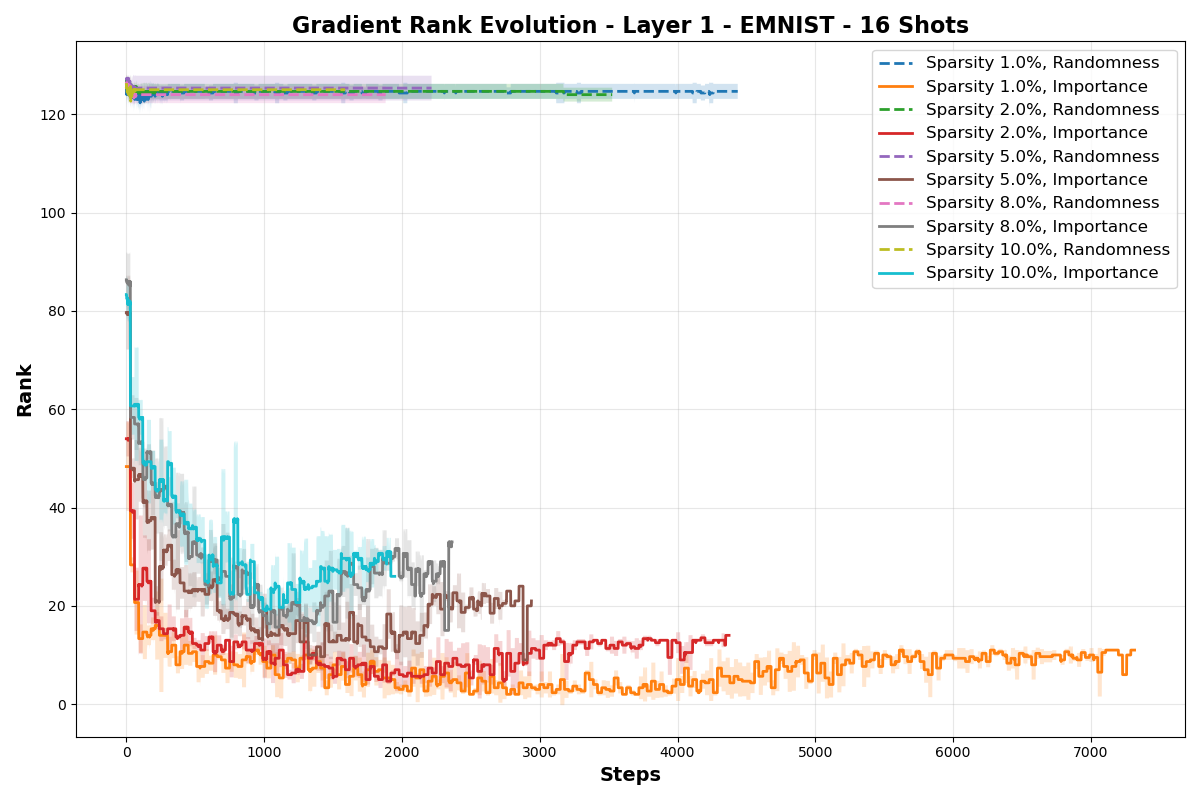}
\end{subfigure} \\ \hline
\begin{subfigure}[b]{0.9\textwidth}
    \centering
    \includegraphics[width=\linewidth]{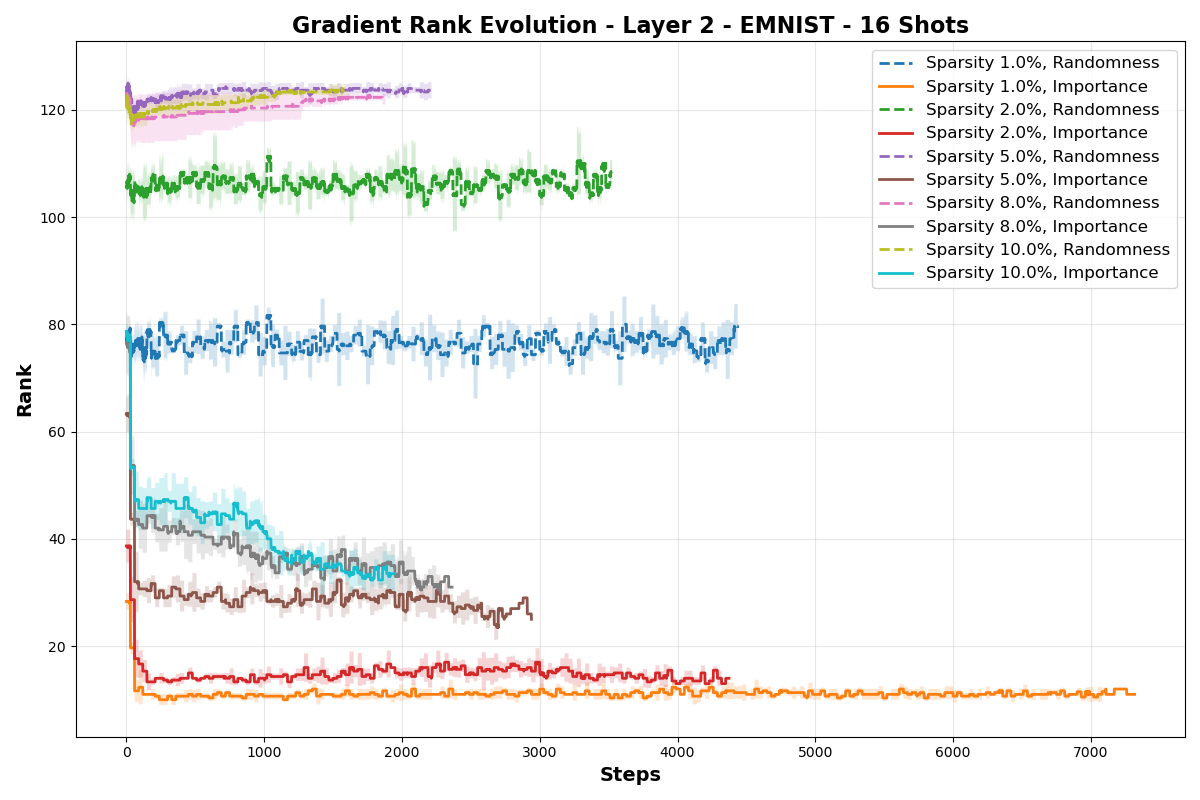}
\end{subfigure} \\ \hline
\end{tabular}
\end{adjustbox}
\end{figure*}

\begin{figure*}
\centering
\captionsetup{justification=centering, singlelinecheck=false} 
\caption{Comparison of Random and Importance Gradient Pruning in Sparse Optimization (SO) – Loss Evolution in Few-Shot Learning (16 shots) on EMNIST Dataset.}
\label{fig:loss_EMNIST_16_shots}
\begin{adjustbox}{max width=0.92\textwidth} 
\begin{tabular}{|c|}
\hline
\begin{subfigure}[b]{0.9\textwidth}
    \centering
    \includegraphics[width=\linewidth]{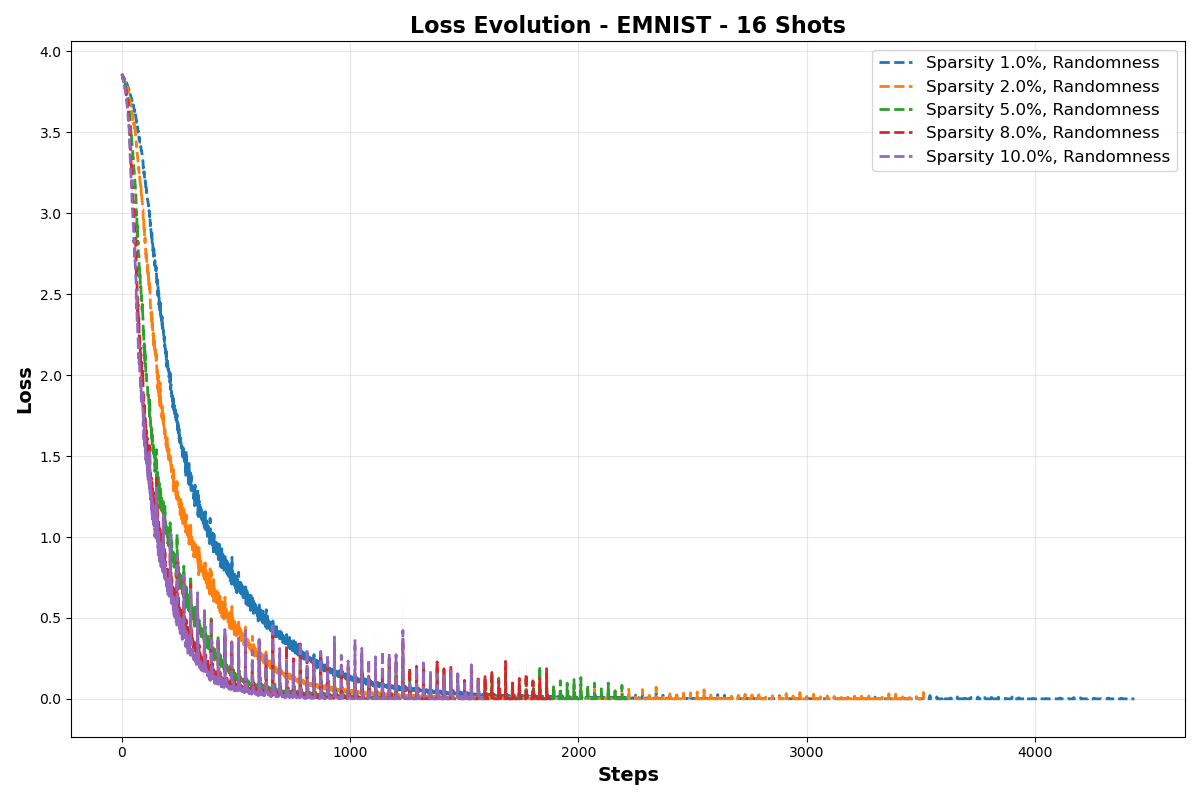}
\end{subfigure} \\ \hline
\begin{subfigure}[b]{0.9\textwidth}
    \centering
    \includegraphics[width=\linewidth]{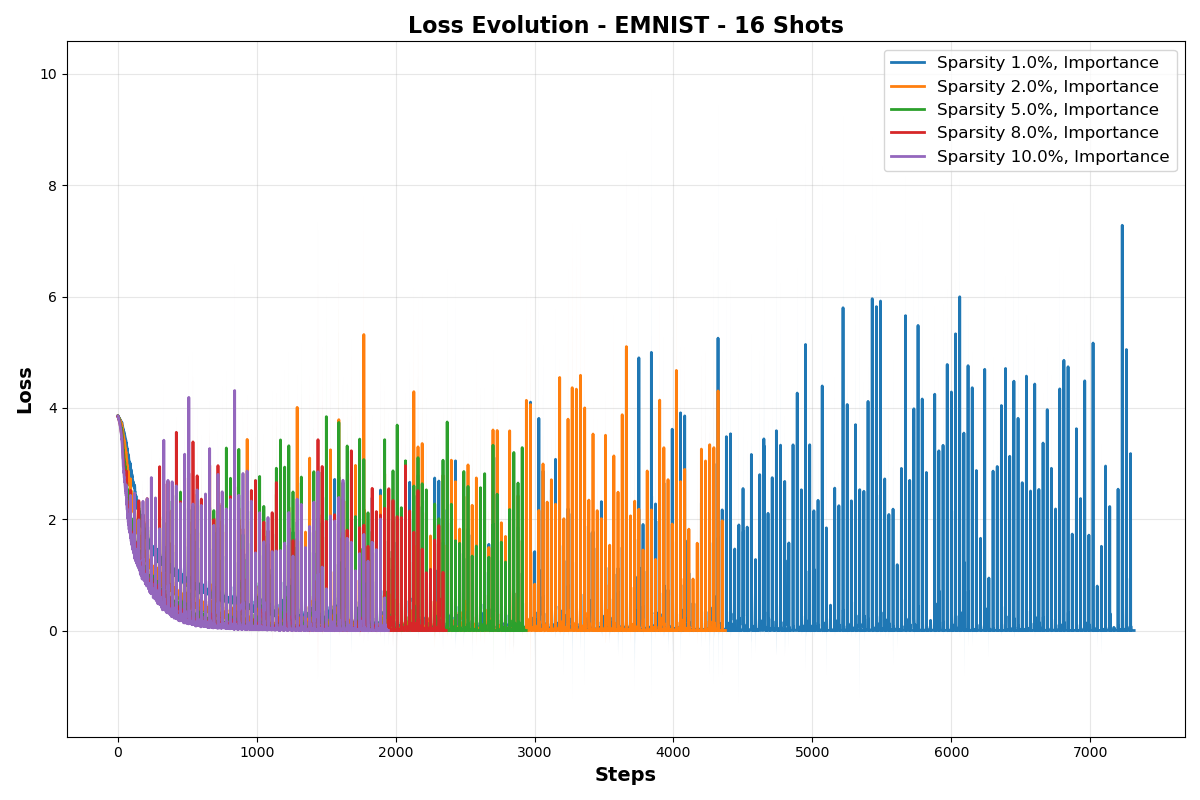}
\end{subfigure} \\ \hline
\end{tabular}
\end{adjustbox}
\end{figure*}

\begin{figure*}
\centering
\captionsetup{justification=centering, singlelinecheck=false} 
\caption{Comparison of Random and Importance Gradient Pruning in Sparse Optimization (SO) – Gradient Rank Evolution in Few-Shot Learning (4 shots) on MNIST Dataset.}
\label{fig:rank_MNIST_4_shots}
\begin{adjustbox}{max width=0.92\textwidth} 
\begin{tabular}{|c|}
\hline
\begin{subfigure}[b]{0.9\textwidth}
    \centering
    \includegraphics[width=\linewidth]{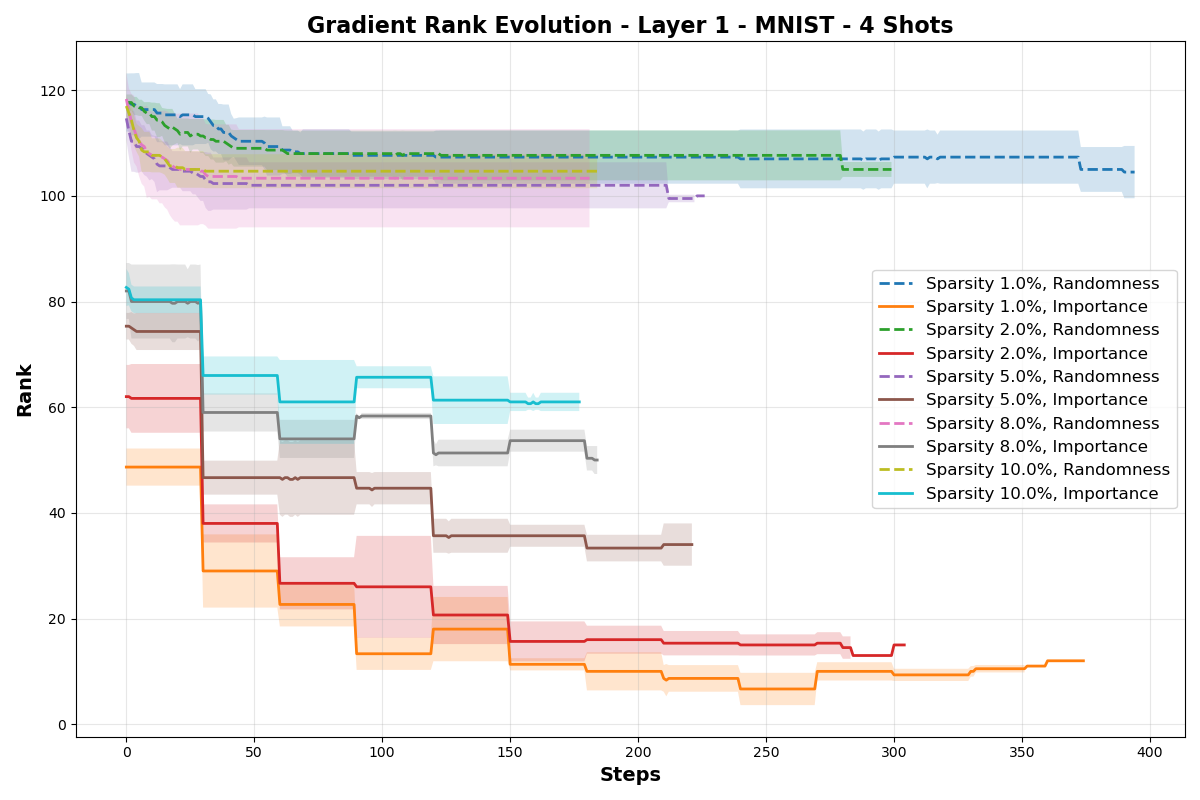}
\end{subfigure} \\ \hline
\begin{subfigure}[b]{0.9\textwidth}
    \centering
    \includegraphics[width=\linewidth]{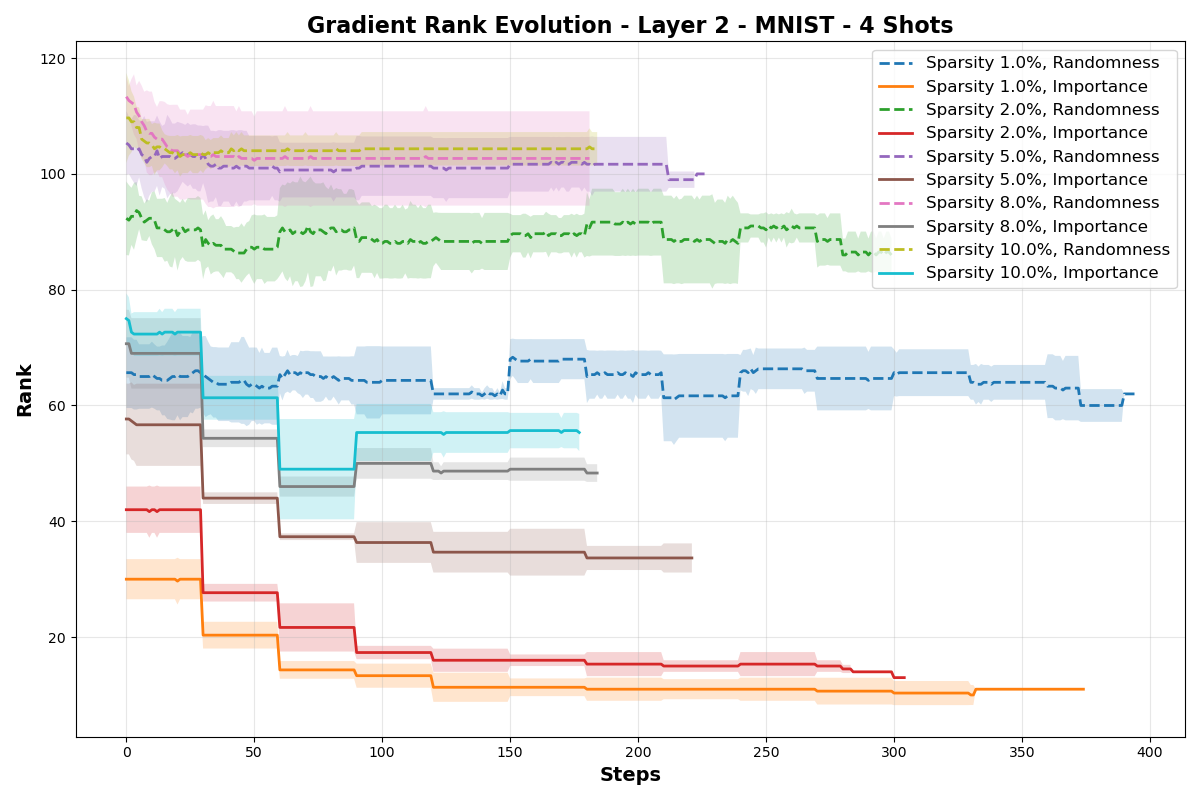}
\end{subfigure} \\ \hline
\end{tabular}
\end{adjustbox}
\end{figure*}

\begin{figure*}
\centering
\captionsetup{justification=centering, singlelinecheck=false} 
\caption{Comparison of Random and Importance Gradient Pruning in Sparse Optimization (SO) – Loss Evolution in Few-Shot Learning (4 shots) on MNIST Dataset.}
\label{fig:loss_MNIST_4_shots}
\begin{adjustbox}{max width=0.92\textwidth} 
\begin{tabular}{|c|}
\hline
\begin{subfigure}[b]{0.9\textwidth}
    \centering
    \includegraphics[width=\linewidth]{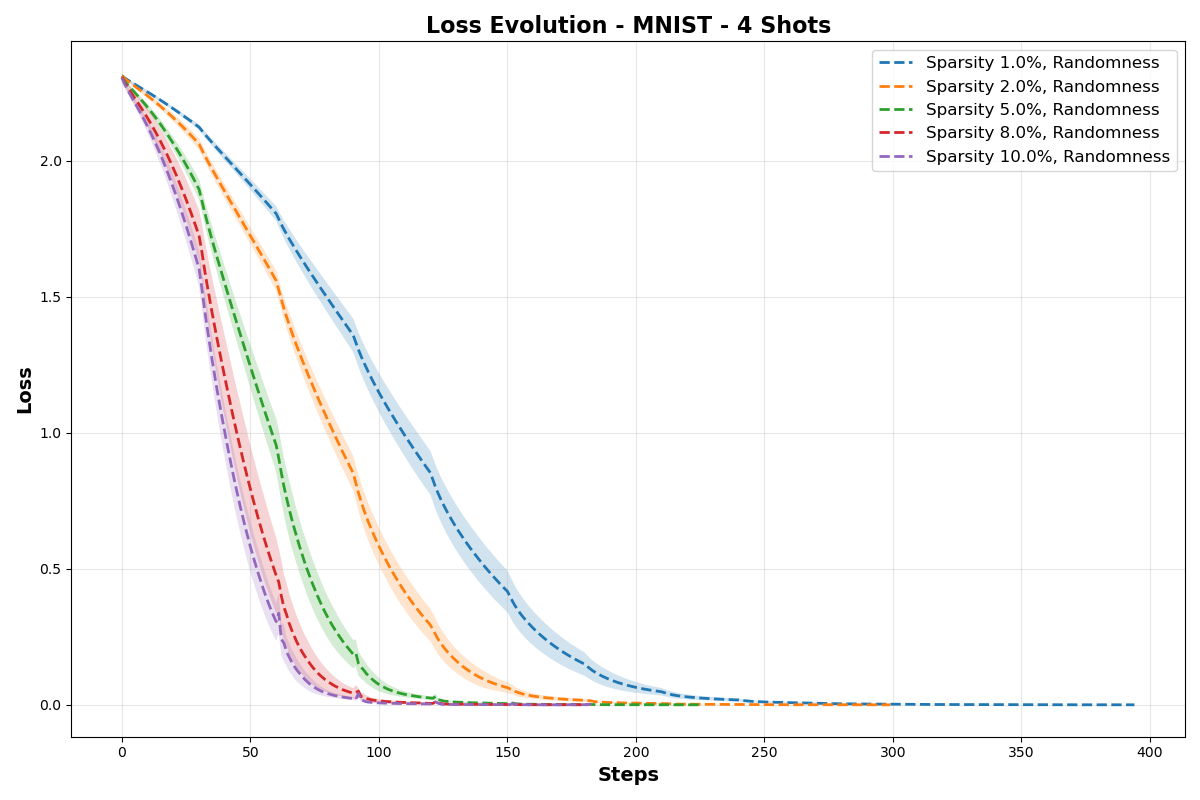}
\end{subfigure} \\ \hline
\begin{subfigure}[b]{0.9\textwidth}
    \centering
    \includegraphics[width=\linewidth]{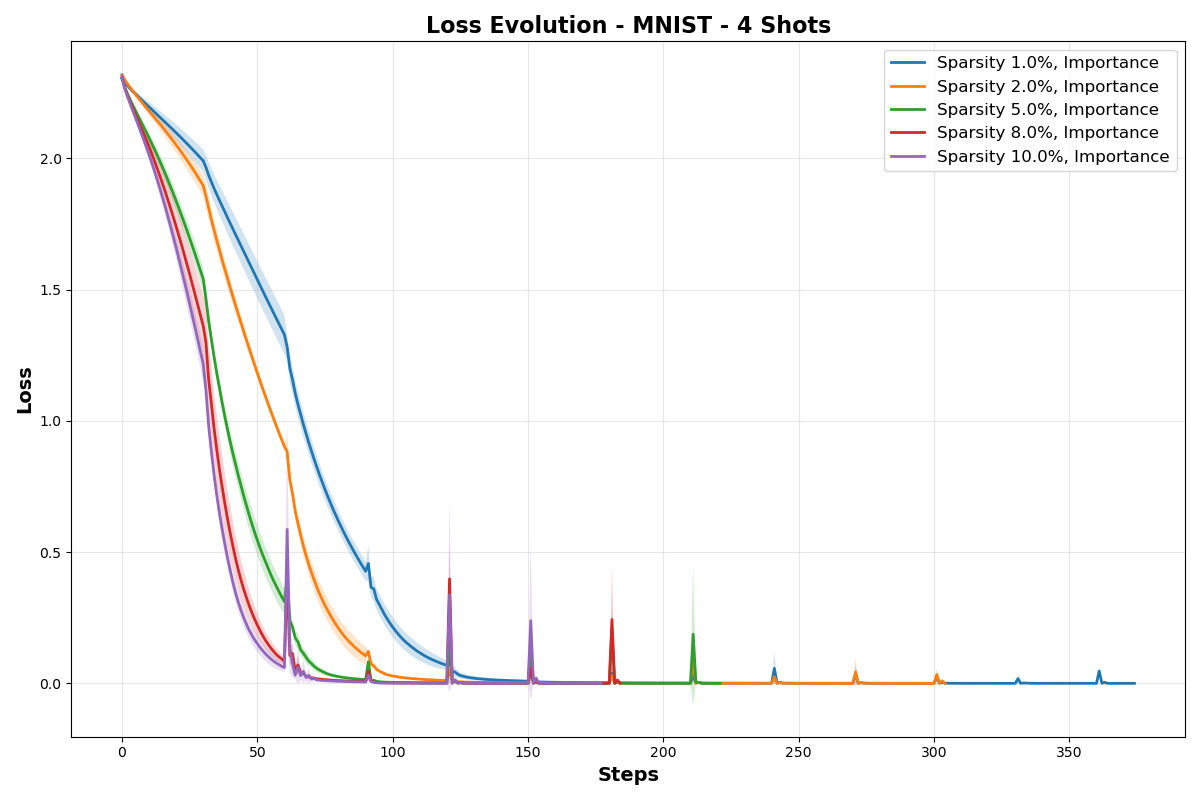}
\end{subfigure} \\ \hline
\end{tabular}
\end{adjustbox}
\end{figure*}

\begin{figure*}
\centering
\captionsetup{justification=centering, singlelinecheck=false} 
\caption{Comparison of Random and Importance Gradient Pruning in Sparse Optimization (SO) – Gradient Rank Evolution in Few-Shot Learning (8 shots) on MNIST Dataset.}
\label{fig:rank_MNIST_8_shots}
\begin{adjustbox}{max width=0.92\textwidth} 
\begin{tabular}{|c|}
\hline
\begin{subfigure}[b]{0.9\textwidth}
    \centering
    \includegraphics[width=\linewidth]{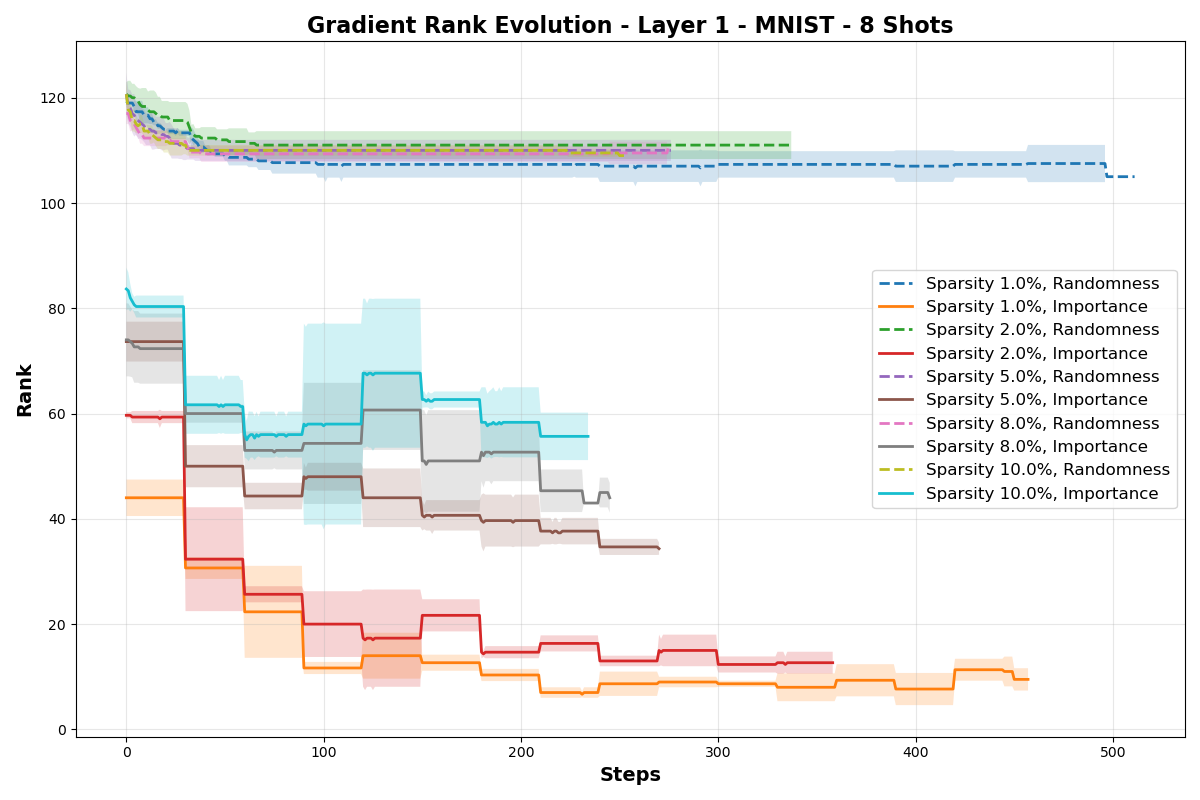}
\end{subfigure} \\ \hline
\begin{subfigure}[b]{0.9\textwidth}
    \centering
    \includegraphics[width=\linewidth]{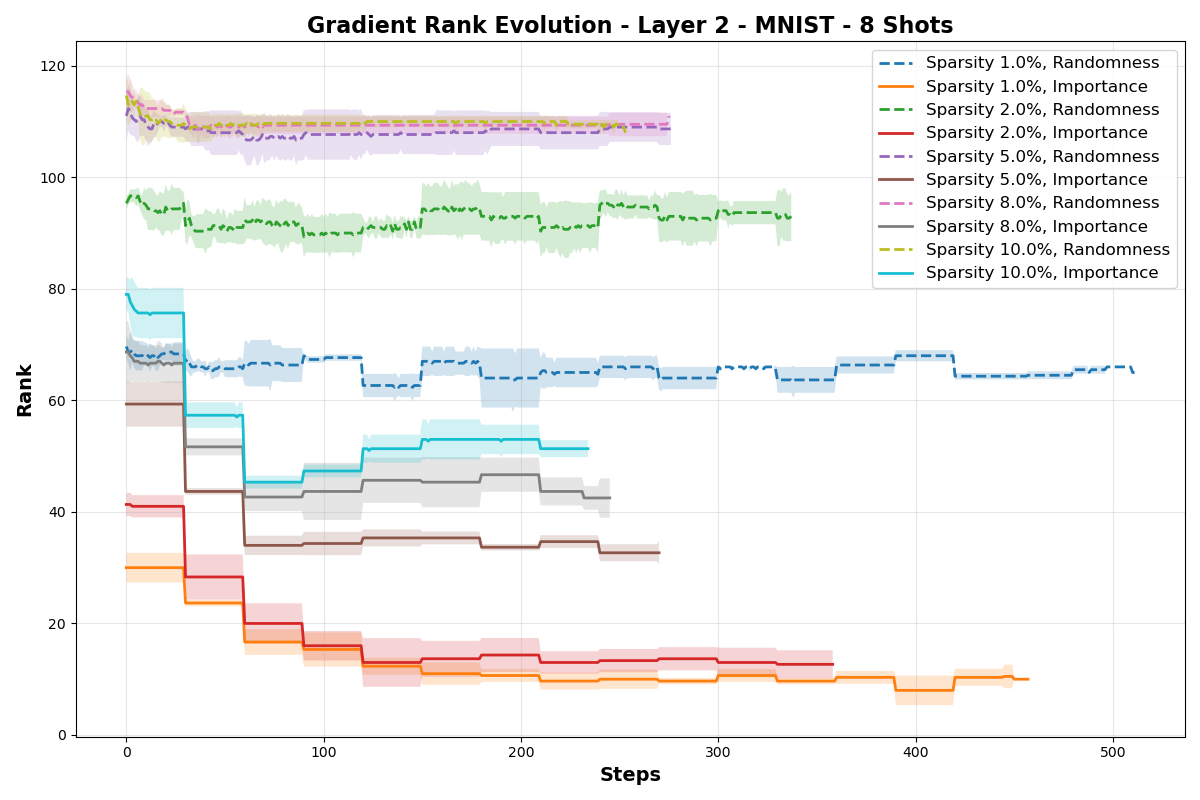}
\end{subfigure} \\ \hline
\end{tabular}
\end{adjustbox}
\end{figure*}

\begin{figure*}
\centering
\captionsetup{justification=centering, singlelinecheck=false} 
\caption{Comparison of Random and Importance Gradient Pruning in Sparse Optimization (SO) – Loss Evolution in Few-Shot Learning (8 shots) on MNIST Dataset.}
\label{fig:loss_MNIST_8_shots}
\begin{adjustbox}{max width=0.92\textwidth} 
\begin{tabular}{|c|}
\hline
\begin{subfigure}[b]{0.9\textwidth}
    \centering
    \includegraphics[width=\linewidth]{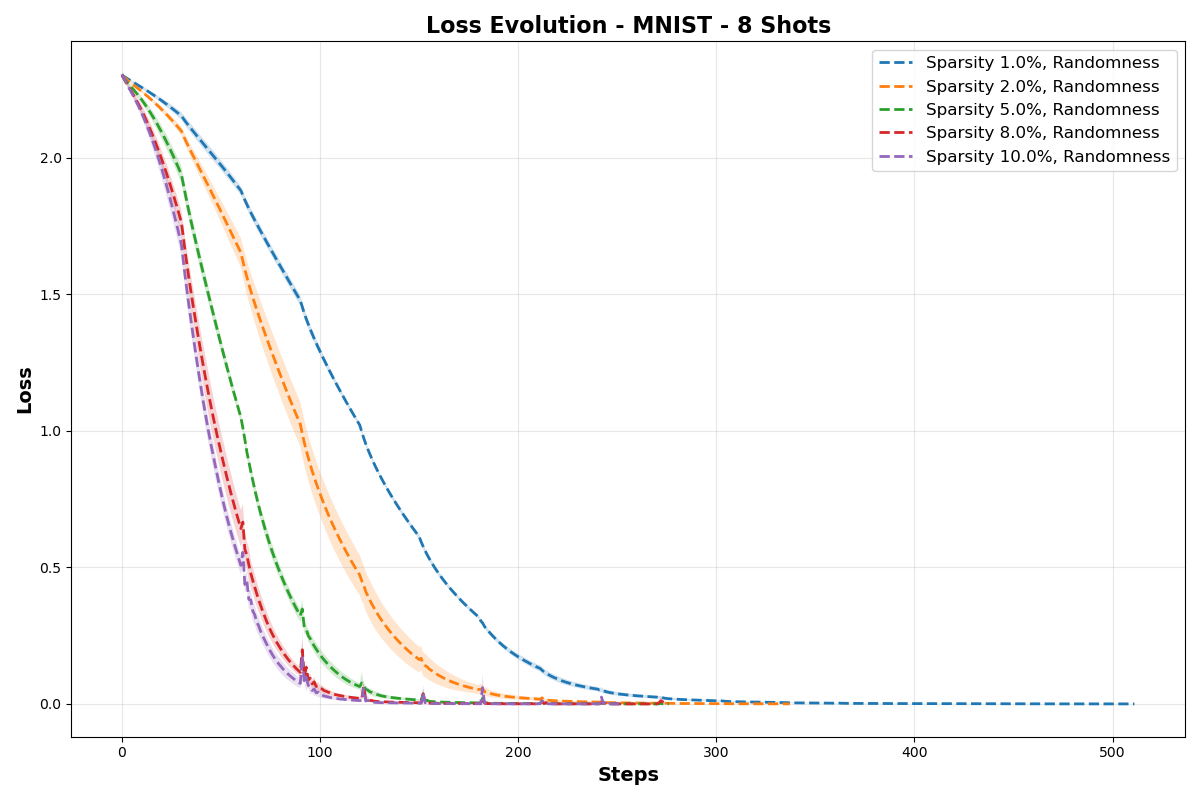}
\end{subfigure} \\ \hline
\begin{subfigure}[b]{0.9\textwidth}
    \centering
    \includegraphics[width=\linewidth]{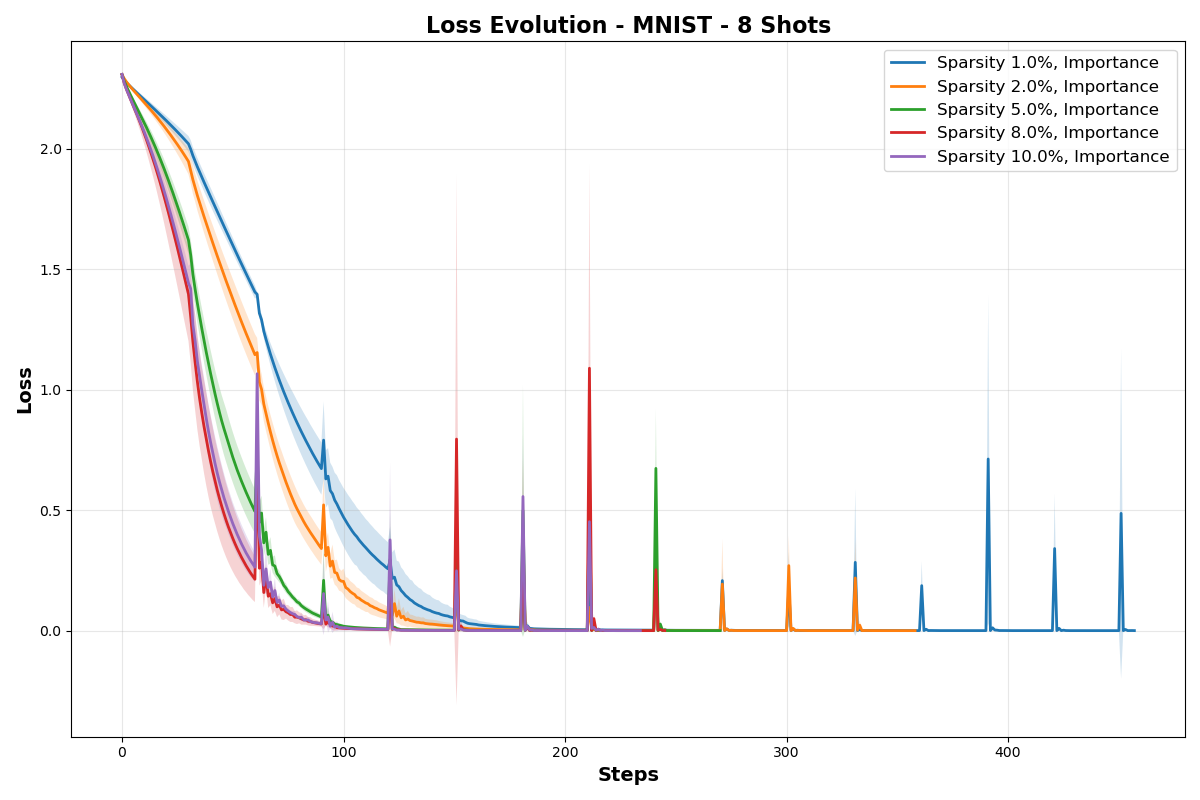}
\end{subfigure} \\ \hline
\end{tabular}
\end{adjustbox}
\end{figure*}

\begin{figure*}
\centering
\captionsetup{justification=centering, singlelinecheck=false} 
\caption{Comparison of Random and Importance Gradient Pruning in Sparse Optimization (SO) – Gradient Rank Evolution in Few-Shot Learning (16 shots) on MNIST Dataset.}
\label{fig:rank_MNIST_16_shots}
\begin{adjustbox}{max width=0.92\textwidth} 
\begin{tabular}{|c|}
\hline
\begin{subfigure}[b]{0.9\textwidth}
    \centering
    \includegraphics[width=\linewidth]{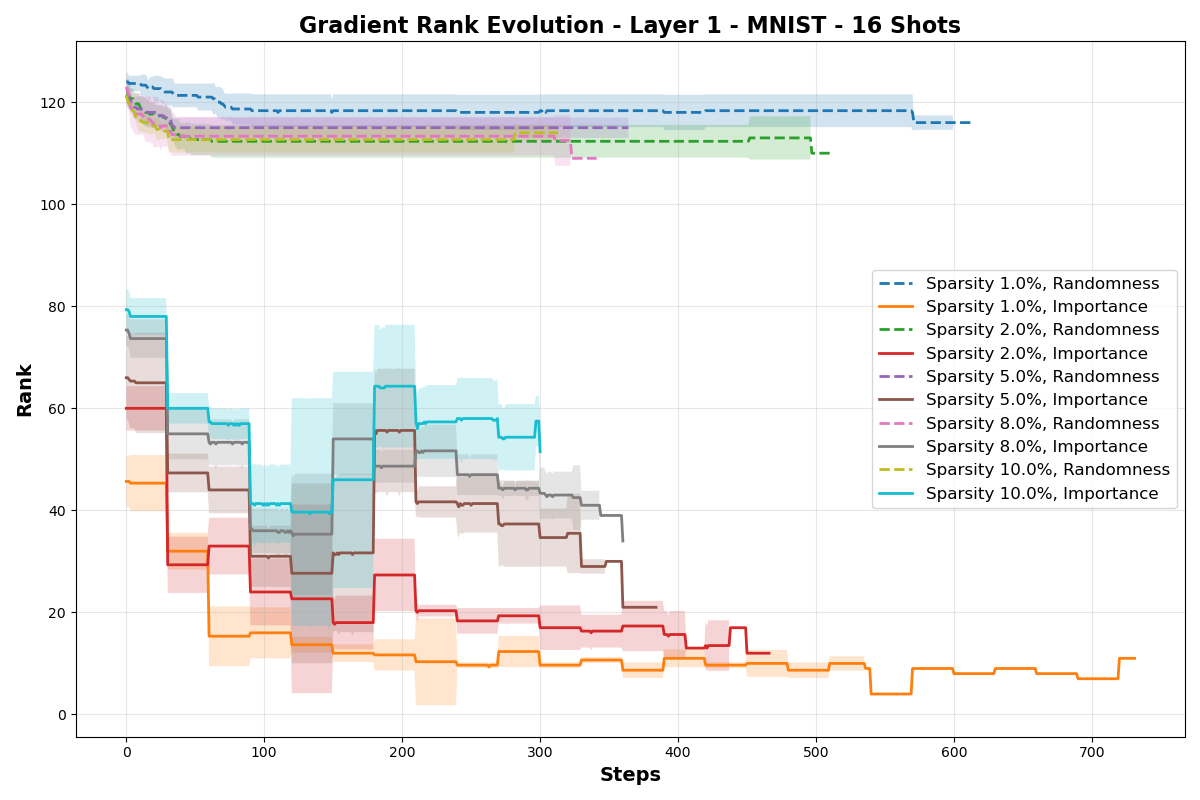}
\end{subfigure} \\ \hline
\begin{subfigure}[b]{0.9\textwidth}
    \centering
    \includegraphics[width=\linewidth]{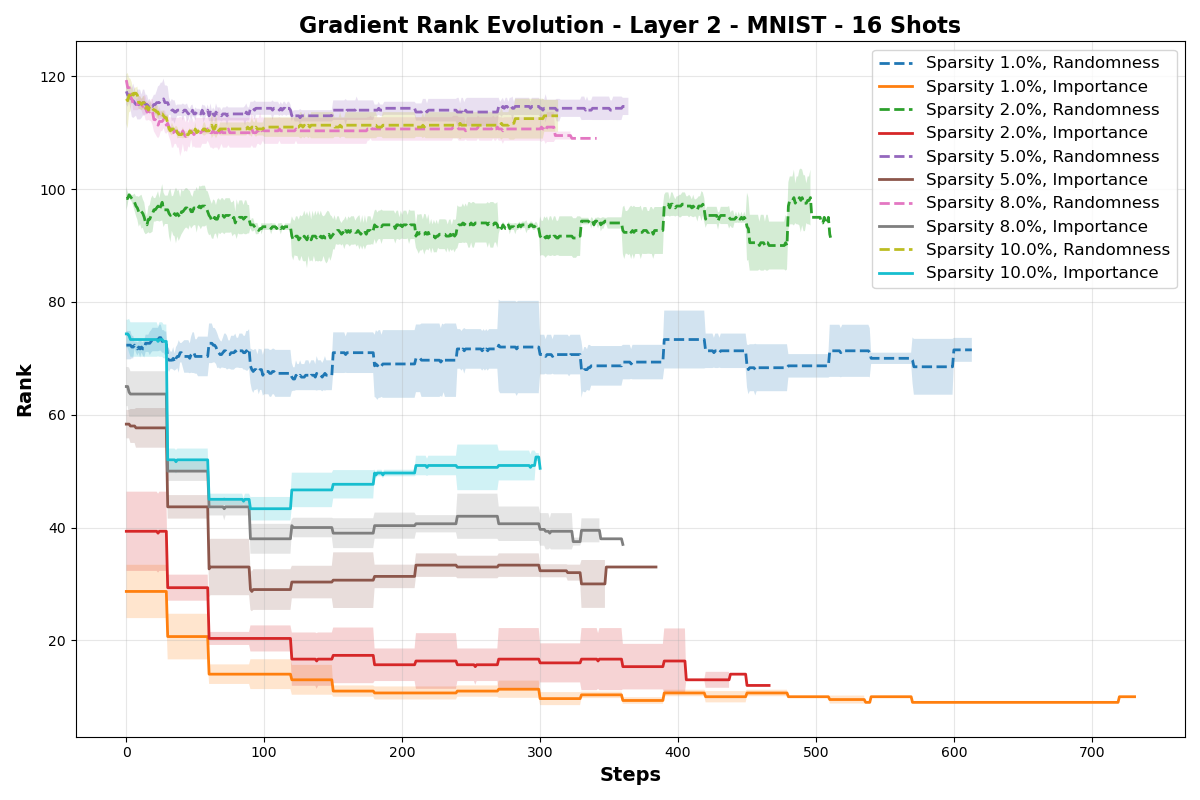}
\end{subfigure} \\ \hline
\end{tabular}
\end{adjustbox}
\end{figure*}

\begin{figure*}
\centering
\captionsetup{justification=centering, singlelinecheck=false} 
\caption{Comparison of Random and Importance Gradient Pruning in Sparse Optimization (SO) – Loss Evolution in Few-Shot Learning (16 shots) on MNIST Dataset.}
\label{fig:loss_MNIST_16_shots}
\begin{adjustbox}{max width=0.92\textwidth} 
\begin{tabular}{|c|}
\hline
\begin{subfigure}[b]{0.9\textwidth}
    \centering
    \includegraphics[width=\linewidth]{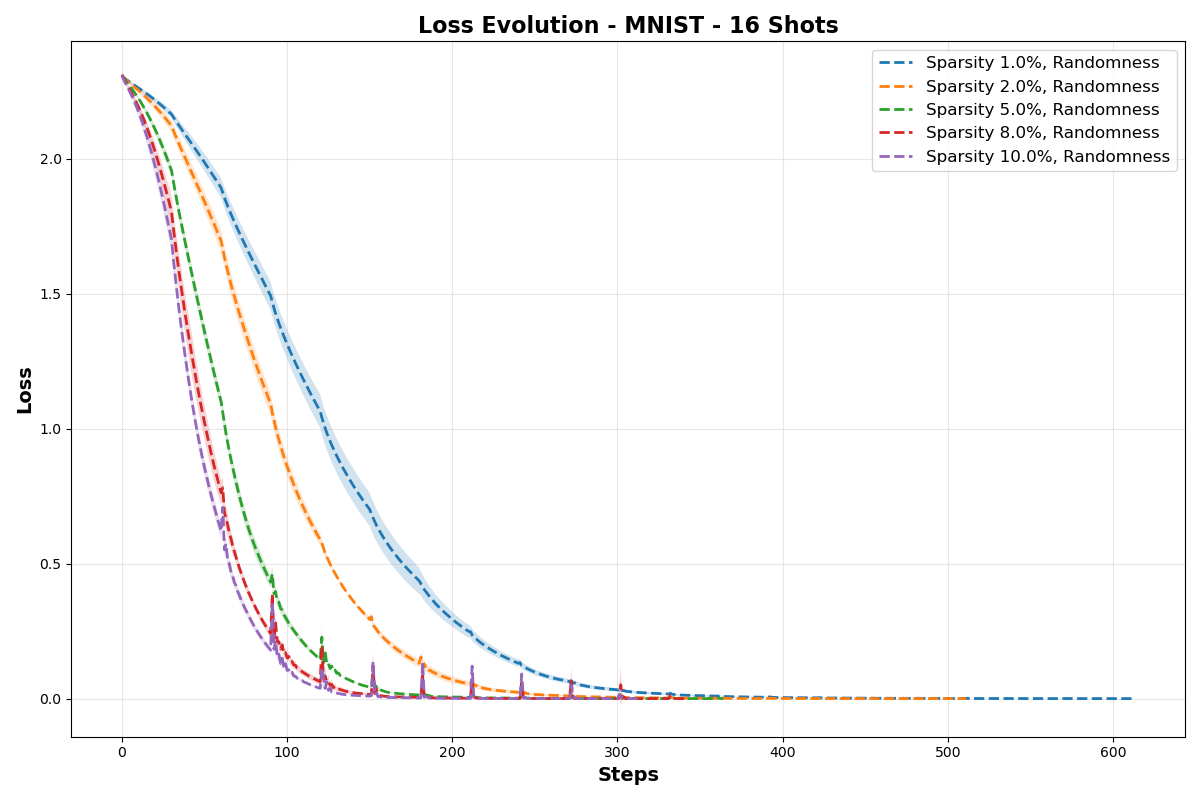}
\end{subfigure} \\ \hline
\begin{subfigure}[b]{0.9\textwidth}
    \centering
    \includegraphics[width=\linewidth]{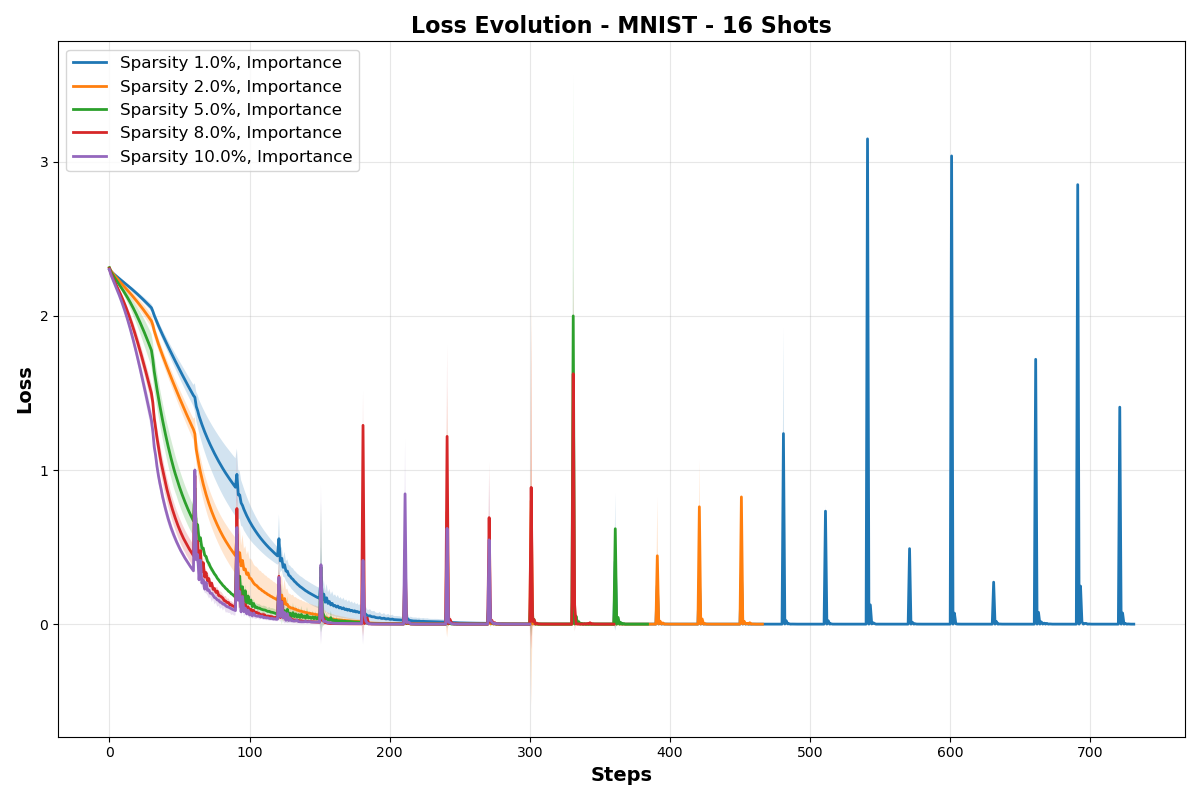}
\end{subfigure} \\ \hline
\end{tabular}
\end{adjustbox}
\end{figure*}

\textbf{Full-Rank Learning.} Figures~\ref{fig:rank_EMNIST_4_shots}--\ref{fig:loss_MNIST_16_shots} illustrate how random gradient pruning maintains a high-dimensional update space, effectively enabling full-rank learning despite extreme sparsity. 

In rank evolution plots, the gradient rank for random pruning remains close to the full rank throughout training, whereas the gradient rank for importance-based pruning often settles to a lower value. This outcome suggests that the random selection of gradient entries explores more diverse directions in parameter space, thereby preserving expressive capacity. Further, the loss curves confirm that random pruning converges stably and less rapidly, while importance-based pruning risks collapsing updates into fewer directions, potentially causing overfitting. Random gradient pruning causes a slow learning process but is more stable and less prone to overfitting, while importance-based gradient learning leads to fast learning and potential overfitting. 

Overall, these figures illustrate that sparsity —even at very low-density ratios— does not diminish the fundamental rank of the gradient and previous results confirm that sparsity does not lower the learning capacity. 

\begin{table*}[h]
\centering
\caption{Classication performance on 7 datasets with a two-layer fully-connected architecture. Results are the average top-1 accuracy of $10$ executions $\pm$ standard deviation.}
\scalebox{0.85}{
\begin{tabular}{lccccccc}
\toprule
\textbf{Model} & \textbf{EMNIST} & \textbf{MNIST} & \textbf{FMNIST} & \textbf{PathMNIST} & \textbf{OrganMNISTAxial} & \textbf{BloodMNIST} & \textbf{BreastMNIST}  \\
\midrule
GaLoRE ($r = 2$) & $80.64 \pm \scriptstyle 0.17$ & $96.06 \pm \scriptstyle 0.12$ & $86.35 \pm \scriptstyle 0.25$ & $49.41 \pm \scriptstyle 0.97$ & $67.85 \pm \scriptstyle 0.48$ & $75.57 \pm \scriptstyle 0.35$ & $77.37 \pm \scriptstyle 1.35$ \\
GaLoRE ($r = 4$) & $80.80 \pm \scriptstyle 0.16$ & $95.89 \pm \scriptstyle 0.12$ & $86.31 \pm \scriptstyle 0.20$ & $53.27 \pm \scriptstyle 0.91$ & $68.32 \pm \scriptstyle 0.44$ & $76.01 \pm \scriptstyle 0.22$ & $77.88 \pm \scriptstyle 0.87$ \\
GaLoRE ($r = 8$) & $81.29 \pm \scriptstyle 0.25$ & $96.04 \pm \scriptstyle 0.16$ & $86.68 \pm \scriptstyle 0.28$ & $53.74 \pm \scriptstyle 0.94$ & $69.19 \pm \scriptstyle 0.38$ & $77.54 \pm \scriptstyle 0.32$ & $78.65 \pm \scriptstyle 1.28$ \\
GaLoRE ($r = 16$) & $81.74 \pm \scriptstyle 0.28$ & $97.07 \pm \scriptstyle 0.13$ & $87.48 \pm \scriptstyle 0.27$ & $53.73 \pm \scriptstyle 1.44$ & $70.56 \pm \scriptstyle 0.40$ & $77.96 \pm \scriptstyle 0.38$ & $78.53 \pm \scriptstyle 1.47$ \\
\midrule
SO ($\kappa = 1\%$) & $81.29 \pm \scriptstyle 0.18$ & $97.16 \pm \scriptstyle 0.12$ & $87.46 \pm \scriptstyle 0.19$ & $53.63 \pm \scriptstyle 0.64$ & $72.67 \pm \scriptstyle 0.35$ & $78.89 \pm \scriptstyle 0.42$ & $79.62 \pm \scriptstyle 1.06$ \\
SO ($\kappa = 2\%$) & $81.55 \pm \scriptstyle 0.16$ & $97.18 \pm \scriptstyle 0.10$ & $87.67 \pm \scriptstyle 0.28$ & $53.43 \pm \scriptstyle 0.82$ & $72.19 \pm \scriptstyle 0.61$ & $79.14 \pm \scriptstyle 0.45$ & $78.78 \pm \scriptstyle 1.16$ \\
SO ($\kappa = 5\%$) & $81.63 \pm \scriptstyle 0.25$ & $97.24 \pm \scriptstyle 0.09$ & $87.66 \pm \scriptstyle 0.36$ & $54.41 \pm \scriptstyle 0.89$ & $72.11 \pm \scriptstyle 0.41$ & $78.81 \pm \scriptstyle 0.37$ & $78.65 \pm \scriptstyle 1.79$ \\
\bottomrule
\end{tabular}}
\label{tab:comparison_classification}
\end{table*}
\begin{table*}[ht]
\centering
\caption{Classication performance on 6 target datasets after pretraining on MNIST with a two-layer fully-connected architecture. Results are the average top-1 accuracy over $10$ executions $\pm$ standard deviation.}

\scalebox{0.85}{\begin{tabular}{lccccccc}
\toprule
\textbf{Model} & \textbf{EMNIST} & \textbf{FMNIST} & \textbf{PathMNIST} & \textbf{OrganMNISTAxial} & \textbf{BloodMNIST} & \textbf{BreastMNIST} \\
\midrule 
LoRA ($r = 2$) & $75.85 \pm \scriptstyle 0.29$ & $84.96 \pm \scriptstyle 0.18$ & $51.24 \pm \scriptstyle 0.73$ & $67.11 \pm \scriptstyle 0.81$ & $76.06 \pm \scriptstyle 0.18$ & $77.37 \pm \scriptstyle 1.32$ \\
LoRA ($r = 4$) & $78.51 \pm \scriptstyle 0.16$ & $85.93 \pm \scriptstyle 0.26$ & $52.03 \pm \scriptstyle 0.77$ & $69.13 \pm \scriptstyle 0.62$ & $77.09 \pm \scriptstyle 0.56$ & $78.21 \pm \scriptstyle 1.15$ \\
LoRA ($r = 8$) & $80.52 \pm \scriptstyle 0.15$ & $86.68 \pm \scriptstyle 0.28$ & $52.28 \pm \scriptstyle 0.94$ & $71.70 \pm \scriptstyle 0.53$ & $77.76 \pm \scriptstyle 0.24$ & $77.95 \pm \scriptstyle 0.92$ \\
LoRA ($r = 16$) & $81.82 \pm \scriptstyle 0.24$ & $87.03 \pm \scriptstyle 0.26$ & $52.58 \pm \scriptstyle 1.36$ & $71.92 \pm \scriptstyle 0.46$ & $78.00 \pm \scriptstyle 0.36$ & $77.63 \pm \scriptstyle 1.26$ \\
\midrule
ReLoRA ($r = 2$) & $78.67 \pm \scriptstyle 0.17$ & $85.58 \pm \scriptstyle 0.20$ & $50.19 \pm \scriptstyle 0.60$ & $70.62 \pm \scriptstyle 0.37$ & $75.79 \pm \scriptstyle 0.44$ & $78.21 \pm \scriptstyle 1.25$ \\
ReLoRA ($r = 4$) & $78.76 \pm \scriptstyle 0.18$ & $85.65 \pm \scriptstyle 0.20$ & $50.43 \pm \scriptstyle 0.43$ & $70.65 \pm \scriptstyle 0.45$ & $76.90 \pm \scriptstyle 0.47$ & $78.01 \pm \scriptstyle 1.32$ \\
ReLoRA ($r = 8$) & $78.73 \pm \scriptstyle 0.17$ & $85.60 \pm \scriptstyle 0.11$ & $50.62 \pm \scriptstyle 0.45$ & $70.80 \pm \scriptstyle 0.55$ & $78.03 \pm \scriptstyle 0.45$ & $77.82 \pm \scriptstyle 1.52$ \\
\end{tabular}}
\label{tab:comparison_acc_finetuning_classification_source_mnist}
\end{table*}

\begin{table*}[ht]
\centering
\caption{Classication performance on 6 target datasets after pretraining on FMNIST with a two-layer fully-connected architecture. Results are the average top-1 accuracy over $10$ executions $\pm$ standard deviation.}
\scalebox{0.85}{\begin{tabular}{lccccccc}
\toprule
\textbf{Model} & \textbf{EMNIST} & \textbf{MNIST} & \textbf{PathMNIST} & \textbf{OrganMNISTAxial} & \textbf{BloodMNIST} & \textbf{BreastMNIST} \\
\midrule 
LoRA ($r = 2$) & $74.83 \pm \scriptstyle 0.18$ & $94.80 \pm \scriptstyle 0.25$ & $50.69 \pm \scriptstyle 0.70$ & $68.14 \pm \scriptstyle 0.34$ & $74.89 \pm \scriptstyle 0.46$ & $77.95 \pm \scriptstyle 1.93$ \\
LoRA ($r = 4$) & $77.44 \pm \scriptstyle 0.20$ & $95.73 \pm \scriptstyle 0.15$ & $50.43 \pm \scriptstyle 0.54$ & $69.97 \pm \scriptstyle 0.41$ & $76.40 \pm \scriptstyle 0.21$ & $78.21 \pm \scriptstyle 1.15$ \\
LoRA ($r = 8$) & $79.59 \pm \scriptstyle 0.16$ & $96.32 \pm \scriptstyle 0.12$ & $50.73 \pm \scriptstyle 1.09$ & $70.73 \pm \scriptstyle 0.26$ & $76.30 \pm \scriptstyle 0.58$ & $78.59 \pm \scriptstyle 0.96$ \\
LoRA ($r = 16$) & $81.18 \pm \scriptstyle 0.14$ & $96.48 \pm \scriptstyle 0.16$ & $51.52 \pm \scriptstyle 0.61$ & $71.17 \pm \scriptstyle 0.55$ & $76.89 \pm \scriptstyle 0.66$ & $78.65 \pm \scriptstyle 1.46$ \\
\midrule
ReLoRA ($r = 2$) & $77.79 \pm \scriptstyle 0.18$ & $95.39 \pm \scriptstyle 0.12$ & $49.99 \pm \scriptstyle 0.39$ & $70.63 \pm \scriptstyle 0.38$ & $74.93 \pm \scriptstyle 0.44$ & $78.01 \pm \scriptstyle 1.55$ \\
ReLoRA ($r = 4$) & $77.74 \pm \scriptstyle 0.24$ & $95.29 \pm \scriptstyle 0.14$ & $50.01 \pm \scriptstyle 0.59$ & $70.49 \pm \scriptstyle 0.37$ & $75.97 \pm \scriptstyle 0.48$ & $78.14 \pm \scriptstyle 0.97$ \\
ReLoRA ($r = 8$) & $77.76 \pm \scriptstyle 0.22$ & $95.36 \pm \scriptstyle 0.10$ & $50.18 \pm \scriptstyle 0.62$ & $70.62 \pm \scriptstyle 0.42$ & $76.45 \pm \scriptstyle 0.48$ & $79.42 \pm \scriptstyle 1.61$ \\
\end{tabular}}
\label{tab:comparison_acc_finetuning_classification_source_fmnist}
\end{table*}

\begin{table*}[ht]
\centering
\caption{Classication performance on 7 datasets with a two-layer fully-connected architecture. Results are the average top-1 accuracy over $10$ executions $\pm$ standard deviation.} 
\scalebox{0.85}{\begin{tabular}{lccccccc}
\toprule
\textbf{Strategy} & \textbf{EMNIST} & \textbf{MNIST} & \textbf{FMNIST} & \textbf{PathMNIST} & \textbf{OrganMNISTAxial} & \textbf{BloodMNIST} & \textbf{BreastMNIST} \\
\midrule
\multicolumn{8}{l}{\textbf{Importance-Based Gradient Pruning}} \\
SO ($\kappa = 1\%$) & $76.81 \pm \scriptstyle 0.45$ & $95.05 \pm \scriptstyle 0.44$ & $85.42 \pm \scriptstyle 0.38$ & $50.22 \pm \scriptstyle 1.23$ & $67.34 \pm \scriptstyle 0.84$ & $74.60 \pm \scriptstyle 0.72$ & $77.63 \pm \scriptstyle 2.28$ \\
SO ($\kappa = 2\%$) & $78.52 \pm \scriptstyle 0.78$ & $95.46 \pm \scriptstyle 0.37$ & $86.29 \pm \scriptstyle 0.37$ & $52.18 \pm \scriptstyle 0.48$ & $67.90 \pm \scriptstyle 0.31$ & $75.18 \pm \scriptstyle 0.59$ & $79.55 \pm \scriptstyle 1.56$ \\
SO ($\kappa = 5\%$) & $80.05 \pm \scriptstyle 0.23$ & $96.27 \pm \scriptstyle 0.26$ & $86.70 \pm \scriptstyle 0.20$ & $53.19 \pm \scriptstyle 0.77$ & $68.17 \pm \scriptstyle 0.75$ & $75.68 \pm \scriptstyle 0.62$ & $78.53 \pm \scriptstyle 1.44$ \\
SO ($\kappa = 8\%$) & $80.59 \pm \scriptstyle 0.38$ & $96.51 \pm \scriptstyle 0.20$ & $86.94 \pm \scriptstyle 0.33$ & $53.39 \pm \scriptstyle 0.76$ & $69.33 \pm \scriptstyle 0.66$ & $75.88 \pm \scriptstyle 0.92$ & $76.92 \pm \scriptstyle 2.48$ \\
SO ($\kappa = 10\%$) & $80.20 \pm \scriptstyle 0.45$ & $96.55 \pm \scriptstyle 0.20$ & $87.30 \pm \scriptstyle 0.34$ & $53.65 \pm \scriptstyle 0.50$ & $69.89 \pm \scriptstyle 0.55$ & $76.35 \pm \scriptstyle 0.34$ & $77.05 \pm \scriptstyle 6.52$ \\
\midrule
\multicolumn{8}{l}{\textbf{Random Gradient Pruning}} \\
\bottomrule
\end{tabular}}
\label{tab:ablation_so_classification}
\end{table*}

\begin{table*}[ht]
\centering
\caption{Classication performance on 6 target datasets after pretraining on MNIST with a two-layer fully-connected architecture. Results are the average top-1 accuracy over $10$ executions $\pm$ standard deviation.}

\scalebox{0.85}{\begin{tabular}{lcccccc}
\toprule
\textbf{Strategy} & \textbf{EMNIST} & \textbf{FMNIST} & \textbf{PathMNIST} & \textbf{OrganMNISTAxial} & \textbf{BloodMNIST} & \textbf{BreastMNIST} \\
\midrule
\multicolumn{7}{l}{\textbf{Importance-Based Gradient Pruning}} \\
SO ($\kappa = 1\%$) & $79.89 \pm \scriptstyle 0.26$ & $86.54 \pm \scriptstyle 0.34$ & $51.42 \pm \scriptstyle 1.51$ & $69.82 \pm \scriptstyle 0.46$ & $76.09 \pm \scriptstyle 0.42$ & $81.35 \pm \scriptstyle 1.89$ \\
SO ($\kappa = 2\%$) & $80.12 \pm \scriptstyle 0.33$ & $86.54 \pm \scriptstyle 0.38$ & $51.15 \pm \scriptstyle 2.15$ & $69.95 \pm \scriptstyle 0.80$ & $76.17 \pm \scriptstyle 0.47$ & $81.35 \pm \scriptstyle 1.56$ \\
SO ($\kappa = 5\%$) & $80.23 \pm \scriptstyle 0.45$ & $87.08 \pm \scriptstyle 0.28$ & $51.54 \pm \scriptstyle 1.75$ & $70.60 \pm \scriptstyle 0.68$ & $76.45 \pm \scriptstyle 0.55$ & $81.41 \pm \scriptstyle 1.22$ \\
SO ($\kappa = 8\%$) & $80.50 \pm \scriptstyle 0.41$ & $87.41 \pm \scriptstyle 0.28$ & $53.23 \pm \scriptstyle 1.22$ & $71.01 \pm \scriptstyle 0.53$ & $77.34 \pm \scriptstyle 0.50$ & $82.05 \pm \scriptstyle 1.81$ \\
SO ($\kappa = 10\%$) & $80.80 \pm \scriptstyle 0.35$ & $87.14 \pm \scriptstyle 0.30$ & $53.66 \pm \scriptstyle 0.87$ & $71.08 \pm \scriptstyle 0.57$ & $77.27 \pm \scriptstyle 0.77$ & $80.77 \pm \scriptstyle 2.20$ \\
\midrule
\multicolumn{7}{l}{\textbf{Random Gradient Pruning}} \\
\bottomrule
\end{tabular}}
\label{tab:ablation_so_adaptation_classification_MNIST}
\end{table*}

\begin{table*}[ht]
\centering
\caption{Classication performance on 6 target datasets after pretraining on FMNIST with a two-layer fully-connected architecture. Results are the average top-1 accuracy over $10$ executions $\pm$ standard deviation.}
\scalebox{0.85}{\begin{tabular}{lcccccc}
\toprule
\textbf{Strategy} & \textbf{EMNIST} & \textbf{MNIST} & \textbf{PathMNIST} & \textbf{OrganMNISTAxial} & \textbf{BloodMNIST} & \textbf{BreastMNIST} \\
\midrule
\multicolumn{7}{l}{\textbf{Importance-Based Gradient Pruning}} \\
SO ($\kappa = 1\%$) & $79.13 \pm \scriptstyle 0.45$ & $96.05 \pm \scriptstyle 0.16$ & $51.80 \pm \scriptstyle 0.92$ & $70.15 \pm \scriptstyle 0.53$ & $75.95 \pm \scriptstyle 0.60$ & $79.42 \pm \scriptstyle 2.04$ \\
SO ($\kappa = 2\%$) & $79.58 \pm \scriptstyle 0.36$ & $96.16 \pm \scriptstyle 0.14$ & $52.25 \pm \scriptstyle 0.95$ & $70.23 \pm \scriptstyle 0.55$ & $76.00 \pm \scriptstyle 0.89$ & $79.04 \pm \scriptstyle 1.81$ \\
SO ($\kappa = 5\%$) & $80.21 \pm \scriptstyle 0.32$ & $96.27 \pm \scriptstyle 0.41$ & $52.13 \pm \scriptstyle 2.10$ & $70.33 \pm \scriptstyle 0.54$ & $76.81 \pm \scriptstyle 0.61$ & $79.62 \pm \scriptstyle 1.90$ \\
SO ($\kappa = 8\%$) & $80.42 \pm \scriptstyle 0.42$ & $96.60 \pm \scriptstyle 0.20$ & $52.26 \pm \scriptstyle 1.24$ & $71.05 \pm \scriptstyle 0.77$ & $76.83 \pm \scriptstyle 0.96$ & $78.59 \pm \scriptstyle 1.72$ \\
SO ($\kappa = 10\%$) & $80.69 \pm \scriptstyle 0.30$ & $96.71 \pm \scriptstyle 0.40$ & $53.64 \pm \scriptstyle 0.93$ & $71.50 \pm \scriptstyle 0.60$ & $77.20 \pm \scriptstyle 0.66$ & $79.29 \pm \scriptstyle 1.77$ \\
\midrule
\multicolumn{7}{l}{\textbf{Random Gradient Pruning}} \\
\bottomrule
\end{tabular}}
\label{tab:ablation_so_adaptation_classification_FMNIST}
\end{table*}

\begin{table*}
\centering
\caption{Few-shot classification performance on 7 datasets using a two-layer fully-connected architecture without pretraining. Results are the average top-1 accuracy of $10$ executions $\pm$ standard deviation.}
\scalebox{0.81}{\begin{tabular}{lccccccccc}
\toprule
\textbf{Shots} & \textbf{Model} & \textbf{EMNIST} & \textbf{MNIST} & \textbf{FMNIST} & \textbf{PathMNIST} & \textbf{OrganMNISTAxial} & \textbf{BloodMNIST} & \textbf{BreastMNIST} \\
\midrule
\multirow{5}{*}{4}  
& GaLoRE ($r = 2$) & $28.71 \pm \scriptstyle 1.75$ & $54.79 \pm \scriptstyle 3.09$ & $59.38 \pm \scriptstyle 2.33$ & $26.20 \pm \scriptstyle 3.54$ & $39.49 \pm \scriptstyle 1.59$ & $47.75 \pm \scriptstyle 3.64$ & $58.46 \pm \scriptstyle 9.95$ \\
& GaLoRE ($r = 4$) & $27.12 \pm \scriptstyle 1.85$ & $56.71 \pm \scriptstyle 3.98$ & $58.99 \pm \scriptstyle 2.96$ & $23.66 \pm \scriptstyle 3.23$ & $37.90 \pm \scriptstyle 2.45$ & $46.08 \pm \scriptstyle 2.46$ &  $57.82 \pm \scriptstyle 10.08$  \\
& GaLoRE ($r = 8$) & $28.94 \pm \scriptstyle 1.82$ & $62.78 \pm \scriptstyle 3.06$ & $60.57 \pm \scriptstyle 2.48$ & $22.26 \pm \scriptstyle 2.99$ & $39.59 \pm \scriptstyle 1.55$ & $43.85 \pm \scriptstyle 3.26$ & $58.33 \pm \scriptstyle 10.47$  \\
& GaLoRE ($r = 16$) & $32.77 \pm \scriptstyle 1.51$ & $62.80 \pm \scriptstyle 3.03$ & $61.56 \pm \scriptstyle 2.48$ & $21.49 \pm \scriptstyle 3.13$ & $40.37 \pm \scriptstyle 2.03$ & $44.76 \pm \scriptstyle 2.98$ &  $57.76 \pm \scriptstyle 10.32$  \\
\cmidrule(lr){2-9} 
& SO ($\kappa = 1\%$) & $34.17 \pm \scriptstyle 1.83$ & $63.20 \pm \scriptstyle 3.03$ & $61.15 \pm \scriptstyle 2.78$ & $26.61 \pm \scriptstyle 5.09$ & $40.72 \pm \scriptstyle 1.79$ & $46.47 \pm \scriptstyle 2.64$ & $57.88 \pm \scriptstyle 11.01$ \\
& SO ($\kappa = 2\%$) & $34.38 \pm \scriptstyle 1.81$ & $63.14 \pm \scriptstyle 3.10$ & $61.32 \pm \scriptstyle 2.79$ & $25.77 \pm \scriptstyle 4.08$ & $41.25 \pm \scriptstyle 2.33$ & $45.94 \pm \scriptstyle 2.47$ & $57.12 \pm \scriptstyle 10.52$ \\
& SO ($\kappa = 5\%$) & $34.07 \pm \scriptstyle 1.84$ & $62.98 \pm \scriptstyle 2.52$ & $61.81 \pm \scriptstyle 2.21$ & $22.93 \pm \scriptstyle 4.88$ & $40.92 \pm \scriptstyle 2.37$ & $46.94 \pm \scriptstyle 2.50$ & $56.86 \pm \scriptstyle 10.26$ \\
\bottomrule
\end{tabular}}
\label{tab:comparison_acc_few_shot}
\end{table*}
\begin{table*}
\centering
\caption{Few-shot classification performance on 6 datasets using a two-layer fully-connected architecture after pretraining on MNIST. Results are the average top-1 accuracy over $10$ executions $\pm$ standard deviation.}

\scalebox{0.85}{\begin{tabular}{lccccccc}
\toprule
\textbf{Shots} & \textbf{Model} & \textbf{EMNIST} & \textbf{FMNIST} & \textbf{PathMNIST} & \textbf{OrganMNISTAxial} & \textbf{BloodMNIST} & \textbf{BreastMNIST} \\
\midrule
\multirow{5}{*}{4}  & LoRA ($r = 2$) & $28.02 \pm \scriptstyle 1.32$ & $55.36 \pm \scriptstyle 2.30$ & $26.69 \pm \scriptstyle 2.21$  & $31.73 \pm \scriptstyle 2.66$ & $45.21 \pm \scriptstyle 4.09$ & $61.28 \pm \scriptstyle 7.25$ \\
                    & LoRA ($r = 4$) & $28.44 \pm \scriptstyle  1.32$ & $55.21 \pm \scriptstyle 2.97$ & $26.81 \pm \scriptstyle 2.45$  & $34.10 \pm \scriptstyle 2.05$ & $ 45.79 \pm \scriptstyle 2.60$ & $60.71 \pm \scriptstyle 8.61$ \\
                    & LoRA  ($r = 8$) & $ 29.08 \pm \scriptstyle 1.23$ & $56.70 \pm \scriptstyle 2.62$ & $27.50 \pm \scriptstyle 2.72$  & $34.87 \pm \scriptstyle 1.70$ & $45.41 \pm \scriptstyle 2.47$ & $60.13 \pm \scriptstyle 8.21$ \\
                    & LoRA ($r = 16$) & $29.14 \pm \scriptstyle 1.42$ & $55.60 \pm \scriptstyle 2.80$  & $27.19 \pm \scriptstyle 1.61$  & $34.74 \pm \scriptstyle 1.12$ & $46.23 \pm \scriptstyle 3.38$ & $60.19 \pm \scriptstyle 8.96$ \\
                    \cmidrule(lr){2-8} 
                    & ReLoRA ($r = 2$) & $27.84 \pm \scriptstyle 1.42$ & $55.36 \pm \scriptstyle 2.30$ & $26.69 \pm \scriptstyle 2.21$  & $31.73 \pm \scriptstyle 2.66$ & $44.61 \pm \scriptstyle 3.36$ & $60.90 \pm \scriptstyle 9.48$ \\
                    & ReLoRA ($r = 4$) & $28.28 \pm \scriptstyle 1.45$ & $55.21 \pm \scriptstyle 2.97$ & $26.81 \pm \scriptstyle 2.45$  & $34.10 \pm \scriptstyle 2.05$ & $45.52 \pm \scriptstyle 2.29$ & $60.64 \pm \scriptstyle 7.82$ \\
                    & ReLoRA ($r = 8$) & $28.89 \pm \scriptstyle 1.32$ & $56.70 \pm \scriptstyle 2.62$ & $27.50 \pm \scriptstyle 2.72$  & $34.87 \pm \scriptstyle 1.70$ & $45.14 \pm \scriptstyle 3.23$ & $59.87 \pm \scriptstyle 7.60$ \\
\bottomrule
\end{tabular}}
\label{tab:comparison_acc_finetuning_few_shot_source_mnist}
\end{table*}

\begin{table*}
\centering
\caption{Few-shot classification performance on 6 datasets using a two-layer fully-connected architecture after pretraining on FMNIST. Results are the average top-1 accuracy over $10$ executions $\pm$ standard deviation.}
\scalebox{0.85}{\begin{tabular}{lccccccc}
\toprule
\textbf{Shots} & \textbf{Model} & \textbf{EMNIST} & \textbf{MNIST} & \textbf{PathMNIST} & \textbf{OrganMNISTAxial} & \textbf{BloodMNIST} & \textbf{BreastMNIST}\\
\midrule
\multirow{5}{*}{4}  & LoRA ($r = 2$) & $29.31 \pm \scriptstyle 1.43$ & $54.60 \pm \scriptstyle 3.53$ & $28.00 \pm \scriptstyle 2.63$ & $35.65 \pm \scriptstyle 2.42$ & $40.59 \pm \scriptstyle 2.68$ & $60.38 \pm \scriptstyle 9.03$ \\
                    & LoRA ($r = 4$) & $29.55 \pm \scriptstyle 1.19$ & $54.73 \pm \scriptstyle 2.93$ & $28.04 \pm \scriptstyle 2.62$ & $36.13 \pm \scriptstyle 1.63$ & $42.27 \pm \scriptstyle 2.08$ & $57.88 \pm \scriptstyle 9.11$ \\
                    & LoRA  ($r = 8$) & $30.01 \pm \scriptstyle 1.68$ & $54.84 \pm \scriptstyle 3.41$ & $28.61 \pm \scriptstyle 2.12$ & $36.43 \pm \scriptstyle 1.57$ & $42.19 \pm \scriptstyle 3.11$ & $57.76 \pm \scriptstyle 8.97$ \\
                    & LoRA ($r = 16$) & $30.21 \pm \scriptstyle 1.46$ & $55.64 \pm \scriptstyle 3.45$ & $28.85 \pm \scriptstyle 1.96$ & $36.03 \pm \scriptstyle 1.61$ & $41.74 \pm \scriptstyle 2.83$ & $58.46 \pm \scriptstyle 9.96$ \\
                    \cmidrule(lr){2-8} 
                    & ReLoRA ($r = 2$) & $29.32 \pm \scriptstyle 1.43$ & $54.60 \pm \scriptstyle 3.53$ & $28.00 \pm \scriptstyle 2.63$ & $35.65 \pm \scriptstyle 2.42$ & $40.59 \pm \scriptstyle 2.68$ & $60.38 \pm \scriptstyle 9.03$ \\
                    & ReLoRA ($r = 4$) & $29.46 \pm \scriptstyle 1.32$ & $54.73 \pm \scriptstyle 2.93$ & $28.04 \pm \scriptstyle 2.62$ & $36.13 \pm \scriptstyle 1.63$ & $42.27 \pm \scriptstyle 2.08$ & $57.88 \pm \scriptstyle 9.11$ \\
                    & ReLoRA ($r = 8$) & $29.85 \pm \scriptstyle 1.64$ & $54.84 \pm \scriptstyle 3.41$ & $28.61 \pm \scriptstyle 2.12$ & $36.43 \pm \scriptstyle 1.57$ & $42.19 \pm \scriptstyle 3.11$ & $57.76 \pm \scriptstyle 8.97$ \\
\bottomrule
\end{tabular}}
\label{tab:comparison_acc_finetuning_few_shot_source_fmnist}
\end{table*}
\begin{table*}[ht]
\centering
\caption{Few-shot classification performance on 7 datasets using a two-layer fully-connected architecture without pretraining. Results are the average top-1 accuracy of $10$ executions $\pm$ standard deviation.}
\scalebox{0.85}{\begin{tabular}{llccccccc}
\toprule
\textbf{Shots} & \textbf{Strategy} & \textbf{EMNIST} & \textbf{MNIST} & \textbf{FMNIST} & \textbf{PathMNIST} & \textbf{OrganMNISTAxial} & \textbf{BloodMNIST} & \textbf{BreastMNIST} \\
\midrule
\multirow{15}{*}{4} 
& \multicolumn{8}{l}{\textbf{Importance-Based Gradient Pruning}} \\
& SO ($\kappa = 1\%$) & $26.59 \pm \scriptstyle 2.77$ & $58.43 \pm \scriptstyle 3.20$ & $58.71 \pm \scriptstyle 2.69$ & $23.85 \pm \scriptstyle 4.78$ & $38.12 \pm \scriptstyle 2.13$ & $46.15 \pm \scriptstyle 3.25$ & $59.23 \pm \scriptstyle 8.68$ \\
& SO ($\kappa = 2\%$) & $29.31 \pm \scriptstyle 1.76$ & $59.94 \pm \scriptstyle 4.18$ & $59.44 \pm \scriptstyle 2.70$ & $23.83 \pm \scriptstyle 4.50$ & $40.27 \pm \scriptstyle 1.80$ & $46.07 \pm \scriptstyle 2.05$ & $57.50 \pm \scriptstyle 8.26$ \\
& SO ($\kappa = 5\%$) & $30.44 \pm \scriptstyle 3.35$ & $61.16 \pm \scriptstyle 4.29$ & $61.32 \pm \scriptstyle 2.23$ & $22.77 \pm \scriptstyle 4.89$ & $39.86 \pm \scriptstyle 1.73$ & $46.16 \pm \scriptstyle 3.43$ & $56.99 \pm \scriptstyle 9.70$ \\
& SO ($\kappa = 8\%$) & $31.97 \pm \scriptstyle 1.52$ & $61.50 \pm \scriptstyle 3.76$ & $60.97 \pm \scriptstyle 2.91$ & $23.39 \pm \scriptstyle 5.31$ & $40.32 \pm \scriptstyle 1.29$ & $47.21 \pm \scriptstyle 2.92$ & $57.63 \pm \scriptstyle 9.66$ \\
& SO ($\kappa = 10\%$) & $32.23 \pm \scriptstyle 1.64$ & $61.80 \pm \scriptstyle 2.85$ & $61.26 \pm \scriptstyle 2.49$ & $21.74 \pm \scriptstyle 3.60$ & $39.92 \pm \scriptstyle 1.50$ & $46.61 \pm \scriptstyle 3.39$ & $56.73 \pm \scriptstyle 9.65$ \\
& \multicolumn{8}{l}{\textbf{Random Gradient Pruning}} \\
\bottomrule
\end{tabular}}
\label{tab:ablation_study_SO_shots}
\end{table*}

\begin{table*}[ht]
\centering
\caption{Few-shot classification performance on 6 datasets using a two-layer fully-connected architecture after pretraining on MNIST. Results are the average top-1 accuracy over $10$ executions $\pm$ standard deviation.}
\scalebox{0.87}{\begin{tabular}{llcccccc}
\toprule
\textbf{Shots} & \textbf{Strategy} & \textbf{EMNIST} & \textbf{FMNIST} & \textbf{PathMNIST} & \textbf{OrganMNISTAxial} & \textbf{BloodMNIST} & \textbf{BreastMNIST} \\
\midrule
\multirow{15}{*}{4} 
& \multicolumn{7}{l}{\textbf{Importance-Based Gradient Pruning}} \\
& SO ($\kappa = 1\%$) & $30.81 \pm \scriptstyle 1.75$ & $59.46 \pm \scriptstyle 2.45$ & $24.87 \pm \scriptstyle 3.73$ & $35.45 \pm \scriptstyle 2.31$ & $44.15 \pm \scriptstyle 3.04$ & $60.71 \pm \scriptstyle 7.35$ \\
& SO ($\kappa = 2\%$) & $31.25 \pm \scriptstyle 1.25$ & $59.37 \pm \scriptstyle 2.65$ & $23.00 \pm \scriptstyle 2.96$ & $36.52 \pm \scriptstyle 1.83$ & $44.72 \pm \scriptstyle 3.01$ & $61.03 \pm \scriptstyle 8.59$ \\
& SO ($\kappa = 5\%$) & $32.17 \pm \scriptstyle 1.69$ & $59.44 \pm \scriptstyle 2.54$ & $22.84 \pm \scriptstyle 3.37$ & $36.91 \pm \scriptstyle 1.56$ & $44.40 \pm \scriptstyle 3.12$ & $60.13 \pm \scriptstyle 7.87$ \\
& SO ($\kappa = 8\%$) & $32.02 \pm \scriptstyle 2.42$ & $60.23 \pm \scriptstyle 2.18$ & $23.01 \pm \scriptstyle 3.31$ & $36.54 \pm \scriptstyle 1.94$ & $44.84 \pm \scriptstyle 1.62$ & $61.28 \pm \scriptstyle 6.62$ \\
& SO ($\kappa = 10\%$) & $31.72 \pm \scriptstyle 1.46$ & $60.12 \pm \scriptstyle 2.51$ & $22.77 \pm \scriptstyle 2.92$ & $36.12 \pm \scriptstyle 2.00$ & $44.60 \pm \scriptstyle 2.65$ & $60.26 \pm \scriptstyle 9.27$ \\
\cmidrule(lr){2-8}
& \multicolumn{7}{l}{\textbf{Random Gradient Pruning}} \\
\end{tabular}}
\label{tab:ablation_study_adaptation_MNIST}
\end{table*}

\begin{table*}[ht]
\centering
\caption{Few-shot classification performance on 6 datasets using a two-layer fully-connected architecture after pretraining on FMNIST. Results are the average top-1 accuracy over $10$ executions $\pm$ standard deviation.}
\scalebox{0.87}{\begin{tabular}{llccccccc}
\toprule
\textbf{Shots} & \textbf{Strategy} & \textbf{EMNIST} & \textbf{MNIST} & \textbf{PathMNIST} & \textbf{OrganMNISTAxial} & \textbf{BloodMNIST} & \textbf{BreastMNIST} \\
\midrule
\multirow{15}{*}{4} 
& \multicolumn{7}{l}{\textbf{Importance-Based Gradient Pruning}} \\
& SO ($\kappa = 1\%$) & $31.08 \pm \scriptstyle 0.86$ & $57.27 \pm \scriptstyle 3.66$ & $24.99 \pm \scriptstyle 2.72$ & $36.25 \pm \scriptstyle 1.71$ & $42.15 \pm \scriptstyle 3.21$ & $57.50 \pm \scriptstyle 8.22$ \\
& SO ($\kappa = 2\%$) & $31.44 \pm \scriptstyle 1.06$ & $58.28 \pm \scriptstyle 3.61$ & $25.26 \pm \scriptstyle 2.91$ & $36.74 \pm \scriptstyle 1.81$ & $42.56 \pm \scriptstyle 3.83$ & $57.12 \pm \scriptstyle 7.89$ \\
& SO ($\kappa = 5\%$) & $32.12 \pm \scriptstyle 1.28$ & $59.20 \pm \scriptstyle 3.82$ & $23.97 \pm \scriptstyle 2.64$ & $37.71 \pm \scriptstyle 1.49$ & $43.18 \pm \scriptstyle 2.93$ & $58.01 \pm \scriptstyle 7.08$ \\
& SO ($\kappa = 8\%$) & $31.48 \pm \scriptstyle 2.11$ & $59.55 \pm \scriptstyle 2.78$ & $22.94 \pm \scriptstyle 2.38$ & $37.20 \pm \scriptstyle 2.13$ & $42.36 \pm \scriptstyle 2.86$ & $56.79 \pm \scriptstyle 7.75$ \\
& SO ($\kappa = 10\%$) & $33.19 \pm \scriptstyle 1.36$ & $58.83 \pm \scriptstyle 3.85$ & $23.68 \pm \scriptstyle 2.32$ & $37.61 \pm \scriptstyle 1.74$ & $42.50 \pm \scriptstyle 2.92$ & $58.21 \pm \scriptstyle 9.28$ \\
\cmidrule(lr){2-8}
& \multicolumn{7}{l}{\textbf{Random Gradient Pruning}} \\
\end{tabular}}
\label{tab:ablation_study_adaptation_FMNIST}
\end{table*}

\end{document}